\documentclass[%
	a4paper,
	headings=optiontotocandhead,
	]{scrartcl}

\makeatletter
\@ifclasslater{\KOMAClassName}{2019/10/12}{%
}{%
}
\makeatother

\usepackage[english]{babel}
\usepackage[%
	backend=bibtex8,
	style=numeric,
	maxbibnames=99
	]{biblatex} % Literaturverzeichnis und Zitation
\usepackage{booktabs} % "schoene" Tabellen ermoeglichen
\usepackage[T1]{fontenc} % interne Font-Generierung
\usepackage[hang]{footmisc} % haengende Fussnoten
\usepackage{graphicx} % Abbildungen ermoeglichen
\usepackage[utf8]{inputenc} % Codierung der Eingabe
\usepackage[scaled]{beramono} % for typewriter family (ttfamily) as used in listings
\usepackage{lmodern} % Schriftfamilie Latin Modern. Als letzten Schrift laden, damit es die aktive ist.
\usepackage{longtable} % "lange" Tabellen ermoeglichen
\usepackage{makeidx} % Index ermoeglichen
\usepackage[%
	expansion=true,
	protrusion=true,
	tracking=true
	]{microtype} % mikrotypographische Feinheiten
\usepackage{setspace} % Zeilenabstand kontrollieren
\usepackage{enumitem} % enumerate/itemize/description modifizieren
\usepackage{textcomp} % zusaetzliche LaTeX-Sonderzeichen
\usepackage[
	pdftex,
	pdfa,
	colorlinks=true,
	citecolor= magenta,
	linkcolor=blue,
	urlcolor=cyan
	]{hyperref} % Hypertextstrukturen ermoeglichen
\usepackage{xurl} % url mit Umbrücken, am besten nach biblatex laden
\usepackage[babel]{csquotes} % "schoene" Anfuehrungszeichen
\usepackage{xspace}
\usepackage{amsmath}
\usepackage{amssymb}
\usepackage{mathtools}
\usepackage{xfrac}
\usepackage{siunitx}
\usepackage[most]{tcolorbox}
\usepackage{etoolbox}
\usepackage{dsfont}
\usepackage{caption}
\usepackage{subcaption}
\usepackage{pgfplots}  % Plots mit tikz/pgf
\pgfplotsset{compat=1.14}
\usepackage{tikz} % draw graphics, diagrams and boxes
\usetikzlibrary{arrows.meta, decorations.text, decorations.pathmorphing, positioning, shapes.misc}
\usepackage{listings} % verbatim and code style
\usepackage{fontawesome5}
\usepackage{placeins}  % \FloatBarrier
\addto\extrasenglish{

}

\AtBeginEnvironment{definition}{}

\let\proglang=\textsf
\let\code=\texttt
\let\variable=\texttt

\newcommand*{\eg}{e.g.,\@\xspace}
\newcommand*{\ie}{i.e.\@\xspace}
\newcommand*{\cf}{cf.\@\xspace}
\newcommand*{\etc}{etc.\@\xspace}
\newcommand*{\wrt}{w.r.t.\@\xspace}
\newcommand*{\iid}{i.i.d.\@\xspace}

\DeclareMathOperator{\E}{\mathbb{E}}  % alternative: \mathbf{E}
\DeclareMathOperator{\Var}{Var}
\DeclareMathOperator{\Cov}{Cov}
\DeclareMathOperator{\CV}{CV}

\DeclareMathOperator*{\argmin}{arg\,min}
\DeclarePairedDelimiter\abs{\lvert}{\rvert}%
\DeclareMathOperator{\sign}{sign}

\newcommand{\dint}{\,\mathrm{d}}

\newcommand{\F}{\mathcal{F}}
\newcommand{\R}{\mathbb{R}}

\newcommand{\one}{\mathds{1}}

\newcommand{\ph}{\varphi}

\renewcommand{\P}{\mathbb P}

\newcommand{\M}{\mathcal M}
\newcommand{\X}{\mathcal X}
\newcommand{\Y}{\mathcal Y}

\newcommand{\bx}{\boldsymbol{x}}
\newcommand{\bX}{\boldsymbol{X}}

\newcommand{\Rtwo}{\mathrm{R}^2}
\newcommand{\Rstar}{\mathrm{R}^\star}

\def\realnumbers{\mathbb{R}}

\newtheorem{definition}{Definition}

\definecolor{coltheo}{RGB}{0, 2, 146}  % #000292
\definecolor{colinsight}{RGB}{254, 209, 72}  % #FED148
\definecolor{colexam}{rgb}{1.0, 0.25, 0.25} % Coral red #ff4040

\DeclareNewSectionCommand[
	style=section,
    level=2,  % or \subsectionnumdepth
    indent=0pt,
	beforeskip=-3.25ex plus -1ex minus -.2ex,
	afterskip=1.5ex plus .2ex,
	tocstyle=subsection,
  	tocindent=1.5em,
  	tocnumwidth=2.3em,
	font=\usekomafont{subsection},
	counterwithin=section
]{theorysubsection}
\DeclareNewSectionCommand[
	style=section,
    level=2,  % or \subsectionnumdepth
    indent=0pt,
	beforeskip=-3.25ex plus -1ex minus -.2ex,
	afterskip=1.5ex plus .2ex,
	tocstyle=subsection,
  	tocindent=1.5em,
  	tocnumwidth=2.3em,
	font=\usekomafont{subsection},
	counterwithin=section
]{insightsubsection}
\DeclareNewSectionCommand[
	style=section,
    level=2,  % or \subsectionnumdepth
    indent=0pt,
	beforeskip=-3.25ex plus -1ex minus -.2ex,
	afterskip=1.5ex plus .2ex,
	tocstyle=subsection,
  	tocindent=1.5em,
  	tocnumwidth=2.3em,
	font=\usekomafont{subsection},
	counterwithin=section
]{examplesubsection}

\makeatletter
\renewcommand{\sectionlinesformat}[4]{%
	\Ifstr{#1}{theorysubsection}{%
		\@hangfrom{\hskip #2\parbox[c][1.1cm]{1cm}{\textcolor{coltheo}{\Huge\faBook}}#3}{#4}% or \faChalkboardTeacher
	}{%
		\Ifstr{#1}{insightsubsection}{%
			\@hangfrom{\hskip #2\parbox[c][1.1cm]{1cm}{\textcolor{colinsight}{\Huge\faInfoCircle}}#3}{#4}%
		}{%
			\Ifstr{#1}{examplesubsection}{%
				\@hangfrom{\hskip #2\parbox[c][1.1cm]{1cm}{\textcolor{colexam}{\Huge\faFlask}}#3}{#4}%
			}{%
				\@hangfrom{\hskip #2#3}{#4}%
			}%
		}%
	}%
}
\makeatother

\newcommand{\TheorySec}{
	\setcounter{theorysubsection}{\value{subsection}}%
	\refstepcounter{subsection}%
	\theorysubsection[nonumber=false, tocentry=Theory, reference=Theory]{Theory}
}

\newcommand{\InsightsSec}{
	\setcounter{insightsubsection}{\value{subsection}}%
	\refstepcounter{subsection}%
	\insightsubsection[nonumber=false, tocentry=Practicalities, reference=Practicalities]{Practicalities}
}

\newcommand{\ExampleSec}[1]{
	\setcounter{examplesubsection}{\value{subsection}}%
	\refstepcounter{subsection}%
	\examplesubsection[nonumber=false, tocentry=Example, reference=Example]{Example: #1}
}

\addto\extrasenglish{

}
\newtcolorbox[use counter=example_env]{emph_box}{%
	colback=black!5!white,colframe=black!75!white
}

\NewDocumentEnvironment{example_env}{ O{} }
{
\newtcolorbox[use counter=example_env]{example_box}{%
	empty,  % Empty previously set parameters
	title={Example: #1},  % use \thetcbcounter to access the example_env counter text
	attach boxed title to top left,
	minipage boxed title,
	boxed title style={empty, size=minimal, toprule=0pt, top=4pt, left=3mm, overlay={}},
	coltitle=colexam,
	fonttitle=\bfseries,
	before=\par\medskip\noindent,
	parbox=true %false,
	boxsep=0pt,
	left=3mm,
	right=0mm,
	top=2pt,
	breakable,
	overlay unbroken={\draw[colexam,line width=2.0pt] ([xshift=-0pt]title.north west) -- ([xshift=-0pt]frame.south west); },
	overlay first={\draw[colexam,line width=2.0pt] ([xshift=-0pt]title.north west) -- ([xshift=-0pt]frame.south west); },
	overlay middle={\draw[colexam,line width=2.0pt] ([xshift=-0pt]frame.north west) -- ([xshift=-0pt]frame.south west); },
	overlay last={\draw[colexam,line width=2.0pt] ([xshift=-0pt]frame.north west) -- ([xshift=-0pt]frame.south west); },%
	}
\begin{example_box}}
{\end{example_box}\endlist}

\newtcbtheorem{summarybox}{\bfseries Summary}{
	enhanced,
}{sb}

\newcommand*\samethanks[1][\value{footnote}]{\footnotemark[#1]}

\begin{document}
% ----------------------------------------------------------------------------
% Titel (erst nach \begin{document}, damit babel bereits voll aktiv ist:
%\titlehead{Kopf über dem Titel mit Le(h/e)rstuhl u.\,ä.}% optional
%\subject{Art des Dokuments}% optional
\title{Model Comparison and Calibration Assessment}% obligatorisch
\subtitle{User Guide for Consistent Scoring Functions in Machine Learning and Actuarial Practice}
%\subtitle{Untertitel}% optional
\author{Tobias Fissler\thanks{Institute for Statistics and Mathematics,
Vienna University of Economics and Business (WU), Welthandelsplatz 1, 1020 Vienna, Austria} \\ {\small tobias.fissler@wu.ac.at}
	\and Christian Lorentzen\thanks{la Mobili\`ere, Bern, Switzerland} \\ {\small christian.lorentzen@mobiliar.ch}
	\and Michael Mayer\samethanks \\ {\small michael.mayer@mobiliar.ch}}% obligatorisch
\date{Prepared for:
\\Working Group \enquote{Data Science}
\\Swiss Association of Actuaries SAV
\\Version of~\today} % sinnvoll
%\publishers{Platz für Betreuer o.\,ä.}% optional
\maketitle% verwendet die zuvor gemachte Angaben zur Gestaltung eines Titels

% ----------------------------------------------------------------------------
\begin{abstract}
One of the main tasks of actuaries and data scientists is to build good predictive models for certain phenomena such as the claim size or the number of claims in insurance.
These models ideally exploit given feature information to enhance the accuracy of prediction.
This user guide revisits and clarifies statistical techniques to assess the calibration or adequacy of a model on the one hand, and to compare and rank different models on the other hand.
In doing so, it emphasises the importance of specifying the prediction target functional at hand a priori (e.g.\@\xspace the mean or a quantile) and of choosing the scoring function in model comparison in line with this target functional.
Guidance for the practical choice of the scoring function is provided.
Striving to bridge the gap between science and daily practice in application, it focuses mainly on the pedagogical presentation of existing results and of best practice.
The results are accompanied and illustrated by two real data case studies on workers' compensation and customer churn.
\newline
\end{abstract}
\textbf{\textit{keywords:}}
actuarial science,
backtesting,
calibration,
classification,
consistency, 
%cross-validation,
data science,
identification functions,
machine learning,
model comparison,
predictive performance,
propriety,
scoring functions,
scoring rules,
supervised learning
% ----------------------------------------------------------------------------
% Inhaltsverzeichnis:
\newpage
\setcounter{tocdepth}{\subsectiontocdepth}
\tableofcontents
% ----------------------------------------------------------------------------
% Gliederung und Text:

%\listoftodos

\newpage
\section{Introduction}

This study has been carried out for the working group \enquote{Data Science} of the Swiss Association of Actuaries SAV, see
\begin{center}
	\url{https://www.actuarialdatascience.org}
\end{center}
The purpose of this user guide is to provide an overview of point forecast evaluation, theory and examples hand in hand.
For better readability and distinction of theory and example sections, we mark them as follows:

\parbox[c][1.1cm]{1cm}{\textcolor{coltheo}{\Huge\faBook}} {\usekomafont{disposition}Theory}\hspace{1cm}
\parbox[c][1.1cm]{1cm}{\textcolor{colinsight}{\Huge\faInfoCircle}} {\usekomafont{disposition}Practicalities}\hspace{1cm}
\parbox[c][1.1cm]{1cm}{\textcolor{colexam}{\Huge\faFlask}} {\usekomafont{disposition}Example}

\minisec{}
Nowadays, validating and comparing the predictive performance of statistical models has become ubiquitous by means like cross-validation, testing on a hold-out data set and machine learning (ML) competitions.
In the light of these developments, this article strives to provide the methodology to answer two key questions:
\begin{emph_box}
\begin{enumerate}
\item Given a model, does it produce calibrated predictions? 
\item Given two models, which model produces more accurate predictions?
%how can their predictive performances be assessed and compared?
\end{enumerate}
\end{emph_box}
In finance, these two tasks are customarily subsumed under the umbrella term backtesting. 
Intuitively, calibration means that the predictions produced by a model are in line with the observations of the response variable.
It is a generalisation of unbiasedness, sometimes referred to as balance property in actuarial science.
Calibration can be checked by means of identification functions, also known as moment functions.
For the latter quest, predictive accuracy is commonly assessed in terms of loss or scoring functions, sometimes also called metrics.
Roughly speaking, they measure the distance between a prediction and the observation of the quantity of interest.
Crucially, for both tasks, the evaluation methodology should be \textcquote{gneiting2014}{designed such that truth telling \textelp{} is an optimal strategy in expectation}.
But what does \emph{truth} amount to in this setting?
In the most general and informative form, it is the true (conditional) probability distribution of the response (given the information contained in the features), called probabilistic prediction \cite{gneiting_2011}.
Alternatively, when the model outputs point predictions, it is a pre-specified summary measure, a property, or a \emph{functional} of the true (conditional) probability distribution of the response, such as the (conditional) mean or a (conditional) quantile.
For the situation of point predictions, the identification and scoring functions used must be chosen to fit to the target functional, \ie they need to be strictly consistent.
Interestingly, strictly consistent scoring functions then simultaneously assess calibration and discrimination ability of the model predictions.

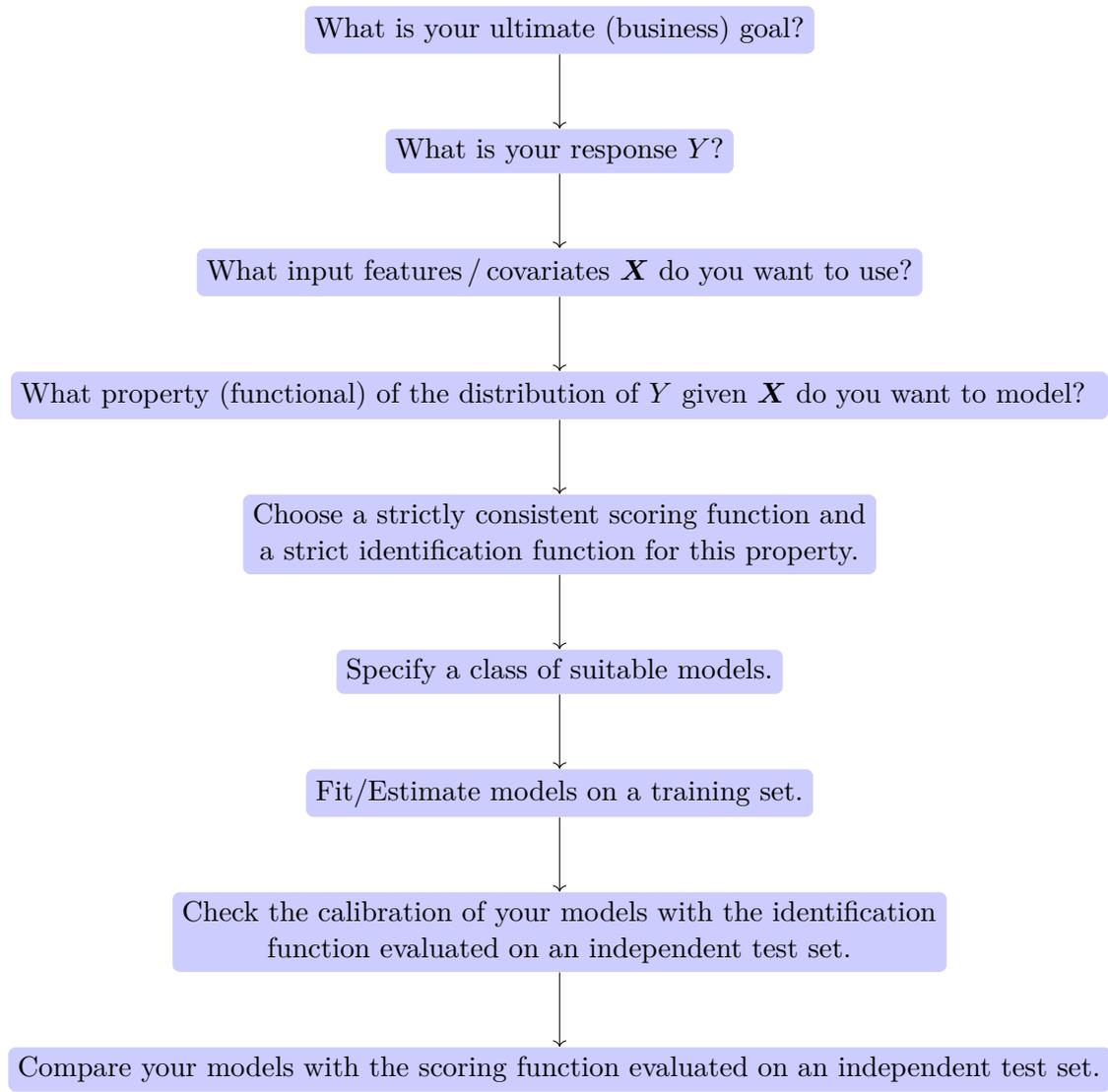
\begin{figure}[htbp]
\centering
\begin{tikzpicture}[every node/.style={fill=blue!20, align=center, rounded corners=0.1cm}]
	\node[rectangle] (goal) {What is your ultimate (business) goal?};
	\node[rectangle, below=of goal] (response) {What is your response $Y$?};
	\node[rectangle, below=of response] (explanatory variables) {What input features\,/\,covariates $\bX$ do you want to use?};
	\node[rectangle, below=of explanatory variables] (functional) {What property (functional) of the distribution of $Y$ given $\bX$ do you want to model?
	};
	\node[rectangle, below=of functional] (score) {Choose a strictly consistent scoring function and \\
	a strict identification function for this property.};
	\node[rectangle, below=of score] (model) {Specify a class of suitable models.};
	\node[rectangle, below=of model] (fit) {Fit/Estimate models on a training set.};
	\node[rectangle, below=of fit] (cali) {Check the calibration of your models with the identification \\
	function evaluated on an independent test set.};
	\node[rectangle, below=of cali](comparison) {Compare your models with the scoring function evaluated on an independent test set.};
	\draw[->] (goal) edge (response) (response) edge (explanatory variables) (explanatory variables) edge (functional) (functional) edge (score) (score) edge (model) (model) edge (fit) (fit) edge (cali) (cali) edge (comparison);
\end{tikzpicture}
\caption{Modeling decision graph}
\label{fig:decision}
\end{figure}

In this user guide, we summarise the theoretical background and give illustrative examples.
As a shortcut, just follow \autoref{fig:decision} and \nameref{par:theory short form}.

\newpage

\paragraph{Theory Short Form} \label{par:theory short form}
\phantom{.}\nopagebreak
\begin{summarybox*}{Theory}
\begin{itemize}
	\item Mind your statistical \textbf{assumption}:
	Response $Y$ given feature information $\bX$ is a random variable. 
	Thus, $Y$ cannot be described as a deterministic function of $\bX$.
	\item Decide if and how to summarise the uncertainty of $Y$ given $\bX$, depending on your (business) goal. 
	\begin{itemize}
		\item Not summarising leads to probabilistic predictions where the ideal prediction is the conditional distribution of $Y$ given $\bX$, $F_{Y|\bX}$.
		\item Summarising with a statistical functional $T$ leads to point predictions.
		The ideal prediction is given in terms of the conditional functional $T(Y|\bX)$, \eg the conditional expectation $\E[Y|\bX]$.
	\end{itemize}
	\item Check whether $T$ is identifiable and elicitable.
	\item Use a \textbf{strict identification function} for $T$ to assess the calibration of a model.
	\item Choose a \textbf{strictly consistent} scoring function for $T$ in order to compare predictions of different models on a test data set that is independent of the training data.
\end{itemize}
\end{summarybox*}

\paragraph{Outline}
The present user guide starts with a short overview of supervised learning in \autoref{sec:Supervised Learning} at the end of which a data set for a regression modelling example is introduced.
\hyperref[sec:statistical learning theory]{Section~\ref*{sec:statistical learning theory}} continues with the core of supervised learning: loss functions and the statistical risk.
It introduces the concept of overfitting and the necessity of a train--test split for model comparison.
In the example part, several models are trained, among them generalised linear models and gradient boosted trees.
How to assess model calibration is then presented in \autoref{sec:Calibration and Identification Functions}.
It defines identification functions and different notions of calibration which are then assessed and visualised in detail.
The following \autoref{sec:model comparison} lays out the theory of scoring functions, which is an alternative name for loss functions.\footnote{\label{footnote:loss}%
Conventions in the literature are different and sometimes loss functions are required to be non-negative. We do not impose this condition. Our convention is to speak of loss functions in the context of learning and of scoring functions in the context of evaluation.}
A central property of scoring functions is (strict) consistency, and general forms of consistent scoring functions for the most common target functionals are provided.
Furthermore, it is shown that scoring functions simultaneously assess calibration and potential discriminative power of a given model.
After practical considerations on how to choose a particular scoring function, the example part illustrates model comparison and concludes which model performs best on the given data set.
\hyperref[sec:probabilistic binary classification]{Section~\ref*{sec:probabilistic binary classification}} sheds light on the peculiarities of probabilistic binary classification and demonstrates them on a classification data set.
Finally, the user guide concludes with \autoref{sec:conclusion}.

\paragraph{Configuration}
All \proglang{R} code was run on the following system.
\begin{itemize}
	%\item OS: Microsoft Windows 10 Pro
	\item Processor: Intel(R) Core(TM) i7-8650U CPU @ 1.90GHz, 2112 Mhz, 4 Core(s)
	\item \proglang{R} version: 4.0.4
\end{itemize}
%Specific package versions are indicated in the code.
The complete R code can be found at \url{https://github.com/JSchelldorfer/ActuarialDataScience}.
Note: Since version 3.6.0, \proglang{R} uses a different random number generator.
Results are not reproducible under older versions.

\section{Supervised Learning} \label{sec:Supervised Learning}

\TheorySec \label{subsec:Supervised Learning.Theory}

According to Wikipedia\footnote{%
	\url{https://en.wikipedia.org/wiki/Supervised_learning} as of 06.01.2021.
}
\textcquote{RusselNorvig2010}{supervised learning is the machine learning task of learning a function that maps an input to an output based on example input-output pairs.}
Using standard notation, we call the input features, also called covariates, regressors or explanatory variables, $\bX$, taking values in some possibly high dimensional feature space $\X$ such as $\R^K$,
and the output $Y$, which is the response or target variable, and which takes values in some space $\Y$, which we assume to be a subset of $\R$.
Observations of input-output pairs are denoted by $(\bx_i, y_i)$, $i=1,\ldots, n$.
The presence of the observable outputs is the reason for the term \emph{supervised}.
In statistical learning theory, we consider both $\bX$ and $Y$ to be \emph{random variables} with joint probability distribution $F_{\bX,Y}$.
To ease the exposition, we dispense with a discussion of a time series framework and solely focus on cross-sectional data.
That is, we assume that each sample of data is a random sample, meaning the data at hand $(\bx_i, y_i)$, $i=1,\ldots, n$, is independent and identically distributed (\iid).%
\footnote{%
	It is also possible to allow for serial dependence and some non-stationarity in the data.
	However, this would dilute the main message of the paper and distract from our main goals.
}

The first key point we would like to stress is: 
It is in general illusive to hope that the input features $\bX$ can fully explain the behaviour of $Y$ in that $Y$ is a deterministic function $g$ of the features, $Y = g(\bX)$.
There is a remaining degree of uncertainty, which can be expressed in terms of the conditional distribution of $Y$, given $\bX$, denoted by $F_{Y|\bX}$.
The most informative prediction approach is probabilistic, aiming at specifying the full conditional distribution $F_{Y|\bX}$.
Often, one is content with point predictions,\footnote{In this tutorial, we will use the terms \enquote{prediction} and \enquote{forecast} interchangeably, possibly ignoring the temporal connotation of the latter.} modelling only a certain property or summary measure of the conditional distribution.
Strictly speaking, such a summary measure is a statistical \emph{functional}, mapping a distribution to a real number, such as the mean or a quantile.
Then the ideal point prediction takes the form $T(F_{Y|\bX})$, which we will denote by $T(Y|\bX)$ for convenience.
If $T$ is the mean functional, we obtain $T(Y|\bX) = \E[Y|\bX]$, or if $T$ is the $\alpha$-quantile, we get $T(Y|\bX) = q_{\alpha}(Y|\bX) 
%= F_{Y|X}^{-1}(\alpha)
= \inf\{t\in\R\mid F_{Y|\bX}(t)\ge \alpha\}
$.
It is also possible to combine different functionals to a vector, \eg when interested in two quantiles at different levels or in the mean and variance of $Y$ given $X$.
For the sake of simplicity, we will mostly stick to one dimensional functionals in this article, though.

The ideal goal of supervised learning is then to find a model $m\colon \X \to \R$ which approximates the actual underlying regression function $\X \ni \bx \mapsto T(Y|\bX=\bx)$.
Since the search space of \emph{all} (measurable) functions $\X \to \R$ is clearly not tractable, one needs to come up with a suitable model class $\M$ of such functions.
For our purpose, we will require that $\M$ contains models of different complexity (also called model capacity) and always contains the trivial model, \ie constant model (also called null model or intercept model).
While the model complexity can theoretically be defined by the Vapnik--Chervonenkis dimension or the Rademacher complexity, \cf \cite{ShalevShwartz2014, vonluxburg2008statistical, 10.1007/3-540-44581-1_15} and references therein, we give some vivid examples:
\begin{itemize}
\item Most classically, for the class of linear models, complexity can be measured by the number of estimable parameters, also known as degrees of freedom (df).
This amounts to including and excluding features as well as adding interaction terms and quadratic and higher order polynomial terms such as splines.
\item For penalised linear models (with fixed input features), such as ridge or lasso regression, the penalty parameter is a---reciprocal---measure of complexity.
\item Non-parametric approaches often impose smoothness or shape constraints on the models, with the most prominent case of isotonic models.%
\footnote{%
	There is some (partial) order $\preceq$ on $\X$ and $x_1 \preceq x_2$ implies that $m(x_1)\le m(x_2)$.
}
Then the degree of smoothness, for example, can be taken as a measure of complexity.
\item For decision trees, it can be the number of leaves or the maximal tree depth.
\item For ensemble models like gradient boosted trees, it can be a combination of the single tree complexity and the number of fitted trees.
\item For neural nets, where the model class $\M$ is implicitly given by the architecture of the net, it can be the combination of the number of weights, early stopping and weight penalisation.
\end{itemize}
Using a training data sample $(\bx_i, y_i)$, $i=1, \ldots, n$, one then fits (or trains or estimates) a model $\widehat m\in\M$.
The fitted model commonly has two purposes: 
On the one hand, it should be interpretable, describing the connection of $Y$ and $\bX$ in terms of $T$ (which is often easier if $\M$ is small, \eg for linear models).
On the other hand, $\widehat m$ should produce predictions which should be as accurate as possible.
That is, when using a new feature point $\bx_{\text{new}}$ (not necessarily contained in the training sample), $\widehat m(\bx_{\text{new}})$ should be close to the ideal $T(Y|\bX = \bx_{\text{new}})$.

\ExampleSec{Mean Regression for Workers' Compensation.}

Throughout this user guide, we illustrate the theoretical results on the Workers' Compensation data set,\footnote{%
	\url{https://www.openml.org/d/42876}
} which consists of $n = \num{100000}$ rows, each representing a single insurance claim caused by injury or death from accidents or occupational disease while a worker is on the job.
Due to possible data inconsistencies, we filter out all rows with $\text{\variable{WeeklyPay}} < 200$ and $\text{\variable{HoursWorkedPerWeek}} < 20$.
This leaves us with $n = \num{82017}$ rows, see Listing~\ref{ls:common data preprocessing regression}.
\begin{lstlisting}[caption=Common data preprocessing for regression example, label=ls:common data preprocessing regression]
library(tidyverse)
library(lubridate)
library(OpenML)


df_origin <- getOMLDataSet(data.id = 42876L)
df <- tibble(df_origin$data)

df <- df %>% 
  filter(WeeklyPay >= 200,
         HoursWorkedPerWeek >= 20) %>% 
  mutate(
    DateTimeOfAccident = ymd_hms(DateTimeOfAccident),
    DateOfAccident = as_date(DateTimeOfAccident),
    DateReported = ymd(DateReported),
    LogDelay = log1p(as.numeric(DateReported - DateOfAccident)),
    HourOfAccident = hour(DateTimeOfAccident),
    WeekDayOfAccident = factor(wday(DateOfAccident, week_start = 1)), 
    LogWeeklyPay = log1p(WeeklyPay),
    LogInitial = log(InitialCaseEstimate),
    DependentChildren = pmin(4, DependentChildren),
    HoursWorkedPerWeek = pmin(60, HoursWorkedPerWeek),) %>% 
  mutate_at(c("Gender", "MaritalStatus", "PartTimeFullTime"), as.factor) %>% 
  rename(HoursPerWeek = HoursWorkedPerWeek)

x_continuous <- c("Age", "LogWeeklyPay", "LogInitial", "HourOfAccident",
                 "HoursPerWeek", "LogDelay")
x_discrete <- c("Gender", "MaritalStatus", "PartTimeFullTime",
                "DependentChildren",
                "DaysWorkedPerWeek", "WeekDayOfAccident")
x_vars <- c(x_continuous, x_discrete)
y_var <- "UltimateIncurredClaimCost"
\end{lstlisting}

The response variable is $Y = \allowbreak\text{\variable{UltimateIncurredClaimCost}}$ with explanatory features $\bX$ represented by the remaining columns listed in \variable{x\_vars}.
As a summary measure $T$, we choose to model the conditional expectation of $Y$ given $\bX$.
It is the most common target functional in machine learning. 
In actuarial science, it informs about adequate pricing of policies.
Moreover, the tower property of the conditional expectation%
\footnote{%
		That is, $\E\left[\E[Y|X]\right] = \E[Y]$.
	}
facilitates to compute an estimate of the unconditional expectation of $Y$ as 
$\frac{1}{n}\sum_{i=1}^n m(\bX_i)$
for claims (or policies) $i=1, \ldots, n$, \ie the expected value for a whole portfolio of claims (or policies).

Listing~\ref{ls:data} shows the first ten rows of the data.

\begin{lstlisting}[caption=Output of command \code{print(df)}, label=ls:data]
> df %>%
+   select(y_var, x_vars) %>%
+   print(n = 10, width = 80)
# A tibble: 82,017 x 13
   UltimateIncurredCl... Age LogWeeklyPay LogInitial HourOfAccident HoursPerWeek
                 <dbl> <dbl>        <dbl>      <dbl>          <int>        <dbl>
 1                102.    45         6.22       9.16              9           38
 2               1451     40         5.65       8.01             15           38
 3                320.    50         6.25       6.91              7           38
 4                108     19         5.30       4.70             14           38
 5               7111.    19         6.64       9.18             14           40
 6               8379.    21         5.30       9.16             12           37
 7              25338.    50         6.79      11.2               0           38
 8                354     25         5.83       7.70              8           35
 9               2269.    20         5.51       8.01             11           40
10             129367.    43         7.55       9.16             16           40
# ... with 82,007 more rows, and 7 more variables: LogDelay <dbl>, Gender <fct>,
#   MaritalStatus <fct>, PartTimeFullTime <fct>, DependentChildren <dbl>,
#   DaysWorkedPerWeek <dbl>, WeekDayOfAccident <fct>
\end{lstlisting}

\begin{figure}%[htbp]
	\centering
	\includegraphics[width=.90\textwidth]{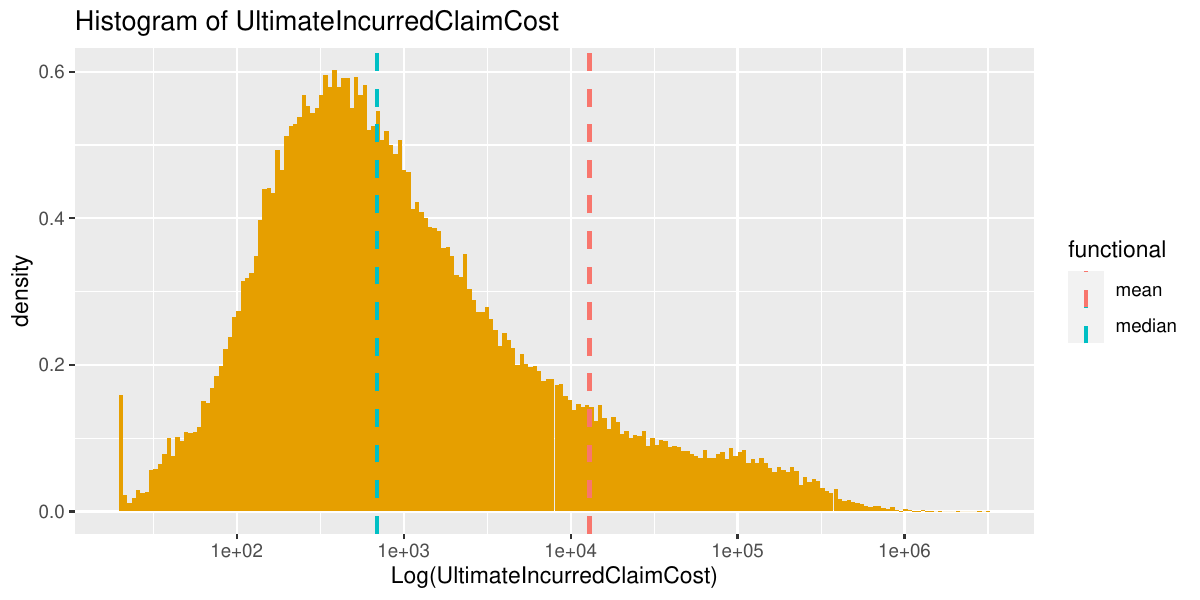}
	\caption{Histogram of response \variable{UltimateIncurredClaimCost} on a logarithmic scale.}
	\label{fig:target histogram}
\end{figure}
In \autoref{fig:target histogram}, the empirical, marginal distribution of $Y$ is visualised.
As the large difference between mean and median indicates, the distribution seems to be very asymmetric, right-skewed, even heavy tailed.
(Mind the logarithmic scale of the $x$-axis.)
For further exploratory data analysis, we refer to \autoref{subsec:EDA regression}.

\section{Statistical Learning Theory} \label{sec:statistical learning theory}

\TheorySec

To provide a complete picture, we shortly outline the basics of statistical learning theory.
This will explain the difference between model estimation on the one hand and model validation and selection on the other hand.
We start by introducing a loss function $L(z, y) \in\R$, 
$z,y\in\R$,
that measures the accuracy of predictions $z=m(\bx)$ from a model $m\in\M$ for observations of the response $y$---with the convention \emph{the smaller the better}.
The leading example is the squared error, $L(z,y) = (z-y)^2$.
The loss function should be chosen in line with the directive $T$ in that it should be \emph{(strictly) consistent} for $T$; see Definition~\ref{defn:consistent} in \autoref{sec:model comparison}.
For a model $m\in\M$ and a loss $L$, the statistical risk, also known as generalisation error, is defined as
\begin{equation} \label{eq:risk 1}
	R(m) = \E\left[L(m(\bX), Y)\right] \,,
\end{equation}
where the expectation is taken with respect to the joint distribution of $Y$ and $\bX$.
The goal of statistical learning is to find the ideal model $m^\star\in\M$ 
minimising the corresponding statistical risk $R(m)$ \eqref{eq:risk 1}.
Supposed a solution exists, this amounts to the Bayes rule,
\begin{equation*}
	m^\star = \argmin_{m\in\M} R(m)\,.
\end{equation*}
The tower property of the conditional expectation yields the representation
\begin{equation*}
	R(m) = \E\big[ \E\left[ L(m(\bX), Y)|\bX\right] \big]\,.
\end{equation*}
If there is some model $m_0$ minimising the \emph{conditional} statistical risk 
\begin{equation} \label{eq:conditional risk}
	R(m|\bx) = \E\left[ L(m(\bx), Y)|\bX=\bx\right]
\end{equation}
for almost all $\bx\in\X$,\footnote{That means it holds for all $\bx$ in some subset $A\subseteq \X$ such that $\bX\in A$ with probability one.} then $m_0$ clearly minimises the unconditional risk and $m^\star = m_0$.

In the absence of knowing the distribution of $(\bX,Y)$ on the population level, 
we cannot calculate the statistical risk $R(m)$, and therefore generally also fail to determine the ideal model $m^\star$.
We need to resort to an approximation of $R(m)$ on a sample level.
Employing a random sample
$D=\left\{(\bx_i, y_i),\ i=1,\ldots,n \right\}$, 
we can easily compute the empirical risk
\begin{equation*}
	\overline{R}(m; D) %= \overline{L} 
	= \frac{1}{n} \sum_{(\bx_i, y_i) \in D} L(m(\bx_i), y_i) \,.
\end{equation*}
The so called M-estimator\footnote{%
	Where the \enquote{M} is for \emph{minimisation} and is due to Huber \cite{Huber1964}; see also \cite{HuberRonchetti2009} for a good textbook.
}
$\widehat m$ is defined as the empirical risk minimiser
\begin{equation} \label{eq:M-estimator}
	\widehat m = \widehat m({\,\cdot\,}; D_{\text{train}})  = \argmin_{m\in\M} \overline{R}(m; D_{\text{train}})
\end{equation}
over a training sample $D_{\text{train}}$.
Since estimating $\widehat{m}$ is based on the empirical risk, which is only an approximation of the statistical risk, it suffers from sampling uncertainty and is thus prone to estimation error.
In particular, $\widehat{m} = \widehat{m}(\cdot; D_{\text{train}})$ inherently depends on the particular training sample $D_{\text{train}}$ at hand.
If another training sample $D'_{\text{train}}$ had been chosen, then generally $\widehat{m}(\cdot; D_{\text{train}}) \neq \widehat{m}(\cdot; D'_{\text{train}})$.
This sampling variability leads us directly to the keyword of overfitting which we will formally introduce in Definition~\ref{defn:overfitting}.
A common practice to reduce this variability is to add an additional penalty term, $\lambda\Omega(m)$, such that \eqref{eq:M-estimator} turns into
\begin{equation*}
	\widehat{m} = \widehat m({\,\cdot\,}; D_{\text{train}}) = \argmin_{m\in\M} \overline{R}(m; D_{\text{train}}) +\lambda\Omega(m) \,.
\end{equation*}
Often, $\Omega \colon \M\to\R$ represents model complexity, see \autoref{subsec:Supervised Learning.Theory}.
For example, it can be a norm of the underlying parameter vector $\theta$, employed in ridge or lasso regression.
The penalisation strength $\lambda\ge0$ is a tuning parameter, or hyperparameter.
It should asymptotically vanish with increasing sample size, which is necessary to obtain a consistent estimate of the statistical risk and thus of the ideal model $m^\star$.
Still, this is only an asymptotic result, and in finite samples, $\widehat m\neq m^\star$ is to be expected.

This fact necessitates a reliable evaluation of the predictive accuracy of $\widehat{m}(\cdot; D_{\text{train}})$, quantifying how well it generalises to unseen data points, $(\bx_{\text{new}}, y_{\text{new}})$.
Furthermore, it calls for a meaningful comparison of different estimators of $m^\star$ which will be the main subject of \autoref{sec:model comparison}.
These estimators could stem from different samples or from different choices of the penalty term $\lambda \Omega(\cdot)$.
The first possibility that comes to mind to estimate the statistical risk of $\widehat{m}(\cdot; D_{\text{train}})$ is to use the same $D_{\text{train}}$ again in the empirical risk, \ie to use the \emph{in-sample} performance, in-sample risk, or training loss, $\overline{R}(\widehat{m}(\cdot; D_{\text{train}}); D_{\text{train}})$.
Having used the training sample twice, it is generally a biased performance measure.
Since any estimator is tailored to fit the training sample involved in the estimation process,%
\footnote{%
	For instance, the OLS estimator is constructed as to minimise the in-sample squared error.
	It approximates the conditional mean $\E[Y|\bX=\bx]$ perfectly on the training sample in so far as the sum of residuals is 0 by construction.
}
the in-sample risk is usually an overly optimistic performance measure that underestimates the statistical risk of $\widehat{m}({\,\cdot\,}; D_{\text{train}})$.\footnote{%
	To be precise, the bias is caused by the dependence between the fitted model and the data used to estimate the risk.
	For squared error and zero-one loss, it can be shown that, in expectation, this bias stems indeed from $\Cov\big[\widehat{m}(\bX_i), Y_i\big]$ and is called expected optimism, see Chapter~7.4 in \cite{hastie01statisticallearning} or Theorem 2.2
	in \cite{kroese2019DSML}.
	Roughly speaking, the amount of \enquote{optimism} of the in-sample risk amounts to too high a sensitivity of the training of the model \wrt changes of the response values in the training sample.
}
This is connected to the notion of overfitting.

\begin{emph_box}
	Instead, the in-sample risk should be discarded in favour of an \emph{out-of-sample} risk, \ie the empirical risk evaluated on an \emph{independent test data set} $D_{\text{test}}$ as $\overline{R}(\widehat{m}(\cdot; D_{\text{train}}); D_{\text{test}})$.
\end{emph_box}

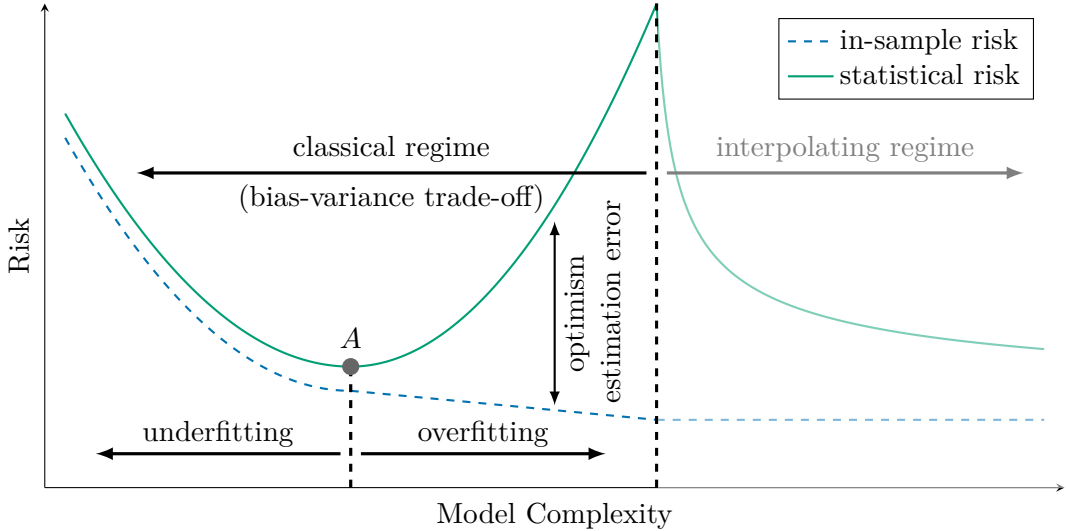
\begin{figure}%[htbp]
	\centering
	\definecolor{c1}{HTML}{0072B2}
	\definecolor{c2}{HTML}{009E73}
	\begin{tikzpicture}
		\tikzstyle{training}=[c1, thick, samples=200, dashed]
		\tikzstyle{testing}=[c2, thick, samples=200]
		\begin{axis}[
			legend pos=north east,
			legend cell align=left,
			axis x line=middle,
			axis y line=middle,
			width=15cm,
			height=8cm,
			xmin=0,       % start the diagram at this x-coordinate
			xmax=100,     % end   the diagram at this x-coordinate
			ymin=0,       % start the diagram at this y-coordinate
			ymax= 1.00,   % end   the diagram at this y-coordinate
			axis background/.style={fill=white},
			xlabel=Model Complexity,
			ylabel=Risk,
			ylabel near ticks,
			xlabel near ticks,
			xtick=\empty,
			ytick=\empty,
			tension=0.08]
			
			% We need the first 2 "addplot" calls to be the first 2 legend entries.
			% in-sample risk
			\addplot[domain=2:30, training] {(x-30)^2/1500 + 0.2};
			% statistical risk
			\addplot[domain=2:30, testing] {(x-30)^2/1500 + 0.25};
			
			% continue in-sample risk
			\addplot[domain=30:60, training] {-0.002*(x-30) + 0.2};
			\addplot[domain=60:98, training] {-0.002*30 + 0.2};
			
			% continue statistical risk
			\addplot[domain=30:60, testing] {(x-30)^2/1200 + 0.25};
			\addplot[domain=60:98, testing] {0.85/sqrt(x-60+1) + 0.15};
			
			\draw[dashed, very thick] (axis cs:30, 0.0) -- (axis cs:30, 0.25);
			\draw[-{latex[scale=3.0]}, very thick] (axis cs:31, 0.07) -- (axis cs:55, 0.07) node[midway, sloped, above=-0.5mm] {overfitting};
			\draw[-{latex[scale=3.0]}, very thick] (axis cs:29, 0.07) -- (axis cs:5, 0.07) node[midway, sloped, above=-0.5mm] {underfitting};
			
			\draw[{latex[scale=3.0]}-{latex[scale=3.0]}, thick] (axis cs:50, 0.17) -- (axis cs:50, 0.55) node[midway, sloped, below, align=center]{optimism\\ \phantom{m}estimation error};
			
			\draw[dashed, very thick] (axis cs:60, 0.0) -- (axis cs:60, 1.0);
			
			%\node[align=center] at (axis cs:80, 0.7) [rectangle, rounded corners, draw=black!50] {interpolating regime};
			\draw[-{latex[scale=3.0]}, very thick] (axis cs:61, 0.65) -- (axis cs:96, 0.65) node[midway, above] {interpolating regime};
			\draw[-{latex[scale=3.0]}, very thick] (axis cs:59, 0.65) -- (axis cs:9, 0.65) node[midway, above]{classical regime} node[midway, below]{(bias-variance trade-off)};
			
			\draw [fill=black!60, draw=black!60] (axis cs:30, 0.25) circle [radius=3pt] node [above=3pt]{$A$};
			%\draw [fill=black!60, draw=black!60] (axis cs:90, 0.30266) circle [radius=3pt] node [above=3pt] {$B$};
			
			% shade the interpolating regime on the right side
			\fill [white, opacity=0.5] (axis cs:60.1,0) rectangle (axis cs:100,1);

			\addlegendentry{in-sample risk}
			\addlegendentry{statistical risk}
		\end{axis}
	\end{tikzpicture}
	\caption{Illustration of over- and underfitting in the classical regime.
	The estimation error corresponds to the last line in \eqref{eq:decomp stat risk}.
	Both regimes together show the typical double descent slope of the statistical risk.}
	\label{fig:overfitting}
\end{figure}

As the term overfitting is often used heuristically, we make it more precise by defining overfitting in a relative sense, \cf \cite{mitchell1997ML}:
\begin{definition} \label{defn:overfitting}
	Model $m \in \M$ overfits (\wrt model complexity given by $\Omega$) the training data $D_{\textrm{train}}$ if there exists another model $m' \in \M$ with $\Omega(m')<\Omega(m)$ such that $\overline{R}(m; D_{\text{train}}) \le \overline{R}(m'; D_{\text{train}})$, but $R(m) > R(m')$.
\end{definition}
Clearly, this definition of overfitting is only relevant on model classes $\M$ containing models of different complexity.
The left side of \autoref{fig:overfitting}, denoted by \enquote{classical regime}, provides an illustration of over- and underfitting with the typical U-shape of the statistical risk and a monotonically decreasing in-sample risk.
% Copied (and adapted) from
% https://github.com/MartinThoma/LaTeX-examples/tree/master/tikz/2d-epochs-overfitting
Intuitively, overfitting appears as soon as a model starts to learn the training set by heart, indicated by point $A$.
The behaviour of this regime is often explained by the bias-variance trade-off \cite{10.1162/neco.1992.4.1.1, hastie01statisticallearning}:
At low complexity, the model is not able to capture the data structure, leading to a high bias.
With larger complexity, the bias is reduced, but the variance, \ie the estimation error, grows. 

The interpolating regime on the right side of \autoref{fig:overfitting} is an area of recent research, initiated by investigating the success of deep neural nets---models with an enormously huge amount of parameters---\cite{neyshabur2015search, zhang2017understanding}.
The point where a model starts to interpolate the data, \ie $m(\bx_i) = y_i$ for all training data $(\bx_i, y_i) \in D_{\textrm{train}}$, has a very high statistical risk.\footnote{%
	Here, we implicitly assume that the target functional $T$ satisfies $T(\delta_y) = y$ for any $y\in\Y$, where $\delta_y$ is a point mass in $y$.
}
But as the model gets even more overparametrised, the statistical risk decreases again, while the in-sample risk remains constant (and minimal).
Therefore, the terms interpolation and overparameterisation are more appropriate in this regime than the notion of overfitting.
The overall shape of the statistical risk has been called \enquote{double descent} and has been confirmed for several model classes \cite{Belkin15849}.
Analytical results are available for linear models \cite{Belkin2020, hastie2020surprises, Bartlett30063}.
Whether there is a point in the interpolating regime with a smaller statistical risk than the local minimum in the classical regime denoted by $A$ depends on the model class as well as the distribution of the data, see \cite{dar2021farewell} for a review.

Another possible way for dealing with the bias of the in-sample risk besides making use of an independent test data set is to use alternative in-sample measures like Akaike's information criterion (AIC) or the Bayesian information criterion (BIC), see \cite{hastie01statisticallearning} and references therein.
They measure the in-sample risk including a penalty term accounting for model complexity.

The following decomposition of the statistical risk of a model $m\in \M$\footnote{%
	Due to Daniel Hsu's slides \url{https://www.cs.columbia.edu/~djhsu/tripodsbootcamp/overview.slides.pdf}.
}
illustrates the constituent parts of the learning procedure:
\begin{alignat}{3} \label{eq:decomp stat risk}
	R(m) &= \inf_{g: \X \to \Y} R(g) && \qquad &&\text{inherent unpredictability}\\ \nonumber
	&+ \inf_{f \in \M} R(f) &&- \inf_{g: \X \to \Y} R(g) \qquad&&\text{approximation error}\\ \nonumber
	&+ \inf_{f \in \M} \overline{R}(f; D) &&- \inf_{f \in \M} R(f) \qquad&&\text{estimation error I}\\ \nonumber
	&+ \overline{R}(m; D) &&- \inf_{f \in \M} \overline{R}(f; D) \qquad\qquad&&\text{optimisation error}\\ \nonumber
	&+ R(m) &&- \overline{R}(m; D) \qquad&&\text{estimation error II}
\end{alignat}
Note that while the approximation and optimisation errors are always non-negative, the two estimation errors can have any sign.\footnote{%
	Figure~7.2 in \cite{hastie01statisticallearning} illustrates a similar decomposition.
}
Equation \eqref{eq:score decomposition} in \autoref{sec:model comparison} provides an alternative decomposition of statistical risk (there named expected score) with an emphasis on evaluation rather than learning.

\InsightsSec

\subsubsection{Train--Validation--Test split}
In actuarial practice and machine learning, one often enjoys the situation of having a large amount of data,\footnote{%
	Interestingly, the amount of data available to actuaries varies a lot depending on insurance sector and country, from (almost) no data for newly invented insurance coverages to usually very good amounts of data for motor insurance.
}
both for fitting and for model evaluation and comparison.
Then, a crucial rule is to divide the data set into mutually exclusive subsets.
Terminology and usage patterns for those data samples vary in the literature.
We give the following definition, \cf \cite[Chapter~7.2]{hastie01statisticallearning}, \cite[Chapter~7.2.3]{Wuthrich2021StatisticalFoundations}, \cite{Raschka2018}:
\begin{itemize}
	\item Training set for model fitting, typically the largest set.
	\item Validation set for model comparison and model selection.\newline
	Typically, this set is used to tune a model of a given model class while building (fitting) models on the training set.
	Examples are variable selection and specification of terms for linear models, finding optimal architecture and early stopping for neural nets, hyperparameter tuning of boosted trees.
	This way, the validation set is heavily used and therefore does not provide an unbiased performance estimate anymore.
	The result is a \enquote{final} model for the given model class that is often refit on the joint training and validation sets.
	\item Test set for assessment and comparison of final models.\newline
	Once the model building phase is finished, this set is used to calculate an unbiased estimate of the statistical risk.
	It may be used to pick the best one of the (few) final models.
	\item Application set.\newline
	This is the data the model is used for in production.
	It consists of feature variables only.
	If the observations of the response become known after a certain time delay, it can serve to monitor the performance of the model.
\end{itemize}
In practise, the usage of the terms \enquote{validation set} and \enquote{test set} is often not clearly distinguished.

A golden rule is: \emph{Never ever} look at the test set while still training models.
Any data leakage from the test to the training set might invalidate the results, \ie it will likely render evaluation results on the test set too optimistic.
Also be aware that the more you use the test or validation data set, the less reliable your results become.\footnote{%
	Assume 1000 random forests, each trained with a different random seed.
	Now, we evaluate on the validation set which one performs best and choose it as our final tuned model.
	We will pick the one that, by chance, optimised the validation set.
	If we finally look at the performance of the picked model on an independent test set, we might likely see a worse performance than on the validation set.
	This can be seen as an instance of the survivorship bias.
}
For the validation set, this risk can be mitigated to some degree by methods like cross-validation, see below.

Data points $(\bx_i, y_i) \in D_{\text{test}}$ of a hold-out set, \ie validation and test set, should be independently drawn from the same distribution as the training set, ensuring that the evaluation is representative on the one hand and preventing overly optimistic results on the other hand.\footnote{%
	Beware of distributions of response or feature variables that change over time.
	Take the trend to improved motorway safety in Switzerland as an example.
	A model for insurance claims frequency fitted in the year 2000 and evaluated on test data from 2015--2020 will show that it is not a good fit for today's situation, but might have been a good model at its time.
}
Below, we review and discuss some examples of data with dependencies.

\begin{emph_box}
	A methodological sound train--validation--test split and usage pattern is essential for building good models and for an unbiased assessment of predictive performance.
\end{emph_box}

\subsubsection{Data with dependencies}
Real life data sets often contain dependent rows.
Taking the dependence structure into account is essential to create independent data sets, \cf \cite{Roberts2017CrossvalidationSF}.
\begin{itemize}
	\item Data rows with the same policy number or customer ID are likely correlated.
	In this case, the same number or ID should only go in either training or test set, but not in both.
	This is called grouped sampling and is a form of a blocking strategy \cite{Roberts2017CrossvalidationSF}.
	In practice, this can be a hidden dependency, \eg if the ID is unknown.
	\item If the data at hand is a time series \cite{Hyndman2021ForecastingPA}, the usual assumption is that the dependence decreases with larger time difference.
	Therefore, choosing a split time such that all training samples are older than the test set is a good strategy, called out-of-time \cite{Stein2007BenchmarkingDP} or forward-validation scheme \cite{Schnaubelt2019comparison}.
	The specific validation scheme depends on different aspects, for instance how the models are to be applied.
	For more details, we refer to the overview and references of \cite{Schnaubelt2019comparison}.
\end{itemize}
Failing to account for important dependencies is information leakage between test and train set and will often lead to too optimistic results.

\subsubsection{Data splitting ensuring identical distributions}
The second aspect next to independence of the split data sets is to ensure that they are identically distributed.
Differently distributed features or response variables will typically lead to different estimates of the statistical risk.
This time, however, the direction is unclear: the result on the test set can be clearly better or worse than on the training set.
Ensuring identically distributed sets makes results comparable.
\begin{itemize}
	\item The data at hand might be ordered in some way.
	A simple random shuffled split then helps to create similarly distributed data sets.
	This procedure, however, does not mitigate possible dependence in the data.
	\item To ensure identically distributed response variables, one can employ stratified sampling.
	It can be combined with the previous point and is then called stratified random sampling.
	Stratification is often applied in classification problems, as do we in \autoref{subsec:statistical learning theory.Example}.
	Interestingly, stratification may induce some degree of dependence between the test and the training set, underpinning that there is no silver bullet in statistics.
	
\end{itemize}

\subsubsection{Cross-Validation}
A statistically more robust and less wasteful alternative to the above strategy of using a validation set to evaluate and compare models is $k$-fold cross-validation (CV) \cite{hastie01statisticallearning, ArlotCelisse2010}.
There, the full data set available for training and validation is first split into $k$ partitions or folds, where $k$ is often a value between five and ten.
For each fold, a model is trained on the remaining $k-1$ folds and a validation score is calculated from the hold-out fold.
The $k$ validation scores are then summarised to an average CV score that can be supplemented by its standard error or a confidence interval.\footnote{%
	Note, however, that CV in fact estimates the statistical risk for different models, each trained on a different fold permutation, meaning on different but dependent training sets.
	Thus, the averaged loss value over all folds may be seen as measuring the fit algorithm rather than as estimate of the statistical risk of a single---fitted---model, see \cite{bates2021crossvalidation}.
}
After all decisions are made, the final model is retrained on all folds, \ie, on the full data set (besides a possible test data set).
In practice, different variants to standard $k$-fold CV exist, \eg repeated or nested CV, see references above.

\ExampleSec{Mean Regression for Workers' Compensation.} \label{subsec:statistical learning theory.Example}

Before we start modelling, we need to make a train--test split.
It could be the case that the same person has several claims represented by different rows in the data.
These rows would then likely be correlated.
As we do not have that information in our data, we assume no ordering, and we use a random shuffled split which helps to create similarly distributed sets.
\begin{lstlisting}
set.seed(1234321L)
.in <- sample(nrow(df), round(0.75 * nrow(df)), replace = FALSE)
train <- df[.in, ]
test <- df[-.in, ]
y_train <- train[[y_var]]
y_test <- test[[y_var]]
\end{lstlisting}
For inspection of this point, we list the mean as well as several quantiles of the response variable in \autoref{tab:train_test_statistics}.
\begin{table}
	\centering
	\begin{tabular}{lrrrrrrr} 
		\toprule
		    & Mean & \multicolumn{5}{c}{Quantile}\\
		Set &      & \SI{20}{\percent} & \SI{40}{\percent} & \SI{50}{\percent} & \SI{60}{\percent} & \SI{80}{\percent} & \SI{90}{\percent}\\
		\midrule
		train & \num{13001} & \num{204} & \num{461} & \num{693} & \num{1097} & \num{4596} & \num{18558}\\
		test  & \num{13025} & \num{202} & \num{460} & \num{700} & \num{1136} & \num{4715} & \num{20346}\\
		\bottomrule
	\end{tabular}
	\caption{Sample statistics of \variable{UltimateIncurredClaimCost} on train and test set.}
	\label{tab:train_test_statistics}
\end{table}
For large quantiles, we observe a certain discrepancy, but otherwise this looks acceptable.

Later on, we will evaluate and compare the models using the Gamma deviance, see \autoref{tab:scoring_functions}.
Therefore, we mainly focus on models that minimise this loss function.
We train four different models:
\begin{itemize}
	\item Trivial model: It will always predict the mean \variable{UltimateIncurredClaimCost} of the training set, that is, $m_{\text{trivial}}(\bx) = 13001.33$.
	\item Ordinary least squares (OLS) on $\log(y)$: As the response is very skewed, but always positive, we fit an OLS model on the log transformed response.
	\begin{lstlisting}
form <- reformulate(x_vars, y_var)
fit_ols_log <- lm(update(form, log(UltimateIncurredClaimCost) ~ .),
                  data = train)
corr_fact <- mean(y_train) / mean(exp(fitted(fit_ols_log)))
	\end{lstlisting}
	Note that the backtransformation for predicting on the untransformed response, $\exp\big(m_{\text{OLS}}(\bx)\big)$, introduces a bias.
	We mitigate this by a multiplicative correction factor \lstinline{corr_fact_ols = 6.644697}; see listing above.
	\item Gamma generalised linear model (GLM) with log link:
	Note that, while the log link gives us positive predictions, it is not the canonical link of the Gamma GLM, and therefore introduces a slight bias (see discussion around Equation \eqref{eq:canonical score equation} for this bias and the balance property).
	As for the OLS in log-space, we account for that with the multiplicative correction factor \lstinline{corr_fact_glm = 1.11445}, which is much smaller than the factor for the OLS.
	\begin{lstlisting}
library(glm2)
# glm2 is more stable.
fit_glm_gamma <- glm2(reformulate(x_vars, y_var), data = train, 
                      family = Gamma(link = "log"))
glm_gamma_predict <- function(X) {
	predict(fit_glm_gamma, X, type = "response")
}				  
corr_fact_glm <- mean(y_train) / mean(glm_gamma_predict(train)) 
glm_gamma_corr_predict <- function(X){
	corr_fact_glm * predict(fit_glm_gamma, X, type = "response")
}
	\end{lstlisting}
	\item Poisson GLM:
	We keep the log link, but now use a Poisson distribution because this gives us a GLM with canonical link with good calibration properties.
	For instance, we do not need a correction factor.
	\item Gradient boosted trees via \href{https://xgboost.readthedocs.io}{XGBoost} \cite{chen2016} with Gamma deviance loss and log link:
	Parameters are tuned by five-fold cross-validation on the training data.
	As for the Gamma GLM, we apply a multiplicative correction factor \lstinline{corr_fact_xgb = 1.194863} in order to be unbiased on the training data.
\end{itemize}

Instead of a multiplicative correction factor, we could have corrected the models by adding a constant.
An additive constant, however, has the disadvantage that it could render some predictions negative.
A multiplicative correction is better suited for models with a log link and models fit on the log transformed response as it is just an additive constant in link-space.
For the Gamma GLM, it corresponds to adjusting the intercept term.

For this user guide the above five models are sufficient.
Further models or improvements could consist in  a separate large claim handling, \eg shown in \cite{fissler2021deep}, more feature engineering with larger model pipelines, usage of other model types like neural nets, stacking of models, and many more.

\section{Calibration and Identification Functions}
\label{sec:Calibration and Identification Functions}

\TheorySec

Once a certain model $m\in\M$ has been fitted, important questions arise: \enquote{Is the model fit for its purpose?}
Or similarly: \enquote{Does the model produce predictions which are in line with the observations?}
To answer these questions, the statistical notion of \emph{calibration} is adequate, which was coined in \cite{Davis2016} and further examined in \cite{NoldeZiegel2017}.

\subsubsection{Conditional calibration and auto-calibration}
To introduce a formal definition of calibration, recall that exactly predicting $Y$ using a feature vector $\bX\in\X$ is illusive.
In almost all applications, $\bX$ does not describe $Y$ entirely in the sense that $Y$ is a deterministic function of $\bX$.
That means $Y$ given $\bX$ is stochastic, and we summerise this stochasticity in terms of a functional $T$.
Our definitions of calibration follow \cite{NoldeZiegel2017, KrugerZiegel2021}.
\begin{definition} \label{defn:calibration}
Given a feature--response pair $(\bX,Y)$, the model $m(\bX)$ is \emph{conditionally calibrated} for the functional $T$ if 
\[
m(\bX) = T(Y|\bX) \qquad \text{almost surely}.
\]
The model $m(\bX)$ is \emph{auto-calibrated} for $T$ if
\[
m(\bX) = T(Y|m(\bX)) \qquad \text{almost surely}.
\]
\end{definition}
In other words, a conditionally calibrated model uses the entire feature information ideally in the prediction task. 
It describes the oracle regression function $\X\ni \bx\mapsto T(Y|\bX = \bx)$.
On the other hand, auto-calibration merely says that the information in the model $m(\bX)$ is used ideally to predict $Y$ in terms of $T$. 
Clearly, the one-dimensional $m(\bX) \in \R$ is less informative than the feature $\bX \in \X$ itself.
In the extreme case when $m(\bX)$ is a constant  model, it contains no information. 
Being auto-calibrated then simply means that the model equals the unconditional functional value $T(Y)$.\footnote{%
	In an insurance context, this would amount to assessing the risk of each client of a car insurance with the very same number, no matter what age, health conditions and experience they have, or if they have ever been involved in a car accident before.
}
For the rich class of \emph{identifiable} functionals so called moment or identification functions are a tool to assess conditional calibration and auto-calibration.
\begin{definition} \label{defn:ident funct}
Let $\F$ be a class of probability distributions where the functional $T$ is defined on.
A \emph{strict $\F$-identification function} for $T$ is a function $V(z,y)$ in a forecast $z$ and an observation $y$ such that 
\begin{equation} \label{eq:ident func}
	\int V(z,y)\dint F(y) = 0 \quad \Longleftrightarrow \quad z=T(F) \qquad \text{for all } z\in\R, \ F\in\F \,.
\end{equation}
If only the implication $\Longleftarrow$ in \eqref{eq:ident func} holds, then $V$ is just called an $\F$-identification function for $T$.
If $T$ admits a strict $\F$-identification function, it is \emph{identifiable} on $\F$.
\end{definition}

\renewcommand{\arraystretch}{1.1}
\begin{table}%[b]
	\centering
	\begin{tabular}{llll} 
		\toprule
		Functional & Strict Identification Function & Domain of $y$, $z$\\
		\midrule
		expectation $\E[Y]$ & $V(z, y) = z - y$ & $\R$\\
		$\alpha$-expectile & $V(z, y) = 2\abs{\one\{z \geq y\}-\alpha}(z - y)$ & $\R$\\
		median $F^{-1}_Y(0.5)$ & $V(z, y) = \one\{z \geq y\}-\sfrac{1}{2}$ & $\R$\\
		$\alpha$-quantile $F^{-1}_Y(\alpha)$ & $V(z, y) = \one\{z \geq y\}-\alpha$ & $\R$\\
		\bottomrule
	\end{tabular}
	\caption{Table of canonical strict identification functions.}
	\label{tab:identification_functions}
\end{table}
\renewcommand{\arraystretch}{1.0}
\autoref{tab:identification_functions} displays the most common strict identification functions for the mean and median as well as their generalisations, $\alpha$-quantiles for the latter and $\alpha$-expectiles for former.\footnote{%
	We use the indicator function $\one\{A\} = \begin{cases} 1\,,\quad \text{if $A$ is true}\\ 0\,,\quad\text{else}.\end{cases}$
}
When $T$ is the mean functional, (conditional) calibration boils down to (conditional) unbiasedness.
Not all functionals are identifiable.
Most prominently, the variance, the expected shortfall (an important quantitative risk measure) and the mode (the point with highest density) fail to be identifiable \cite{Osband1985, gneiting_2011, SteinwartPasinETAL2014, HeinrichFissler2021}.
Strict identification functions are not unique: 
If $V(z,y)$ is a strict $\F$-identification function for $T$, so is $\widetilde{V}(z,y) = h(z) V(z,y)$, where $h(z)\neq 0$ for all $z$.\footnote{%
	It is actually also possible to show that all strict identification functions must be of this form, see Theorem S.1 in \cite{DFZ2020}.
}

The connection between calibration and identification functions is as follows:
Let $V$ be any strict $\F$-identification function for $T$ and suppose further that $\F$ contains the conditional distributions $F_{Y|\bX=\bx}$ for almost all $\bx\in\X$.
Then, an application of \eqref{eq:ident func} to these conditional distributions yields that $m(\bx) = T(Y|\bX=\bx)$ if and only if 
$\int V(m(\bx),y)\dint F_{Y|\bX = \bx}(y) = 0$ .
This shows that $m$ is conditionally calibrated for $T$ if and only if 
\begin{equation}\label{eq:cond cal}
\E[V(m(\bX),Y)|\bX]=0\qquad \text{almost surely}.
\end{equation}
Similarly, if the conditional distributions $F_{Y|m(\bX) = z}$ are in $\F$ for almost all $z\in\R$, then $m$ is auto-calibrated for $T$ if and only if 
\begin{equation}\label{eq:auto cal}
\E[V(m(\bX),Y)|m(\bX)]=0\qquad \text{almost surely}.
\end{equation}
Hence, for identifiable functionals with a sufficiently rich class $\F$, the tower property of the conditional expectation yields that a conditionally calibrated model is necessarily auto-calibrated.\footnote{%
	This implication does not necessarily hold for non-identifiable functionals.
	Consider the variance functional, let $Y$ be standard normal and consider the single feature $X = -Y$.
	Clearly, $X$ fully describes $Y$. Hence, $\Var[Y|X] = 0$. The zero model $m(X)=0$ is conditionally calibrated.
	On the other hand, it provides no information about $Y$ at all. Consequently, conditioning on it has no effect.
	Since $\Var[Y]=1$, the zero model is not auto-calibrated.
}
The reverse implication generally does not hold as can be seen, \eg with constant models.

\subsubsection{Unconditional calibration}
Besides conditional calibration and auto-calibration, one can also find the notion of \emph{unconditional} calibration in the literature on identifiable functionals. 
If $V$ is a strict identification function for $T$, then we say that $m(\bX)$ is unconditionally calibrated for $T$ \emph{relative to $V$} if 
$\E[V(m(\bX),Y)]=0$. 
Unless $m(\bX)$ is constant, and in stark contrast to conditional calibration and auto-calibration, the notion of unconditional calibration depends on the choice of the identification function $V$ used.
Moreover, there is no definition solely in terms of $T$ and $m(\bX)$ as in Definition~\ref{defn:calibration}, unless the model is constant such that the notions of unconditional calibration and auto-calibration coincide.
As pointed out in \cite{NoldeZiegel2017}, non-constant models that are not conditionally calibrated (or auto-calibrated) can still be unconditionally calibrated.
The different notions of calibration are summarised in \autoref{tab:calibration}.
For a more detailed discussion, we refer to \cite{Pohle2020} and \cite{gneiting2021regression}.
\begin{table}
	\centering
	\begin{tabular}{lll} 
		\toprule
		Notion & Definition & Check\\
		\midrule
		conditional calibration & $m(\bX) = T(Y|\bX)$ & $\E[V(m(\bX),Y)|\bX]=0 \quad a.s.$\\
		auto-calibration & $m(\bX) = T(Y|m(\bX))$ & $\E[V(m(\bX),Y)|m(\bX)]=0 \quad a.s.$\\
		unconditional calibration & $\E[V(m(\bX),Y)]=0 $  & $\E[V(m(\bX),Y)]=0 $\\
		\bottomrule
	\end{tabular}
	\caption{Types of calibration for an identifiable functional $T$ with strict identification function $V$.}
	\label{tab:calibration}
\end{table}

\subsubsection{Assessment of calibration}
On a sample $D = \{(\bx_i,y_i), \ i=1, \ldots, n\}$, unconditional calibration can be empirically tested by checking whether $\overline{V}(m; D) = \frac{1}{n} \sum_{(\bx_i,y_i) \in D} V(m(\bx_i),y_i)$ is close to 0.
This property justifies the term generalised residuals for the quantities $V(m(\bx_i),y_i)$.
Next to pure diagnostics, this can also be accompanied with a Wald test in an asymptotic regime.
To check conditional calibration, one needs to assess \eqref{eq:cond cal} on a sample level.
The presence of the conditional expectation in \eqref{eq:cond cal} complicates this check.
Formally, \eqref{eq:cond cal} is equivalent to
\begin{equation} \label{eq:cond cal test functions}
	\E[\ph(\bX)V(m(\bX),Y)] = 0 \quad \text{for \emph{all} (measurable) test functions } \ph\colon \X \to \R.
\end{equation}
Clearly, implementing \eqref{eq:cond cal test functions} in practice is usually not feasible.%
\footnote{%
	An important instance where this is possible is when the features are categorical only such that $\X = \{\bx_1, \ldots, \bx_k\}$.
	Then \eqref{eq:cond cal} is equivalent to $\E[\one\{\bX = \bx_j\}V(m(\bX),Y)] =0$ for all $\bx_j \in \X$.
}
One needs to choose a finite number of test functions $\ph_1, \ldots, \ph_k$ and check whether $\E[\ph_j(\bX)V(m(\bX),Y)] =0$ for all $j=1, \ldots, k$, 
which can be done with a Wald test. 
On the other hand, \eqref{eq:cond cal} can be assessed visually with appropriate plots.
or by checking suitable plots.
The choice of the specific test functions and the overall number of test functions can greatly influence the power of the corresponding Wald test.
On the population level, the more test functions, the higher the power of the test.
But in the practically relevant situation of finite samples, increasing the number of test functions also increases the overall estimation error.
Hence, one faces a trade-off.
In many situations, one is left with practically testing a broader, \ie less informative, null hypothesis than \eqref{eq:cond cal}.
So if the Wald test fails to reject the null $\E[\ph_j(\bX)V(m(\bX),Y)] =0$ for all $j=1, \ldots, k$, one cannot be sure if \eqref{eq:cond cal} actually holds.%
\footnote{%
	On top of that, recall the general limitations in statistical tests that it is not possible to establish a null hypothesis by not rejecting it.
}
It might thus be the case that a model is deemed conditionally calibrated even if it does not make ideal use of the information contained in $\bX$.

The counterpart of \eqref{eq:cond cal test functions} for testing auto-calibration is
\begin{equation*} 
	\E[\ph(m(\bX))V(m(\bX),Y)] = 0 \quad \text{for \emph{all} (measurable) test functions\;} \ph\colon \R \to \R.
\end{equation*}
This time $\ph$ is a univariate function of $m(\bX)$ only, which greatly simplifies the practical implementation.
The notion of auto-calibration and its assessment become important when the feature vector $\bX$ is unknown in the evaluation process, but only the model (prediction) $m(\bX)$ along with the response $Y$ are reported, thus adhering to the weak prequential principle of \cite{DawidVovk1999}.
\par\medskip

Calibration gives us only a very limited possibility to actually compare two models.
What if both of them are (auto-)calibrated, but one is more precise in that it incorporates strictly more information than the other one?
What about two miscalibrated models?%
\footnote{%
	This is particularly relevant under model-misspecification.
	That means that the model class $\M$ is too narrow such that there is no $m^\star$ with $m^\star(\bX) = T(Y|\bX)$.
	This can arise, \eg when a monotone model is used to capture a relation with a clear saturation point.
	This can be observed, for instance, when $\bX$ is a driver's age and $T(Y|\bX) = \P(Y=1|\bX)$ is the probability of having a car accident ($Y=1$), given that age.
	The effect of age on the accident probability is often bathtub-like: young drivers have more accidents, then become safer drivers over time until the risk of accident increases again.
}
And how should we think of the situation where we have a calibrated, but uninformative model (\eg making use of only one covariate) versus a very informative one, which is, however, slightly uncalibrated?
On the one hand, such comparisons can be made by the expert at hand, say, the actuary, taking into regard the economic relevance of these differences.%
\footnote{%
	For example, is a more discriminative price policy (say, based on age, health, sex, race etc.) practically implementable, or legal at all?
	If not, then a less informative but well calibrated model might be more adequate.
}
On the other hand, we shall introduce the notion of consistent scoring functions in \autoref{sec:model comparison}.
Assessing conditional calibration and resolution (a measure of the information content used in the model or its discrimination ability) simultaneously, they constitute an adequate tool to overall quantify prediction performance, and to ultimately compare and rank different models.

\InsightsSec

\subsubsection{Best practice recommendations}
\begin{itemize}
	\item Check for calibration on the training as well as on the test set.
	On the training set, this yields insights on further modelling improvements, such as including an important omitted feature or excluding a less informative one.
	(One should, however, mind the overfitting trap when working on the training set.)
	Once training is finished, assessing calibration on the test set provides a more unbiased view of the calibration and indications whether there is severe bias in the model or whether the model is fit for predictive usage.
	\item Quantify calibration by making a choice for the test function $\ph$ and report $\overline{V}_{\ph}(m; D) = \frac{1}{n}\sum_{i=1}^n \ph(\bx_i)V(m(\bx_i),y_i)$.
	For practical and diagnostic purposes, we recommend to consider at least $\ph(\bx)= 1$ as well as all projections to single components of the feature vector $\bx$ (supposed we do not have too many feature components) in addition with the (fitted) model $m(\bx)$, where the latter explicitly assesses auto-calibration.
	Moreover, the test functions $\ph$ can be chosen to bin continuous feature variables; see \cite{FKP2022} for more details.\footnote{%
	If, \eg a numerical feature variable \texttt{income} is reported, we can bin it in, say, three categories \texttt{low}, \texttt{middle}, and \texttt{high}, defined through two thresholds, say $c_{\text{middle}}$ and $c_{\text{high}}$. 
	Then we can consider the test functions 
	$\ph_{\text{low}}(\bx) = \one\{\texttt{income}\le c_{\text{middle}}\}$, 
	$\ph_{\text{middle}}(\bx) = \one\{c_{\text{middle}} < \texttt{income} \le c_{\texttt{high}}\}$, and
	$\ph_{\text{high}}(\bx) = \one\{\texttt{income} >c_{\texttt{high}}\}$.
	}
	\item 
	Assess calibration visually:
	For a test function $\ph$, plot the generalised residuals $V(m(\bx_i), y_i)$ versus $\ph(\bx_i)$ for the above choices of test functions $\ph$.
	Check whether the average of the generalised residuals is around 0 for all values of the test function.
	Another possibility is to plot the values of $\overline{V}_{\ph}(m; D)$ for different $\ph$, for instance projections to single feature columns; see \cite{FKP2022} for more details.

	In addition, auto-calibration can be visually assessed by means of a reliability diagram: a graph of the mapping $m(\bx) \rightarrow T(Y|m(\bx))$, see \cite{gneiting2021regression}.
	For identifiable $T$, $T(Y|m(\bx))$ can be estimated by isotonic regression of $y_i$ against $m(x_i)$ as shown in \cite{Jordan2022}.
	For an auto-calibrated model, the graph is the diagonal line.
	Note that due to the estimation error of $T(Y|m(\bx))$ the graph usually shows some deviation from the diagonal even for an auto-calibrated model.
	In \autoref{fig:reliability diagrams}, an example is given for binary classification where $T(Y|m(\bx))=\E[Y|m(\bx)]$ is called conditional event probability.
\end{itemize}

\subsubsection{GLMs with canonical link}
By construction, a fitted GLM with canonical link fulfils the simplified score equations
\begin{equation} \label{eq:canonical score equation}
	0 = \sum_i x_{ij} (m(\bx_i) - y_i) %= \left(\mathbf{x}' (y - m(\mathbf{x}))\right)_j 
	\quad \text{for all components $j$ of the feature vector}
\end{equation}
on the training set.\footnote{%
	Often in the GLM context, $m(\bx_i)$ is written as $\mu_i$.
}
This means that \eqref{eq:cond cal test functions} is satisfied on the training set for all projections $\ph$ to single features.
By incorporating a constant feature to account for the intercept, \eqref{eq:canonical score equation} yields the 
so called balance property \cite{Wuthrich2021StatisticalFoundations}, $\overline{V}(m; D_{\text{train}}) = \frac{1}{n_{\text{train}}} \sum_{(\bx_i,y_i) \in D_{\text{train}}} m(\bx_i) - y_i = 0$, which amounts to unconditional calibration on the training set.
Moreover, dummy coding and \eqref{eq:canonical score equation} ensure that this balance property also holds on all labels of categorical features.
Note that \eqref{eq:cond cal test functions} is not necessarily fulfilled for other choices of $\ph$ on the training set, nor is it clear that \eqref{eq:cond cal test functions} holds on the test set in the presence of an estimation error.

\ExampleSec{Mean Regression for Workers' Compensation.}

As we are modelling the expectation, we use the identification function $V(z, y) = z - y$, corresponding to the bias or negative residual.
This immediately leads to the well-known analysis of residuals.

\subsubsection{Unconditional calibration}
Checking for unconditional calibration amounts to checking for the average bias $\overline{V}(m; D) \allowbreak = \frac{1}{n} \sum_{i \in D} m(\bx_i) - y_i = \frac{1}{n}\sum_{i \in D} \mathrm{bias}_i$.
Normalising with  the sample size $n$ ensures that results on training and test set are directly comparable despite having different sizes.	
\begin{table}
	\centering
	\begin{tabular}{lrr@{\hskip 1cm}ll}
		\toprule
		& \multicolumn{2}{c}{$\overline{V}(m; D)$} & \multicolumn{2}{c}{$p$-value of $t$-test} \\
		Model & $D=\text{train}$ & $D=\text{test}$ & $p$ train & $p$ test \\
		\midrule
		Trivial        & \num{0}      & \textbf{\num{-24}}   & \num{1.0e+  0} & \num{9.5e-  1} \\
		OLS            & \num{-11045} &         \num{-11049} & \num{0       } & \num{6.4e-188} \\
		OLS corr       & \num{0}      &         \num{109}    & \num{1.0e+  0} & \num{7.6e-  1} \\
		GLM Gamma      & \num{-1335}  &         \num{-1207}  & \num{1.1e-  9} & \num{8.8e-  4} \\
		GLM Gamma corr & \num{0}      &         \num{146}    & \num{1.0e+  0} & \num{6.9e-  1} \\
		GLM Poisson    & \num{0}      &         \num{125}    & \num{1.0e+  0} & \num{7.3e-  1} \\
		XGBoost        & \num{-2120}  &         \num{-2044}  & \num{1.6e- 22} & \num{1.4e-  8} \\
		XGBoost corr   & \num{0}      &         \num{96}     & \num{1.0e+  0} & \num{7.9e-  1} \\
		\bottomrule
	\end{tabular}
	\caption{Assessment of unconditional calibration in terms of bias.}
	\label{tab:unconditional calibration}
\end{table}
From the numbers in \autoref{tab:unconditional calibration}, we confirm that the correction factor eliminated the bias over the whole training set, while there is some bias on the test set.
The bias on the test set might be due to estimation error on the training set on the one hand or to sampling error of the test set on the other hand.
As expected, the Poisson GLM is unbiased on the training set out of the box as a consequence of the balance property for GLMs with canonical link.
This, however, no longer holds on the test set, where the corrected XGBoost is the best one apart from the trivial model.

\autoref{tab:unconditional calibration} additionally provides $p$-values for the two-sided $t$-tests with null hypotheses $\E[V(m(\bX), Y)] = 0$ for the respective models $m$.
The advantage of $p$-values lies in a scale free representation of the statistic $\overline{V}$ within the range $[0, 1]$ and the meaning: the larger the better calibrated.

\subsubsection{Auto-Calibration}
We analyse auto-calibration by visualising bias versus predicted values.
On the training set, this is a vertically flipped version of the well-known residual versus fitted plot.
\begin{figure}%[htbp]
	\centering
	\includegraphics[width=.95\textwidth]{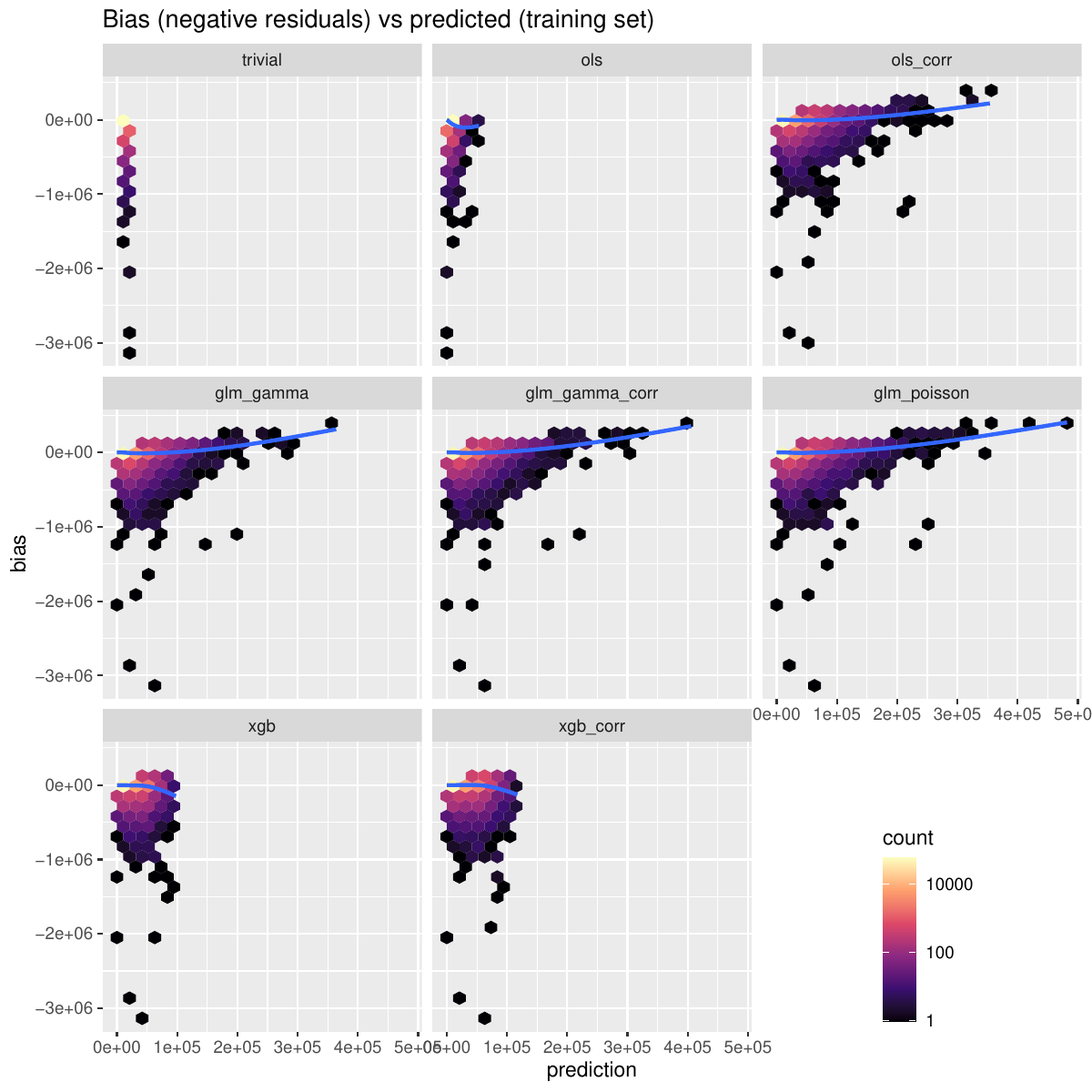}
	\caption{Bias (negative residuals) vs predicted values with smoothed lines for the average bias.
	Note that the trivial model in the top left plot depicts a straight line, the distortion is due to the hexagon based visualisation.}
	\label{fig:auto calibration training set}
\end{figure}
The smoothed lines in \autoref{fig:auto calibration training set} are an estimate of the average bias per prediction.
We observe large differences in the range of predictions among the models.
The three GLMs and the corrected OLS model show a wide range of predicted values with the Poisson GLM having the largest range with the largest predicted value, while the OLS has the smallest range and also the smallest predicted value---neglecting the trivial model.
These peculiarities of the OLS are due to the backtransformation after modelling on the log-scale.
By comparing the bias term with maximum absolute value, we get another valuable information.
The OLS model has the bias term with largest absolute value, on the training set (\num{-3.131e6}) as well as on the test set (\num{-1.666e6}), the corrected Gamma GLM has the smallest such value on the training set (\num{-3.073e6}), the corrected XGBoost exhibits the smallest one on the test set (\num{-1.634e6}).
For the three GLMs and the corrected OLS model, the bias tends to increase with increasing prediction.
On the other hand, both XGBoost models show the opposite behaviour.
Both phenomena might be due to the fact that there are only a few observations of features leading to large predictions such that the estimation error might be relatively high for such feature constellations.
\begin{figure}%[htbp]
	\centering
	\includegraphics[width=.95\textwidth]{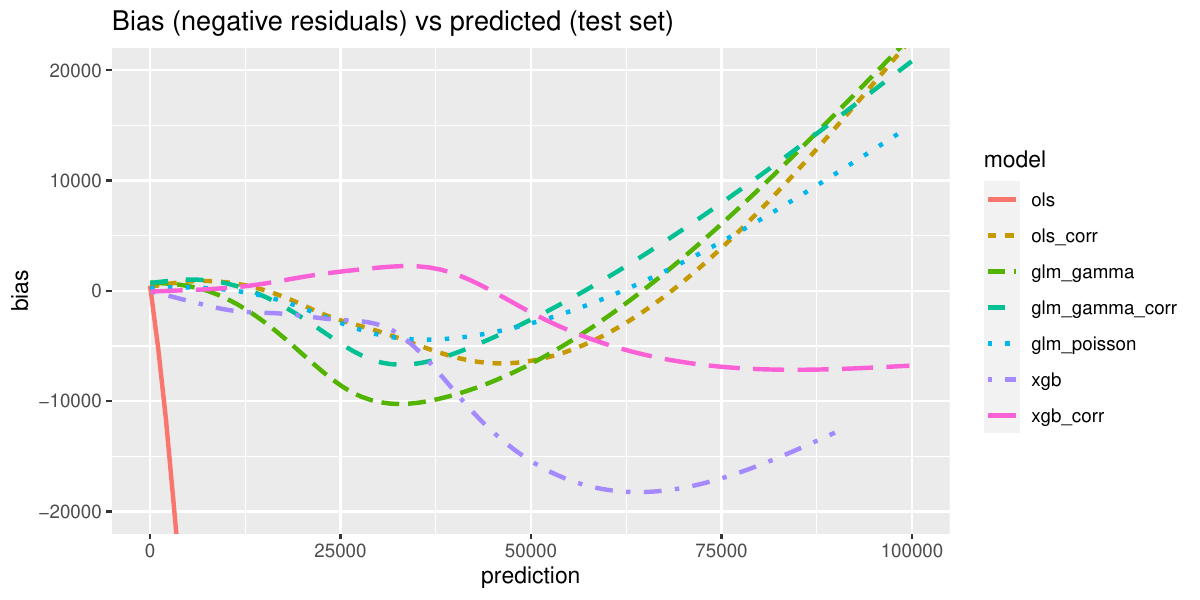}
	\caption{Smoothed lines for average bias (negative residuals) vs predicted values, truncated at $10^5$.
	}
	\label{fig:auto calibration test set}
\end{figure}
For a better view on the range with most exposure, we draw the smoothed lines only on the restricted range $m(\bx) \le 10^5$, but this time on the test set, see \autoref{fig:auto calibration test set}.
Here, it becomes visible that the corrected XGBoost model seems to have the best auto-calibration in that range.

\begin{table}
	\centering
	\begin{tabular}{lrr}
		\toprule
		& \multicolumn{2}{c}{$\overline{V}_\ph(m; D)$} \\
		Model & $D=\text{train}$ & $D=\text{test}$ \\
		\midrule
		Trivial        & \num{0}         & \textbf{\num{-3.139e5}}\\
		OLS            & \num{-6.201e+7} &         \num{-5.987e7}\\
		OLS corr       & \num{-2.266e+6} &         \num{3.110e7}\\
		GLM Gamma      & \num{-3.332e+7} &         \num{-9.977e6}\\
		GLM Gamma corr & \num{1.162e+7}  &         \num{3.996e7}\\
		GLM Poisson    & \num{3.525e+7}  &         \num{7.485e7}\\
		XGBoost        & \num{-9.254e+7} &         \num{-6.777e7}\\
		XGBoost corr   & \num{-3.721e+7} &         \num{-6.723e6}\\
		\bottomrule
	\end{tabular}
	\caption{Assessment of auto-calibration with test function $\ph(\bx) = m(\bx)$.}
	\label{tab:auto-calibration}
\end{table}
The next step is to evaluate different test functions.
We list the values of $\overline{V}_\ph$ for the test function $\ph(\bx) = m(\bx)$ in \autoref{tab:auto-calibration}.
The results are similar in quality to the ones concerning unconditional calibration.
Among the non-trivial models, it is again the calibrated XGBoost model that is best in class on the test set.

\subsubsection{Calibration conditional on Gender}
We now have a look at the categorical variable \variable{Gender}.
Therefore, we evaluate $\overline{V}_\ph(m; D)$ for the two projection test functions $\ph(\bx) = \one\{\variable{Gender}=\text{\enquote{F}}\}$ and $\ph(\bx) = \one\{\variable{Gender}=\text{\enquote{M}}\}$.

\begin{table}
	\centering
	\begin{tabular}{lrrrr}
		\toprule
		& \multicolumn{4}{c}{$\overline{V}_\ph(m; D)$} \\
		& \multicolumn{2}{c}{train} & \multicolumn{2}{c}{test}\\
		Model & bias F & bias M & bias F & bias M\\
		\midrule
		Trivial        & \num{-969}  & \num{969}   &         \num{-934}  &         \num{910}\\
		OLS            & \num{-3269} & \num{-7776} &         \num{-3250} &         \num{-7799}\\
		OLS corr       & \num{-246}  & \num{246}   &         \num{-152}  &         \num{261}\\
		GLM Gamma      & \num{-274}  & \num{-1061} &         \num{-178}  &         \num{-1029}\\
		GLM Gamma corr & \num{130}   & \num{-130}  &         \num{237}   &         \num{-91}\\
		GLM Poisson    & \num{0}     & \num{0}     & \textbf{\num{105}}  & \textbf{\num{20}}\\
		XGBoost        & \num{-858}  & \num{-1262} &         \num{-779}  &         \num{-1265}\\
		XGBoost corr   & \num{-284}  & \num{284}   &         \num{-191}  &         \num{286}\\
		\bottomrule
	\end{tabular}
	\caption{Assessment of calibration conditional on \variable{Gender} with test functions $\ph(\bx) = \one\{\variable{Gender}=\text{\enquote{F}}\}$ and $\ph(\bx) = \one\{\variable{Gender}=\text{\enquote{M}}\}$.}
	\label{tab:conditional calibration by Gender}
\end{table}
\begin{figure}%[htbp]
	\centering
	\includegraphics[width=.90\textwidth]{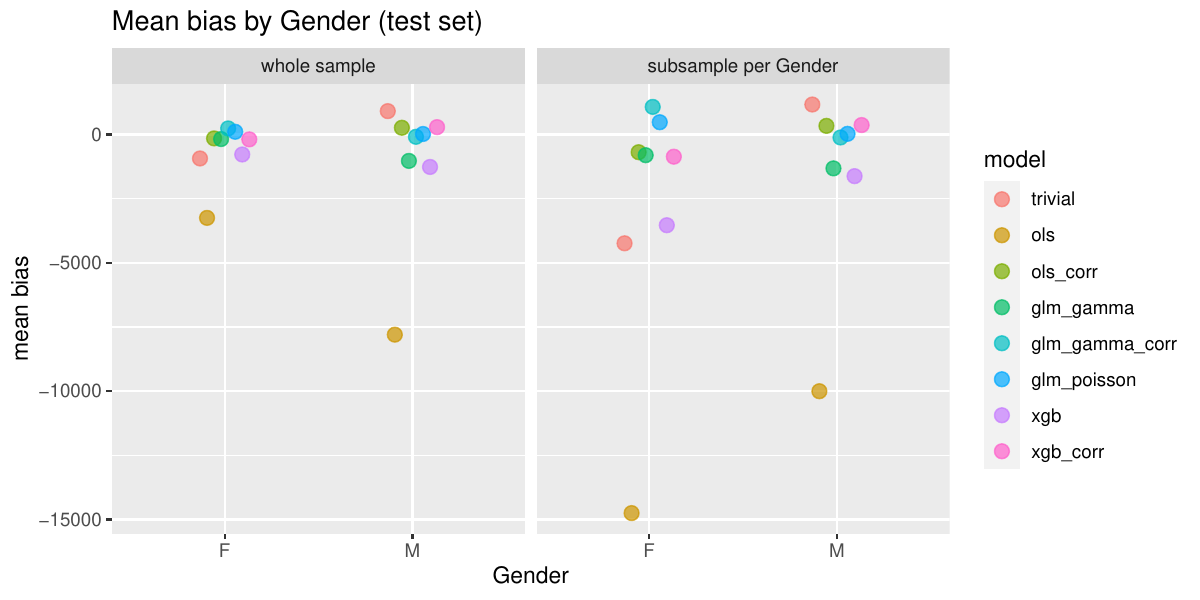}
	\caption{Plot of mean bias on the test set.
	Left: $\overline{V}_\ph(m; D)$ for the two test functions $\ph(\bx) = \one\{\variable{Gender}=\text{F}\}$ and $\ph(\bx) = \one\{\variable{Gender}=\text{M}\}$.
	Right: $\overline{V}$ evaluated on the subsamples with $\variable{Gender}=\text{\enquote{F}}$ and $\variable{Gender}=\text{\enquote{M}}$, respectively.}
	\label{fig:residual barplot by Gender}
\end{figure}
From the bare numbers in \autoref{tab:conditional calibration by Gender} as well as from the implied bar plot on the left-hand side of \autoref{fig:residual barplot by Gender}, we observe again the perfect calibration of the Poisson GLM on the training set due to the balance property \eqref{eq:canonical score equation}.
This time, this carries over to the test set to a considerable degree.
Interestingly, only the Poisson and the corrected Gamma GLM have a positive mean bias for $\variable{Gender}=\text{\enquote{F}}$ on the test set.
These two models seem to have the best out-of-sample calibration conditional on \variable{Gender}.

A different perspective arises if, instead of evaluating $\overline{V}_\ph(m; D)$ on the whole training or test sample, we evaluate the mean bias $\overline{V}(m; D)$ on the two projected subsamples $D$ with $\variable{Gender}=\text{\enquote{M}}$ and $\variable{Gender}=\text{\enquote{F}}$ separately.
This corresponds to a different normalisation of the averaging step such that one obtains a mean bias per categorical level---a valuable information about possible discrimination.
It can be seen from the right-hand side of \autoref{fig:residual barplot by Gender} that the bias for \enquote{F} is larger than for \enquote{M}, in particular the opposite of the left plot.

Remarkably, it turns out that the models with correction factor are not only better unconditionally calibrated than the uncorrected ones, but they are also better conditionally calibrated on \variable{Gender}, in-sample as well as out-of-sample.
If we were to place great importance on calibration, we could exclude the three model variants without correction factor, OLS, Gamma GLM and XGBoost, from further analysis as they persistently show worse calibration than their counterparts with calibration factor.

We could go on and investigate interactions with other features, for instance, use $\ph(\bx) = \variable{LogWeeklyPay} \cdot \one\{\variable{Gender}=\text{\enquote{M}}\}$.

\subsubsection{Calibration conditional on LogWeeklyPay}
\begin{figure}%[htbp]
	\centering
	\includegraphics[width=.90\textwidth]{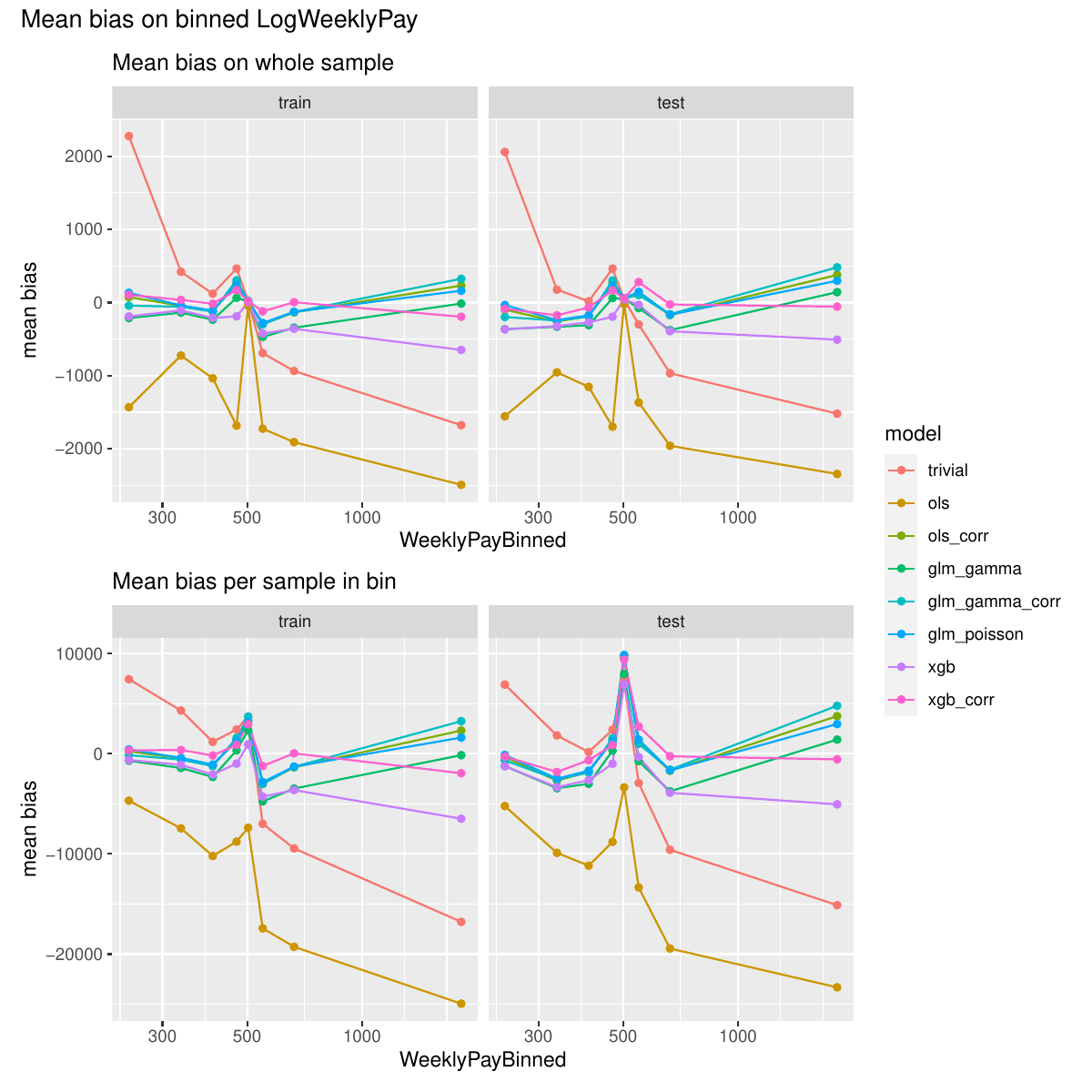}
	\caption{Mean bias (negative residual) conditional on binned (quantile based) \variable{LogWeeklyPay}.
	Top: $\overline{V}_\ph(m; D)$ for the test functions $\ph$ projecting on each bin.
	Bottom: $\overline{V}(m; D_{\text{bin}})$ for the subsample of each bin.}
	\label{fig:residual plot LogWeeklyPay}
\end{figure}
In order to assess the calibration conditional on the continuous variable \variable{LogWeeklyPay}, we bin it on the whole data set into intervals of \num{11} evenly distributed quantiles (\num{0}, \SI{10}{\percent}, \SI{20}{\percent}, \ldots, \SI{100}{\percent}).
Since the lowest value $\variable{WeeklyPay} = 200$ occurs in over \SI{22}{\percent} of the data rows, we end up with only 8 unique bins.
Then we compute
\begin{itemize}
	\item $\overline{V}_\ph(m; D)$ for the projection test functions $\ph_k(\bx)=\one\{\variable{LogWeeklyPay} \in \text{bin }k\}$ for all bins $k=1, \ldots, 8$; and
	\item the average bias $\overline{V}(m; D_{\text{bin}})$ per bin.
\end{itemize}
To plot the bins on the $x$-axis, we choose the midpoint of each bin.
While it is the logarithm of \variable{WeeklyPay} that enters the models, we display the results in \autoref{fig:residual plot LogWeeklyPay} on the original scale.
It is obvious that the OLS model without correction is very biased as a consequence of the backtransformation after modelling on the log scale of $y$.
We rate the corrected XGBoost model as the most unbiased one, conditional on \variable{LogWeeklyPay}.

\begin{table}
	\centering
	\begin{tabular}{lrr}
		\toprule
		& \multicolumn{2}{c}{$\overline{V}_{\ph}(m; D) $} \\
		Model & $D=\text{train}$ & $D=\text{test}$ \\
		\midrule         
		Trivial        & \num{ -3730} &         \num{ -3502} \\
		OLS            & \num{-69362} &         \num{-69102} \\
		OLS corr       & \num{   125} &         \num{  1158} \\
		GLM Gamma      & \num{ -8107} &         \num{ -6985} \\
		GLM Gamma corr & \num{   313} &         \num{  1549} \\
		GLM Poisson    & \num{     0} &         \num{  1118} \\
		XGBoost        & \num{-13569} &         \num{-12788} \\
		XGBoost corr   & \num{  -298} & \textbf{\num{   611}} \\
		\bottomrule
	\end{tabular}
	\caption{Assessment of conditional calibration with respect to \variable{LogWeeklyPay}, using test function $\ph(\bx)=\variable{LogWeeklyPay}$.}
	\label{tab:conditional calibration test funcion}
\end{table}
Similar to what we have done for assessing auto-calibration, we also check with the test function $\ph(\bx) = \variable{LogWeeklyPay}$.
This leads to computing $\overline{V}_{\ph}(m; D) = \allowbreak \frac{1}{n} \allowbreak \sum_{(\bx_i,y_i) \in D} \variable{LogWeeklyPay}_i (m(\bx_i) - y_i)$, see \autoref{tab:conditional calibration test funcion}.
We see that the Poisson GLM passes this check perfectly on the training set which is a result of the score equation Eq.~\eqref{eq:canonical score equation}.
On the test set, the corrected XGBoost has again the smallest value, which confirms our earlier findings:
overall, the corrected XGBoost seems to be the best calibrated model.

\section{Model Comparison and Selection with Consistent Scoring Functions} \label{sec:model comparison}

\TheorySec

\renewcommand{\arraystretch}{1.1}
\begin{table}
	\centering
	\begin{tabular}{llll} 
		\toprule
		Functional & Scoring Function & Formula $S(z, y)$ & Domain\\ 
		\midrule 
		expectation $\E[Y]$ & squared error &$(y - z)^2$ & $y,z \in \R$\\
		& Poisson deviance & $2 \, (y \log\frac{y}{z} + z - y)$ & $y \geq 0, z>0$\\
		& Gamma deviance & $2 \, (\log\frac{z}{y} + \frac{y}{z} - 1)$ & $y,z > 0$\\
		& Tweedie deviance & $2 \, \Big(\frac{ y^{2-p}}{(1-p)\cdot(2-p)}$ & $y,z > 0$\\
		& \hspace{12pt} $p \in \R \setminus \{1, 2\}$ & \hspace{12pt} $-\frac{y\cdot z^{1-p}}{1-p}+\frac{z^{2-p}}{2-p}\Big)$ & \hspace{7pt} $y \geq 0$ for $p < 2$\\
		& homogeneous score & $\abs{y}^a - \abs{z}^a$ & $y,z \in \R$\\
		& \hspace{12pt} $a>1$ & \hspace{12pt} $- a \sign(z)\abs{z}^{a-1}(y - z)$ & \\
		& log loss & $-y\log{z} - (1-y)\log(1-z) $ & $0 \leq y \leq 1$ \\
		&          & \hspace{12pt} $+y\log{y} + (1-y)\log(1-y) $            & \hspace{7pt} $0<z<1$ \\
		\midrule
		$\alpha$-expectile & APQSF & $\abs{\one\{z \geq y\}-\alpha}(z - y)^2$ & $\R$\\
		\midrule
		median $F^{-1}_Y(0.5)$ & absolute error & $\abs{y - z}$ & $\R$\\
		\midrule
		$\alpha$-quantile $F^{-1}_Y(\alpha)$ & pinball loss & $(\one\{z \geq y\}-\alpha)(z - y)$ & $\R$\\
		\bottomrule
	\end{tabular}
	\caption{Examples of strictly consistent scoring functions.\newline
	For more details about the Tweedie deviance, see \autoref{sec:tweedie deviance}.
	The homogeneous score arises from Eq.~\eqref{eq:Bregman} with $\phi(x)=\abs{x}^a$.
	Note that homogeneous score and Tweedie deviance coincide on the common domains, up to a multiplicative constant.
	The terms in the second line of the log loss render it non-negative for all $0\le y \le 1$, but are zero for $y \in \{0,1\}$, see \autoref{sec:probabilistic binary classification}.
	APQSF stands for asymmetric piecewise quadratic scoring function.
	The pinball loss is also known as asymmetric piecewise linear scoring function.
	}
	\label{tab:scoring_functions}
\end{table}
\renewcommand{\arraystretch}{1.0}

Suppose we have two models $m_A, m_B \colon\X \to \R$ and we would like to compare and rank their accuracy or predictive performance.
Here, we are agnostic about how one has
come up with these models.
They might be the result of a statistically sound parametric estimation procedure, the yield of fitting the same model with different loss functions, the output of black box algorithms (which is often the case in Machine Learning), or they might be based on expert opinion (or an aggregate of different opinions), underlying physical models or pure gut guess.
Since we ignore how the models have been produced, particularly if and what training sample might have been used, we only care about \emph{out-of-sample performance}.

\subsubsection{Model comparison}
\label{subsubsec:model comparison}
So suppose we have a random test sample $D = D_{\text{test}} = \{(\bx_i, y_i), \ i=1, \ldots, n\}$.
The standard tool to assess prediction accuracy are \emph{scoring functions} $S(z,y)$, which is an alternative name for loss functions frequently used in the forecasting literature; see Footnote~\ref{footnote:loss} for further comments.
Just like identification functions, they are functions of the prediction $z = m(\bx)$ and the observed response $y$.
In a sense, they measure the distance between a forecast $z$ and the observation $y$, with the standard examples being the squared and the absolute prediction error, $S(z,y) = (z-y)^2$ and $S(z,y) = |z-y|$, 
such that smaller scores are preferable.
We estimate the expected score $\E[S(m(\bX), Y)]$, earlier called statistical risk, as
\begin{equation} \label{eq:empirical score}
	\overline{S}(m; D) = \frac{1}{n} \sum_{(\bx_i, y_i) \in D} S(m(\bx_i), y_i) \,,
\end{equation}
which we called empirical risk before.
Model $m_A$ is deemed to have an inferior predictive performance than model $m_B$ in terms of the score $S$ (and on the sample $D$) if
\begin{equation}\label{eq:score_diff}
	\overline{S}(m_A; D) - \overline{S}(m_B; D) = \frac{1}{n} \sum_{(\bx_i, y_i) \in D} S(m_A(\bx_i), y_i) - S(m_B(\bx_i), y_i) > 0.
\end{equation}
This pure assessment can be accompanied by a testing procedure to take into account the estimation error of the empirical risk.
Hence, we can test for superiority of $m_A$ by formally testing (and rejecting) the null hypothesis $\E[S(m_A(\bX), Y) - S(m_B(\bX), Y) ] \ge0$.
Similarly, one can also test the composite null that $\E[S(m_A(\bX), Y) - S(m_B(\bX), Y) ] \le 0$, or test the null hypothesis of equal predictive performance via the two-sided null $\E[S(m_A(\bX), Y) - S(m_B(\bX), Y) ] = 0$.
In a cross-sectional framework, these null hypotheses can all be addressed with a $t$-test which provides asymptotically valid results. In the context of predictive performance comparisons, such $t$-tests are also known as Diebold--Mariano tests, \cite{Diebold1995}.\footnote{%
	For more details about hypothesis testing with scoring functions, we refer to Chapter~3 of \cite{gneiting2014}.
	Serial correlation as in time-series needs to be accounted for in the estimation of the variance of the score difference.
	For correction in a cross-validation setting, see \cite{Nadeau_2003, Bouckaert_Frank_2004}.
	Finally, we hint at recent developments for e-values, see \cite{HenziZiegel2021}.
}

\subsubsection{Consistency and elicitability}
An important question in this model ranking procedure is the choice of the scoring function $S$.
At first glance, there is a multitude of scores to choose from.
To get some guidance, let us first ignore the presence of features and consider only constant models.
Then, the constant model $m(\bx) = c$ is correctly specified if $c = T(Y)$.
To ensure that the correctly specified model outperforms any misspecified model on average if evaluated according to \eqref{eq:score_diff}, one needs to impose the following consistency criterion on the score $S$, which was coined in \cite{MurphyDaan1985} and is discussed in detail in \cite{gneiting_2011}.

\begin{definition} \label{defn:consistent}
Let $\F$ be a class of probability distributions where the functional $T$ is defined on.
A scoring function $S(z,y)$ is a function in a forecast $z$ and an observation $y$.
It is \emph{$\F$-consistent} for $T$ if 
\begin{equation} \label{eq:consistent}
	\int S(T(F),y) \dint F(y) \le \int S(z,y) \dint F(y) \qquad \text{for all }z\in \R, \ F\in\F.
\end{equation}
The score is \emph{strictly} $\F$-consistent for $T$ if it is $\F$-consistent for $T$ and if equality in \eqref{eq:consistent} implies that $z=T(F)$.
If $T$ admits a strictly $\F$-consistent scoring function, it is \emph{elicitable} on $\F$.
\end{definition}

If the functional of interest is possibly set-valued (e.g.~in the case of quantiles), the usual modification of \eqref{eq:consistent} is that any value in the correctly specified functional $T(F)$ should outperform a forecast $z\notin T(F)$ in expectation.
Roughly speaking, strict consistency acts as a \enquote{truth serum}, encouraging to report correct forecasts.

As can be seen in \autoref{tab:scoring_functions}, quantiles and expectiles (and therefore the median and mean) are not only identifiable, but also elicitable, subject to mild conditions.
Indeed, it can be shown that under some richness assumptions on the class $\F$ and continuity assumptions on $T$, identifiability and elicitability are equivalent for one-dimensional functionals, see \cite{SteinwartPasinETAL2014}.
In line with this, variance and expected shortfall fail to be elicitable \cite{Osband1985, gneiting_2011}.
An important exception from this rule is the mode.
If $\Y$ is categorical, then the mode is elicitable with the zero-one loss, but no strict identification function is known.
For continuous response variables, the mode generally fails to be elicitable or identifiable \cite{HeinrichFissler2021}.

\subsubsection{Characterisation results}
(Strictly) consistent scoring functions for a functional are generally not unique.
One obvious fact is that if $S(z,y)$ is (strictly) consistent for $T$, then $\widetilde{S}(z,y) = \lambda S(z,y) + a(y)$ is also (strictly) consistent for $T$ if $\lambda>0$.
But the flexibility is usually much higher:
For example, under richness conditions on $\F$ and smoothness assumptions on the score, any (strictly) $\F$-consistent score for the mean takes the form of a Bregman function
\begin{equation} \label{eq:Bregman}
	S(z,y) = \phi(y) -\phi(z) + \phi'(z)(z-y) + a(y),
\end{equation}
where $\phi$ is (strictly) convex and $a$ is some function only depending on the observation $y$.
For $a=0$ and $\phi(z) = z^2$, this nests the usual squared error; see Table \ref{tab:scoring_functions} for some examples.
Note also that deviances of the exponential dispersion family are Bregman functions, see \cite{Saerens2000} and Eq.~(2.2.7) of \cite{Wuthrich2021StatisticalFoundations}. %\autoref{sec:scoring to deviance}.
An example of this fact is given in \autoref{sec:tweedie deviance} for the Tweedie familiy.
For the $\alpha$-quantile, under similar conditions, any (strictly) $\F$-consistent score takes the form
\begin{equation} \label{eq:piecewise linear}
	S(z,y) = (\one\{y\le z\} - \alpha)(g(z) - g(y)) + a(y),
\end{equation}
called generalised piecewise linear, where $g$ is (strictly) increasing \cite{gneiting_2011}.
The standard choice arises when $g(z) = z$ is the identity. 
Then \eqref{eq:piecewise linear} is often called piecewise linear loss, tick loss, or pinball loss.

There is an intimate link between strictly consistent scoring functions and strict identification functions, called Osband's principle \cite{Osband1985, gneiting_2011, FisslerZiegel2016}.
For any strictly $\F$-consistent score $S(z,y)$ and strict $\F$-identification function $V(z,y)$ there exists a function $h(z)$ such that
\begin{equation} \label{eq:Osband}
	\frac{\partial}{\partial z} \E[S(z,Y)] = h(z)\E[V(z,Y)]\qquad \text{for all }z\in \R, \ \text{for all }Y\sim F\in\F.
\end{equation}
Using the standard identification functions for the mean and the quantile, 
Osband's principle \eqref{eq:Osband} applied to \eqref{eq:Bregman} yields that $h(z) = \phi''(z)$, and applied to \eqref{eq:piecewise linear} it yields $h(z) = g'(z)$; see \autoref{tab:scoring_functions} for some specific examples.
Again, the mode functional constitutes an important exception.
Here, for categorical $Y$, any strictly consistent score is of the form 
\begin{equation} \label{eq:zero-one loss}
	S(z,y) = \lambda \one \{z\neq y\} + a(y), \quad \lambda>0,
\end{equation}
see \cite{Gneiting2017}.

\subsubsection{Score decompositions}
So far, we have motivated consistency in the setup without any feature information.
In the presence of feature information, however, \cite{HolzmannEulert2014} showed that a conditionally calibrated model $m_A(\bX)$ outperforms a conditionally calibrated model $m_B(\bX')$, where the latter one is based only on a subvector $\bX'$ of $\bX$.
This led to the observation that consistent scoring functions assess the information contained in a model and how accurately it is used \emph{simultaneously}.
It can be formalised in the following two versions of the calibration--resolution decomposition due to \cite{Pohle2020}:
\begin{align} \label{eq:score decomposition}
	\E [S(m(\bX),Y) ]
	&= \Big\{ \underbrace{\E [S(m(\bX),Y)] - \E [S(T(Y|\bX),Y) ]}_{\text{conditional miscalibration}} \Big\}\\ \nonumber
	&- \Big\{ \underbrace{\E [S(T(Y),Y)] - \E [S(T(Y|\bX),Y) ]}_{\text{conditional resolution\,/\,conditional discrimination}} \Big\} 
	+  \underbrace{\E [S(T(Y),Y)]}_{\text{uncertainty\,/\,entropy}} \\ \nonumber
	&= \Big\{ \underbrace{\E [S(m(\bX),Y)] - \E [S(T(Y|m(\bX)),Y) ]}_{\text{auto-miscalibration}} \Big\}\\\nonumber
	&- \Big\{ \underbrace{\E [S(T(Y),Y)] - \E [S(T(Y|m(\bX)),Y) ]}_{\text{auto-resolution\,/\,auto-discrimination}} \Big\} 
	+ \underbrace{\E [S(T(Y),Y)]}_{\text{uncertainty\,/\,entropy}}.
\end{align}
The term $\E [S(m(\bX),Y)] - \E [S(T(Y|\bX),Y) ]$ expresses the degree of \emph{conditional miscalibration} of $m(\bX)$ with respect to the information contained in the full feature vector $\bX$.
If the score is $\F$-consistent and assuming that the conditional distribution $F_{Y|\bX}$ is in $\F$ almost surely, this term is non-negative.
Under strict $\F$-consistency, it is \num{0} if and only if $m(\bX) = T(Y|\bX)$ almost surely.\footnote{%
This follows from an application of the definition of (strict) consistency \eqref{eq:consistent} using the conditional distribution $F = F_{Y|\bX}$.
Then, the expected score amounts to the conditional risk defined in \eqref{eq:conditional risk}.
}
The second term, $\E [S(T(Y),Y)] - \E [S(T(Y|\bX),Y) ]$, called \emph{conditional resolution} or \emph{conditional discrimination}, expresses the potential of a calibrated and informed prediction $T(Y|\bX)$ with respect to a calibrated, but uninformative prediction $T(Y)$.
Using the tower property of the conditional expectation and consistency, it is non-negative (see also \cite{HolzmannEulert2014}).
Under strict consistency, it is \num{0} if and only if $T(Y|\bX) = T(Y)$ almost surely.
Finally, $\E [S(T(Y),Y)]$ quantifies the inherent uncertainty of $Y$ when predicting $Y$ in terms of $T$. It is also called \emph{entropy} or \emph{Bayes risk} in the literature \cite{GR07}.
The second identity in \eqref{eq:score decomposition} constitutes a similar score decomposition, but now with respect to the auto-calibration.
Here, one is more modest in terms of the information set which is used.
It is generated by the model $m(\bX)$ itself (hence the term \enquote{auto}).
This contains generally less information than $\bX$ itself.%
\footnote{%
	As an example, consider a linear model where $\bX$ is high dimensional, but the model $m(\bX)$ is only one-dimensional.
}
Therefore, the first term $\E [S(m(\bX),Y)] - \E [S(T(Y|m(\bX)),Y) ]$, called \emph{auto-miscalibration}, assesses the deviation from perfect auto-calibration, while the second term $\E [S(T(Y),Y)] - \E [S(T(Y|m(\bX)),Y) ]$, called \emph{auto-resolution} or \emph{auto-discrimination}, is still a resolution term, quantifying the potential improvement in reducing uncertainty if the information in the model $m(\bX)$ had been used ideally.
This second decomposition reflects the general trade-off a modeller faces when deciding whether to incorporate an additional feature into the model in terms of out-of-sample performance.
On the one hand, including additional feature information increases the auto-resolution, but on the other hand, correctly fitting the model will become more difficult making it prone to a potential increase in auto-miscalibration.
The two decompositions \eqref{eq:score decomposition}  inform that minimising the expected score in the model $m(\bX)$ amounts to \emph{simultaneously improving calibration and maximising resolution} \cite{Pohle2020}.

To make the calibration--resolution decomposition \eqref{eq:score decomposition} tangible, 
we provide it for the squared error:
\begin{equation}\label{eq:squared score decomp}
	\E [(m(\bX) - Y)^2]
	= \underbrace{\E [(m(\bX) - \E[Y|\bX])^2]}_{\text{conditional miscalibration}}\quad 
	- \underbrace{\Var[\E [Y|\bX]]}_{\text{conditional resolution}}
	+  \underbrace{\Var[Y]}_{\text{uncertainty}}.
\end{equation}

The intuitive link between the conditional miscalibration term or the auto-mis\-ca\-li\-bra\-tion term on the one hand and the characterisation of conditional calibration \eqref{eq:cond cal} or auto-calibration \eqref{eq:auto cal} on the other hand can be made as follows:
Invoking the close connection between scoring functions and identification functions via Osband's principle \eqref{eq:Osband}, the respective miscalibration terms can be regarded as the antiderivative of the expected identification function $\E[V(m(\bX),Y)]$ in $m(\bX)$, with an additive term such that this antiderivative is non-negative and \num{0} if and only if the corresponding calibration property holds.
(Of course, formally this reasoning only holds for the case of constant  models and for auto-calibration / unconditional calibration.)

Practical estimation of $T(Y|m(\bX))$ and therefore of the auto variant of the decomposition can often be performed, see \cite{gneiting2021regression} or \autoref{fig:reliability diagrams}, while the estimation of the ultimate goal $T(Y|\bX)$ can be harder.
Hence, estimation of the miscalibration and resolution terms in the decompositions \eqref{eq:score decomposition} are challenging.
In any case, the score decompositions yield beneficial conceptional insights.
For practical checks of calibration, one should usually resort to the methods pointed out in \autoref{sec:Calibration and Identification Functions}.
However, since the unconditional functional $T(Y)$ and hence the uncertainty are relatively easy to estimate, a simplified decomposition consisting of miscalibration minus resolution\,/\,discrimination on the one hand and uncertainty on the other hand is feasible to estimate.
Therefore, \cite{gneiting2021regression} have promoted the ratio of these two constituents of the simplified score decomposition as a universal coefficient of determination (tacitly assuming that the scores are non-negative).
As an in-sample version, it generalises the classical coefficient of determination, $\Rtwo$, from OLS estimation.
The out-of-sample version of this universal coefficient of determination is often referred to as skill score.\footnote{%
	More generally, the skill score for model $m$ and scoring function $S$ is defined as relative improvement over a reference model $m_{\mathrm{ref}}$ as $\mathrm{score}_{\mathrm{skill}}(m; D) = \frac{\overline{S}(m; D) - \overline{S}(m_{\mathrm{ref}}; D)}{\overline{S}(m^\star; D) - \overline{S}(m_{\mathrm{ref}}; D)} \leq 1$ where $\overline{S}(m^\star; D) = 0$ for a suitable choice of the constant $a(y)$, see \cite{GR07}.
	It has therefore a reversed orientation, meaning larger values are better, and the optimal model achieves $\mathrm{score}_{\mathrm{skill}}(m^\star; D) = 1$.
}

We close the discussion by commenting on the differences of the calibration--resolution decomposition \eqref{eq:score decomposition} from the decomposition of the statistical risk in \eqref{eq:decomp stat risk}.
Upon identifying the loss function $L$ with the score $S$, the terms which are decomposed are the same: $R(m) = \E[S(m(\bX),Y)]$.
The two main differences are, however, that first the decomposition \eqref{eq:decomp stat risk} is concerned with estimation of the risk on a sample level while the decomposition \eqref{eq:score decomposition} stays on a population level.
Second, the baseline in \eqref{eq:score decomposition} is the entropy $\E[S(T(Y),Y)]$, which corresponds to the smallest risk in the class of trivial models $\{1\}\to\Y$.
In contrast, the baseline in \eqref{eq:decomp stat risk} is $\inf_{g: \X \to \Y} R(g)$, which corresponds to the term $\E[S(T(Y|\bX),Y)]$ in \eqref{eq:score decomposition}.

\subsubsection{Further topics}
The following five topics also have potential relevance in applications:
\begin{itemize}
	\item
	When aiming at estimating the expectation functional, often the root mean squared error $\mathrm{RMSE}(m; D) = \overline{S}(m; D)^{\frac{1}{2}}$ is reported and used for comparisons instead of the mean squared error (MSE) $\overline{S}(m; D) = \frac{1}{n} \sum_{(\bx_i, y_i) \in D} (m(\bx_i) - y)^2$ in order to have a score on the same unit as the response variable.
	Since taking the square root is a monotone transformation, RMSE and MSE induce the same model ranking.
	\item
	Absolute percentage error (APE) $S(z, y) = \abs{\frac{z - y}{y}}$ and relative error (RE) $S(z, y) = \abs{\frac{z - y}{z}}$, $z,y>0$, are two frequently used scores, in particular because they have no unit and can be reported as percentages.
	According to \cite{gneiting_2011}, Theorem~5 (Theorem~2.7 in arxiv version), they both are strictly consistent scoring functions for the $\beta$-median with $\beta=-1$ for the APE and $\beta=1$ for the RE.
	The $\beta$-median of a distribution $F$ with density $f$ is defined as the median of the re-weighted distribution $F_\beta$ with density proportional to $y^\beta f(y)$.
	\item
	One could be interested in a whole vector of different point predictions, resulting in a vector of functionals $(T_1,\ldots, T_k)$.
	Scoring functions for such pairs have been studied, \eg in \cite{FisslerZiegel2016}.
	Even though variance and expected shortfall fail to be elicitable, the pair of expected shortfall and the quantile at the same probability level admit strictly consistent scores, see \cite{FisslerZiegel2016} for details.
	Similarly, the pair (mean, variance) is elicitable.
	Scores of this pair are provided in \cite{10.1214/19-EJS1552} Eq.~(4.13).
	One particular example derived from the squared loss and the Poisson deviance is
	\begin{equation*}
		S(\mu, \sigma, y) = (\mu - y)^2 + y^2\log\frac{y^2}{\mu^2 + \sigma^2} + \mu^2 + \sigma^2 - y^2
	\end{equation*}
	with predicted mean $\mu$ and predicted standard deviation $\sigma$.
	Note that this is a homogeneous function of degree two, see \eqref{eq:homogeneous function}, and $S(\mu, \sigma, y) \ge S(y, 0, y) = 0$.

	It might also be interesting to consider pairs of functionals which are elicitable on their own.
	Two quantiles at different levels $\alpha<\beta$ induce a prediction interval with nominal coverage of $\beta - \alpha$.
	Scores for two quantiles can easily be obtained as the sum of two generalised linear losses \eqref{eq:piecewise linear}.
	For the case of a central $(1-\alpha)$-prediction interval, \ie when predicting two quantiles at levels $\frac{\alpha}{2}$ and $1-\frac{\alpha}{2}$, denoted $l$ (lower) and $u$ (upper), a standard interval score \cite{GR07} arises as
	\begin{equation*}
		S(l, u, y) = \frac{\alpha}{2} (u - l) + (l - y) \one\{y<l\} + (y - u) \one\{y>u\} \;.
	\end{equation*}
	The first term penalises large intervals (corresponding to bad discrimination), the other terms penalise samples outside the predicted interval (corresponding to miscalibration).
	For more details on evaluating prediction intervals, we refer to \cite{BrehmerGneiting2021, FFHR2021}.
	\item If the goal is a probabilistic prediction, \ie to predict the whole conditional distribution $F_{Y|\bX}$, then one uses \emph{scoring rules} to evaluate the predictions, see \cite{GR07, gneiting2014}.
	They are maps of the predictive distribution or density and the observation.
	Here, (strict) consistency is often termed (strict) propriety.
	It arises in \eqref{eq:consistent} upon considering the identity functional, $T(F) = F$.
	Standard examples are the logarithmic scoring rule and the continuously ranked probability score.
	\item
	The notion of consistency in Definition~\ref{defn:consistent} has its justification in a large sample situation, relying on a Law of Large Numbers argument to approximate expected scores.
	In a small sample setting it depends on how scores are used. 
	If modellers are rewarded for their prediction accuracy according to the average score their models achieve, consistency still has its justification in a repeated setting.
	If, on the other hand, the average score is only a tool to establish a ranking of different models, such that there is a winner-takes-it-all reward, one should care about the probability of achieving a smaller score than competitor models, $\P(\overline{S}(m_A; D) < \overline{S}(m_B; D))$. 
	This probability is not directly linked to the expected score and such schemes do generally not incentivise truthful forecasting \cite{WitkowskiETAL2021}.
	Hence, it provides an example where a criterion other than consistency is justified. 
	While this field is still subject to ongoing research, the  following \autoref{subsubsec:Neyman-Pearson} provides some links between winner-takes-it-all rewards and efficiency considerations in Diebold--Mariano tests.
\end{itemize}

\subsubsection{Role of data generating process for efficiency}
\label{subsubsec:Neyman-Pearson}
As discussed in \autoref{subsubsec:model comparison}, Diebold--Mariano tests for superior predictive performance of model $m_B$ over model $m_A$ formally test the null hypothesis $\E[S(m_A(\bX), Y) - S(m_B(\bX), Y) ] \le0$ for a scoring function $S$.
Without any further distributional assumptions on $(\bX,Y)$, one needs to invoke asymptotic normality of the average score differences and uses a $t$- or a $z$-test.
With this normal approximation, the power of such a test, \ie the probability for rejecting under the alternative, is the higher the smaller the ratio
\begin{equation*}
	\frac{\Var[S(m_A(\bX), Y) - S(m_B(\bX), Y) ]}{\E[S(m_A(\bX), Y) - S(m_B(\bX), Y) ]^2}
\end{equation*}
is. 
This ratio clearly depends on the true joint distribution of $(\bX,Y)$ and on the choice of $S$. 
While we are unaware of any clear theoretical results which $S$ maximises this efficiency, the simulation study in \autoref{sec:Efficiency aspects} and the following arguments provide evidence that scores derived from the negative log-likelihood of the underlying distribution behave advantageously.
The rationale is deeply inspired by the exposition of Section~3.3 in \cite{LerchETAL2017} highlighting the role of the classical Neyman--Person Lemma.

To facilitate the presentation, we omit the presence of features for a moment such that we only need to care about the distribution of $Y$.
Hence, the null simplifies to $H_0\colon \E[S(z_A, Y) - S(z_B, Y) ] \le0$ for two reals $z_A,z_B$.
Suppose $Y$ follows some parametric distribution $F_\theta$ with density $f_\theta$, $\theta\in\R$. 
Moreover, assume that on this family, the parameter corresponds to the target functional $T$, \ie $T(F_\theta) = \theta$ for all $\theta\in\R$. For example, $f_\theta$ could be the family of normal densities with mean $\theta$ and variance 1. 
Then one could also consider the simple test of the null $H'_0\colon Y_i\sim f_{z_A}$ against the simple alternative $H'_1\colon Y_i\sim f_{z_B}$.
The Neyman--Pearson Lemma then asserts that---under independence---the most powerful test at level $\alpha$ is based on the likelihood ratio $\prod_{i=1}^n f_{z_B}(y_i)/\prod_{i=1}^n f_{z_A}(y_i)$ and it rejects if this ratio exceeds some critical level $c_\alpha$, which is determined such that the probability of false rejection is exactly $\alpha$ under $H'_0$.

Suppose there exists a strictly consistent scoring function $S$ for $T$ such that for some function $a$ it holds that for all $\theta \in\R$, $y\in\Y$,
\begin{equation}
\label{eq:exp density}
	f_\theta(y) = \exp\left(-S(\theta,y) + a(y)\right).
\end{equation}
For the above example for normally distributed densities, $S$ could be half of the squared error.
Hence, the likelihood ratio exceeds $c_\alpha$ if and only if the log of the likelihood ratio, which corresponds to 
\begin{align*}
\log\left(\frac{\prod_{i=1}^n f_{z_B}(y_i)}{\prod_{i=1}^n f_{z_A}(y_i)}\right) 
= \sum_{i=1}^n \log\big(f_{z_B}(y_i)\big) -  \log\big(f_{z_A}(y_i)\big)
= \sum_{i=1}^n S(z_A,y_i) - S(z_B,y_i)
\end{align*}
exceeds $\log(c_\alpha)$. 
As a matter of fact, the decision based on the likelihood ratio is equivalent to a decision based on the empirical score difference, or equivalently on the average thereof.
 
Advantageously, this is not an asymptotic argument, but holds for any finite sample size $n$. 
On the other hand, the Neyman--Pearson Lemma is only applicable if the level of the test is exactly achieved under the null.
The correct critical value $c_\alpha$ can, however, only be calculated when exactly knowing $H'_0$ and $H'_1$.
And it is not clear per se if the test has the correct size under the broader null $H_0$.

\InsightsSec

\subsubsection{Which scoring function to choose?}
\label{subsubsec:choice of score}
In the presence of a clear purpose for a model, be it business, scientific or anything else, accompanied by an intrinsically meaningful loss function to assess the accuracy of this model, one should just use this very loss function for model ranking and selection.\footnote{%
	E.g.~the owner of a windmill might be interested in predictions for windspeed.
	Windspeed is directly linked to the production capability of electricity for the windmill.
	The owner uses this capability to enter a short-term contract on the electricity marked with explicit costs for over- or undersupply, resulting in an explicit economic loss for inaccurate forecasts.
}
In the absence of knowing such a specific cost function, but still knowing the modelling target $T$, one should clearly use a strictly consistent scoring function for the functional $T$.

But out of the (usually) infinitely many consistent scoring functions, which one to choose?
We see four different directions which might help in selecting a single one:
\begin{itemize}
	\item Different scoring functions have different domains for forecasts $z$ and observations $y$.
	For instance, the Poisson deviance can only be used for non-negative $y$ and positive $z$.
	This might at least exclude some of the choices.
	
	\item Scoring functions can exhibit beneficial invariance or equivariance properties. 
	One of the most relevant ones is positive homogeneity.
	This describes the scaling behaviour of a function if all its arguments are multiplied by the same positive number.
	The degree of homogeneity of the score $S$ is defined as the number $h$ that fulfils
	\begin{equation} \label{eq:homogeneous function}
		S(tz,ty) = t^h S(z, y) \quad \text{for all}\; t>0 \,\text{ and for all }z,y\,.
	\end{equation}
	Up to a multiplicative constant, the Tweedie deviances, see \autoref{tab:scoring_functions} and \autoref{sec:tweedie deviance}, are essentially the only homogeneous scoring functions that are strictly consistent for the expectation functional.
	The Tweedie deviance with power parameter $p$ has degree $h=2-p$, the Gaussian deviance (corresponding to the squared error) has $h=2$, the Poisson deviance $h=1$, the Gamma deviance $h=0$ and the inverse Gaussian deviance $h=-1$.
	This reflects the well-known fact that the squared error penalises large deviations of its arguments quadratically more than small ones.
	On the other hand, the Poisson deviance scales linearly with its arguments and preserves the unit.
	For instance, if $z$ and $y$ are amounts in some currency like CHF, the unit of Poisson deviance is CHF, too.
	The scale-invariant Gamma deviance only measures relative differences and has no notion of scales or units.
	As a 0-homogeneous score it can downgrade the (conditional) heteroskedasticity in the data (volatility clusters etc.).
	Lastly, larger values of the Tweedie power correspond to heavier-tailed distributions (still, all moments are finite).
	The larger the Tweedie power, the smaller the degree of homogeneity of the corresponding deviance, which puts less and less emphasis on the deviation of large values and more and more weight on deviations of small values.\footnote{%
		Consider the inverse Gaussian deviance with $p=3$ and degree of homogeneity $h=-1$, $S(z, y) = \frac{(z - y)^2}{z^2y}$. The score $S(2, 1) = \frac{1}{4}$ is ten times larger than the score $S(20, 10) = \frac{1}{40}$.
	}
	\item 
	Another aspect in the choice of the scoring function is the efficiency in test decisions for predictive dominance, see \autoref{subsubsec:Neyman-Pearson}. 
	Scoring functions based on the negative log-likelihood of the conditional distribution of $Y$ given $\X$ appear to be beneficial. 
	E.g.~if this distribution is a certain Tweedie distribution, the corresponding Tweedie deviance given in \autoref{tab:scoring_functions} is appropriate.
	See \autoref{sec:Efficiency aspects} for a corresponding simulation study.
	\item 
	In order to assess the ranking of two models with respect to \emph{all} consistent scoring functions, one can exploit the mixture representations of consistent scores established for quantiles and expectiles in \cite{ehm2015quantiles} and visually assess the dominance in terms of Murphy diagrams, see \autoref{fig:murphy diagrams}.
\end{itemize}

\subsubsection{Sample weights and scoring functions}
For model fitting, a priori known sample weights $w_i > 0$, also known as case weights and---in the actuarial literature---exposure or volume, are a way to account for heteroskedasticity and to improve the speed of convergence of the estimation.
A simple example is given by the weighted least squares estimation in comparison to OLS estimation.
More generally, estimating the conditional expectation via maximum likelihood with a member of the exponential dispersion family amounts to minimising the weighted score
\begin{equation*}
	\frac{1}{\sum_i w_i} \sum_i w_i S(m(\bx_i), y_i)
\end{equation*}
for scoring functions $S$ of the Bregman form, see \eqref{eq:Bregman} and comments thereafter.

For fitting as well as for model validation, it is noteworthy that using weights often changes the response variable.
Take as an example the actuarial task of modelling the average claim size, called severity.
Let $w_i$ denote the number of claims of a policy $i$ with features $\bX_i$, where $w_i$ is given in the feature vector, which happened in a certain amount of time, and $C_i$ denotes the corresponding sum of claim amounts. 
We define the severity as $Y_i = \frac{C_i}{w_i}$.
Then, the response modelled and predicted by $m(\bX_i)$ is $Y_i$ and not $C_i$, with the important assumption that $\Var[Y_i|\bX_i] \propto \frac{1}{w_i}$.\footnote{%
	This can be seen as follows.
	We use the $\alpha$-$\beta$--parametrisation of the Gamma distribution, such that $Z \sim \mathrm{Ga}(\alpha, \beta)$ has $\E[Z] = \frac{\alpha}{\beta} = \mu$ and $\Var[Z] = \frac{\alpha}{\beta^2} = \phi \mu^2$ upon setting $\phi = \frac{1}{\alpha}$.
	A typical assumption is that $C_i$ is a sum of $w_i$ independent $\mathrm{Ga}(\alpha, \beta_i)$-distributed claim amounts each with mean $\mu_i$ and variance $\phi \mu_i^2$, such that $C_i | \bX_i \sim \mathrm{Ga}(w_i \alpha, \beta_i)$. Then, the severity is also Gamma distributed, $Y_i | \bX_i \sim \mathrm{G}(w_i \alpha, w_i \beta_i)$, with mean $\E[Y_i|\bX_i] = \frac{\alpha}{\beta_i} = \mu_i$ and variance $\Var[Y_i|\bX_i] = \frac{1}{w_i} \phi \mu_i^2$.
}

Back to our main purpose of estimating the expected score $\E[S(m(\bX), Y)]$ based on an \iid test sample $(\bX_i,Y_i)_{i=1,\ldots,n}$.
We assume that $\E[S(m(\bX),Y)|\bX] = \mu$ for some constant $\mu$ and heteroskedasticity in the form of a conditional variance $\Var[S(m(\bX), Y)| \bX] = \sigma^2 h(\bX)$ for some $\sigma^2>0$ and some positive and measurable function $h\colon\mathcal X\to(0,\infty)$.
Again, consider the weighted score average 
\[
\widetilde{S}_n = \frac{1}{\sum_i w_i} \sum_{i=1}^n w_i S(m(\bX_i), Y_i).
\]
Here, each weight $w_i$ is a function of the feature $\bX_i$. 
For the case that all weights are the same, one simply retrieves the average score $\overline{S}_n$.
Just like $\overline{S}_n$, the weighted average $\widetilde{S}_n$ is an unbiased estimate of the expected score. 
(This is achieved by the normalisation. 
Simply consider $\E[\widetilde{S}_n|\bX_1,\ldots, \bX_n] = \mu$.)
Thanks to the total variance formula,\footnote{%
	It asserts that $\Var[Z] = \E[\Var[Z|\bX]] + \Var[\E[Z|\bX]]$.
}
we can calculate the variance of $\widetilde{S}_n$ as 
\begin{equation}
\label{eq:variance}
\Var[\widetilde{S}_n]
= \Var[\widetilde{S}_n|\bX_1,\ldots, \bX_n]
= \Big(\sum_{i=1}^n w_i\Big)^{-2}\sigma^2 \sum_{i=1}^n w_i^2 h(\bX_i) \,.
\end{equation}
Minimising \eqref{eq:variance} with respect to the weights yields that any solution can be written as
$w_i = \lambda / h(\bX_i)$ for some constant $\lambda>0$.\footnote{%
This can be easiest derived by a constraint optimisation with the Lagrange method, where \eqref{eq:variance} is minimised subject to $\sum_{i=1}^n w_i = c$ for some constant $c>0$.)
}

The major drawback with this re-weighting approach in practice is that the conditional variance of the scores is usually unknown.
Even if the conditional variance of the response variable is known or assumed, as in maximum likelihood estimation, the authors are not aware of a general link to the level of the score.
Whether weighted averages with weights accounting for heteroskedasticity of the response provide more efficient estimates for the expected score seems to be an open question.
That being said, weighted averages of scores are nonetheless unbiased.

\ExampleSec{Mean Regression for Workers' Compensation.}

\subsubsection{Model comparison with Gamma deviance}
\begin{figure}[p]
	\centering
	\includegraphics[width=.90\textwidth]{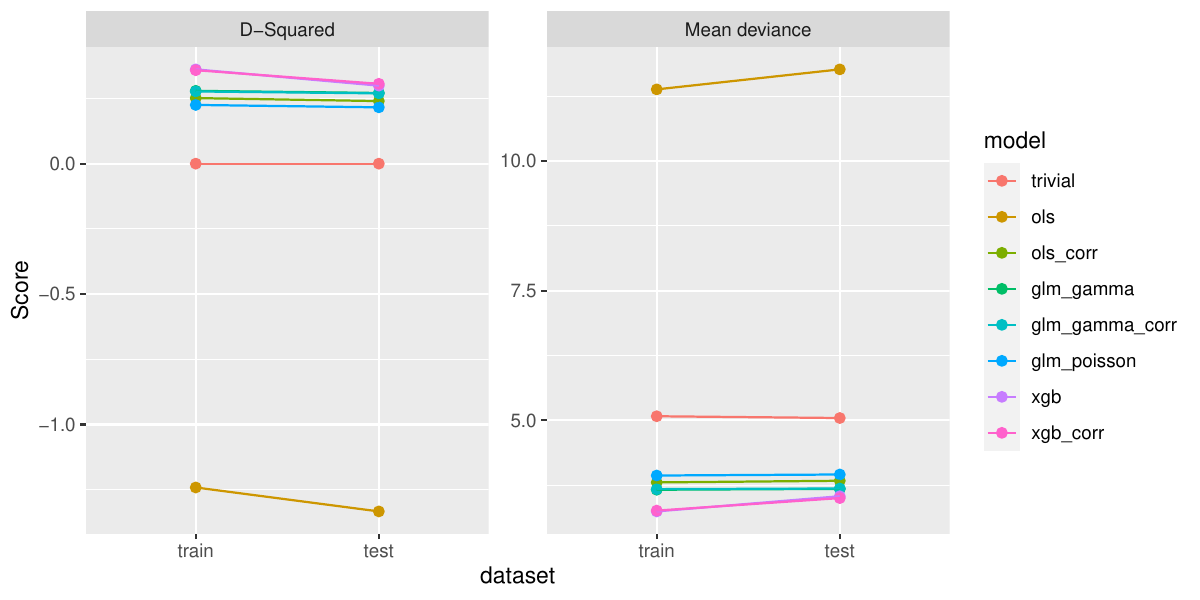}
	\caption{Performance on training and validation data in terms of mean Gamma deviance (smaller is better) and its relative reduction $\Rstar$ (larger is better).}
	\label{fig:performance regression}
\end{figure}
As stated in the \autoref{subsec:statistical learning theory.Example}, we evaluate our models with the Gamma deviance, see \autoref{tab:scoring_functions} and Equation~\eqref{eq:empirical score}.
The reasons for this choice follow the discussion of \autoref{subsubsec:choice of score}:
\begin{itemize}
	\item
	The target functional is the expectation and the Gamma deviance is strictly consistent for it.
	\item
	Observations and predictions are positive, thus falling within the domain of the Gamma deviance.
	\item
	The 0-homogeneity of the Gamma deviance implies that the induced model ranking is independent of the unit of the response (currency in our case).
	\item
	The response is right-skewed. 
	The Gamma distribution is a good candidate distribution for such data within the class of Tweedie distributions.
	In addition, only Tweedie deviances provide homogeneous scores.
\end{itemize}

In analogy to the coefficient of determination, $\Rtwo$, we additionally report the (relative) Gamma deviance reduction $\Rstar = 1 - \frac{\overline{S}(m; D)}{\overline{S}(m_\text{trivial}; D)}$, which is closely related to McFadden's pseudo $\Rtwo$ and which corresponds to the universal coefficient of determination coined in \cite{gneiting2021regression}.
It holds that $\Rstar \le 1$, where a larger value indicates a better fit.
\autoref{tab:performance regression} displays the scores on the training as well as on the test set.
\autoref{fig:performance regression} visualises these numbers.
\begin{table}[p]%[hbt]
	\centering
	\begin{tabular}{lrrrr}
		\toprule
		Model   &       $\Rstar$ train & $\Rstar$ test & Mean deviance train & Mean deviance test \\
		\midrule
		Trivial        &    \num{0    } &         \num{0    }  & \num{5.08} &         \num{5.04} \\
		OLS            &    \num{-1.24} &         \num{-1.33}  & \num{11.4} &         \num{11.8} \\
		OLS corr       &    \num{0.251} &         \num{0.240}  & \num{3.80} &         \num{3.83} \\
		GLM Gamma      &    \num{0.279} &         \num{0.271}  & \num{3.66} &         \num{3.68} \\
		GLM Gamma corr &    \num{0.277} &         \num{0.270}  & \num{3.67} &         \num{3.68} \\
		GLM Poisson    &    \num{0.225} &         \num{0.216}  & \num{3.94} &         \num{3.95} \\
		XGB            &    \num{0.362} &         \num{0.299}  & \num{3.24} &         \num{3.54} \\
		XGB corr       &    \num{0.358} & \textbf{\num{0.306}} & \num{3.26} & \textbf{\num{3.50}} \\
		\bottomrule
	\end{tabular}
	\caption{Performance on training and validation data in terms of mean Gamma deviance (smaller is better) and its relative reduction $\Rstar$ (larger is better).}
	\label{tab:performance regression}
\end{table}
\begin{table}[p]%[hbt]
	\centering
	\begin{tabular}{lrrrr}
		\toprule
		Model   &       Mean deviance & Auto-miscalibration & Auto-resolution & Uncertainty \\
		\midrule
		Trivial        &         \num{5.04}  &         \num{0    }  &         \num{0   }  &    \num{5.04} \\
		OLS            &         \num{11.8}  &         \num{8.25 }  &         \num{1.52}  &    \num{5.04} \\
		OLS corr       &         \num{3.83}  &         \num{0.314}  &         \num{1.52}  &    \num{5.04} \\
		GLM Gamma      &         \num{3.68}  &         \num{0.190}  &         \num{1.56}  &    \num{5.04} \\
		GLM Gamma corr &         \num{3.68}  &         \num{0.197}  &         \num{1.56}  &    \num{5.04} \\
		GLM Poisson    &         \num{3.95}  &         \num{0.482}  &         \num{1.57}  &    \num{5.04} \\
		XGB            &         \num{3.54}  &         \num{0.124}  &         \num{1.63}  &    \num{5.04} \\
		XGB corr       & \textbf{\num{3.50}} & \textbf{\num{0.088}} & \textbf{\num{1.63}} &    \num{5.04} \\
		\bottomrule
	\end{tabular}
	\caption{Decomposition of the Gamma deviance on the test set into auto-miscalibration (smaller is better), auto-resolution (larger is better) and uncertainty.}
	\label{tab:score decomposition regression}
\end{table}
The best out-of-sample performance is achieved by the corrected XGBoost with a test $\Rstar$ of \num{0.308}.
We could use a $t$-test to check whether it is significantly better than the second best model, but refrain here to do so.
We also observe for all models that the in-sample $\Rstar$ is better (higher) than the out-of-sample $\Rstar$, reflecting the general difficulty to generalise to new data.
According to our definition of overfitting in Definition~\ref{defn:overfitting}, the uncorrected XGBoost model overfits the training data.
Indeed, the corrected XGBoost model exhibits a smaller complexity (the correction acts as additional constraint and reduces the degrees of freedom in fitting by one) and it holds that
$\overline{S}(m_{\text{xgb}}; D_{\text{train}}) < \overline{S}(m_{\text{xgb\_corr}}; D_{\text{train}})$, 
while $\overline{S}(m_{\text{xgb}}; D_{\text{test}}) > \overline{S}(m_{\text{xgb\_corr}}; D_{\text{test}})$.
Another observation is that both Gamma GLMs perform better in terms of Gamma deviance than the Poisson GLM.
This is to be expected as the Gamma GLMs minimise the Gamma deviance on the training set, while the training objective for the Poisson GLM is the Poisson deviance.
Similar to our finding for calibration, the bare OLS shows very poor performance, even negative $\Rstar$, indicating worse predictive performance than the trivial model.
Again, this is the consequence of modelling on the log transformed response variable and then backtransforming for prediction.
On the other hand, the corrected OLS performs even better than the Poisson GLM.
Regarding the correction term for the Gamma GLM and the XGBoost model, it slightly reduces the performance on the training set.
On the test set, however, there is hardly a penalty left.
On the contrary, the XGBoost out-of-sample performance is even improved by it.

For a more detailled inspection, we add the score decomposition of the Gamma deviance according to the auto-variant of \eqref{eq:score decomposition} in \autoref{tab:score decomposition regression}.
We observe that the uncertainty term is by far the largest component.
The model ranking is then dominated by the auto-resolution term (larger is better).
Clearly, the XGBoost models make most use of the information contained in themselves.
The only exception is again the OLS model, whose auto-miscalibration component is even larger than the uncertainty term.
Interestingly, auto-miscalibration is the decisive component for the ranking within the different GLMs, with the Gamma GLMs being better auto-calibrated than the Poisson GLM.
To the largest part, these findings are in line with \autoref{tab:auto-calibration}.

To summarise the results for our models: Overall, the corrected XGBoost model seems to be the best calibrated one and has the best out-of-sample performance.
We are confident to have mitigated the risk of overly optimistic scores with the evaluation on an independent test set.
The corrected Gamma GLM shows better performance than the Poisson GLM, at the cost of a slightly worse calibration conditional on \variable{Gender}, \cf \autoref{tab:conditional calibration by Gender}, and on \variable{LogWeeklyPay}, \cf \autoref{tab:conditional calibration test funcion}.

\subsubsection{Murphy diagram}
\begin{figure}
	\centering
	\begin{subfigure}[t]{1.0\textwidth}
		\includegraphics[width=.90\textwidth]{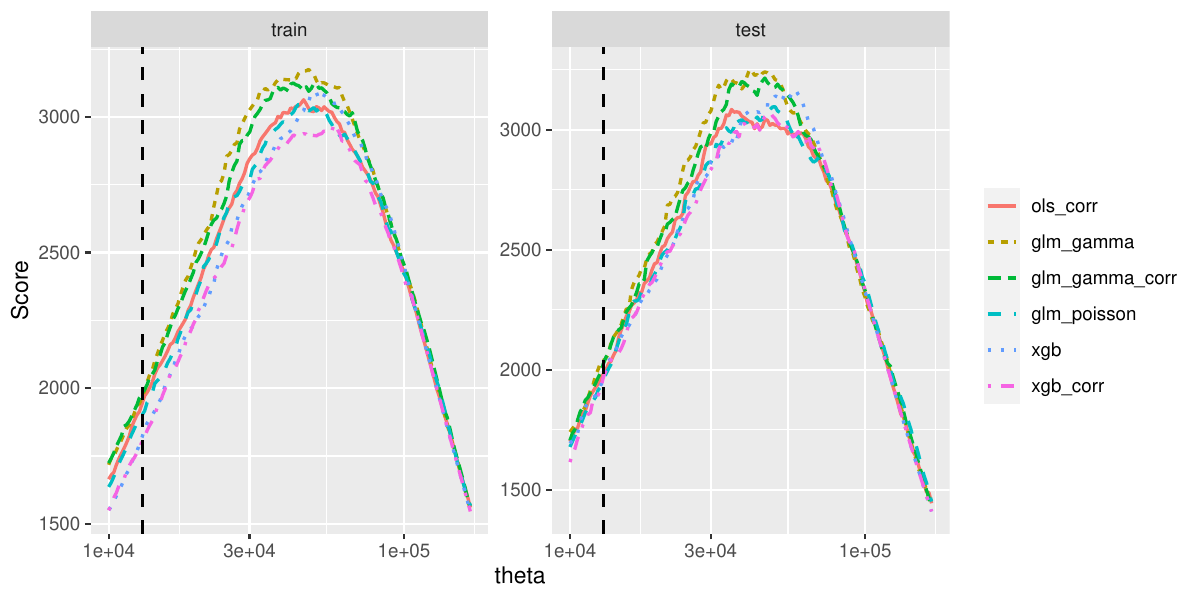}
		\caption{Murphy diagram with elementary scores for $\theta$ from \num{1e4} to \num{1.7e5} (instead of $\min(y)$ to $\max(y)$, for better readability).
		The black vertical line indicates the average response on the training data set.}
		\label{fig:murphy diagram}
	\end{subfigure}
	\par\bigskip
	\begin{subfigure}[b]{1.0\textwidth}
		\includegraphics[width=.90\textwidth]{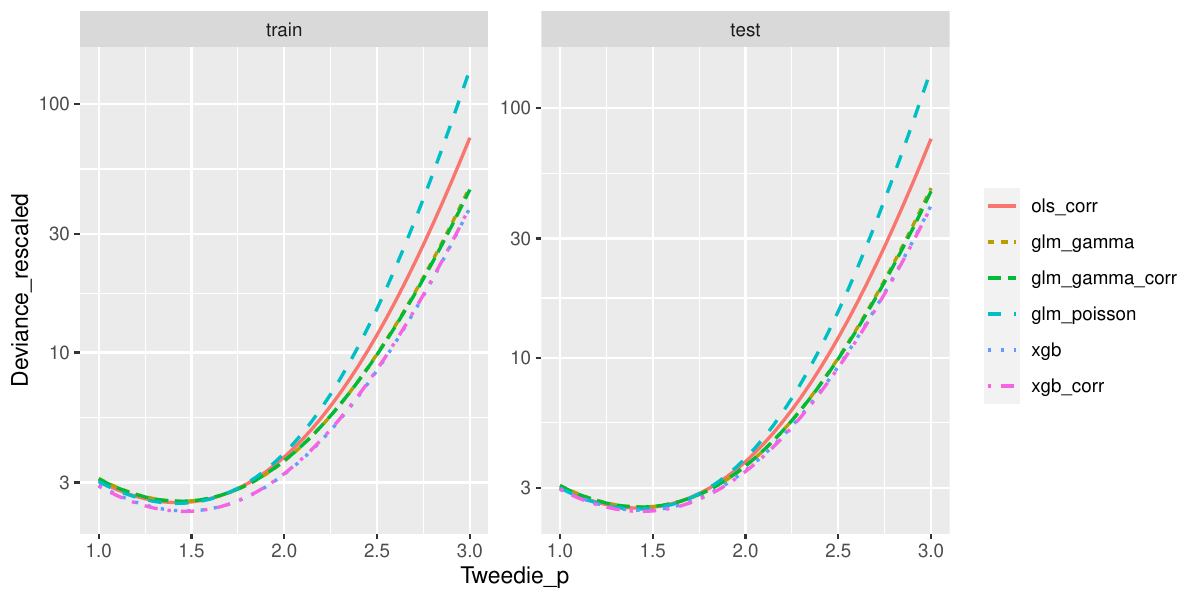}
		\caption{Murphy-like diagram with Tweedie deviances across range of Tweedie powers.
		For better readability, the $y$-axis shows a rescaled version of the Tweedie deviance: $d_p \cdot \bar{y}^{p-2}$ with the mean response value $\bar{y}$.}
		\label{fig:tweedie diagram}
	\end{subfigure}
	\caption{Murphy diagrams}
	\label{fig:murphy diagrams}
\end{figure}
How sensitive are our results with respect to the choice of the scoring function?
We show Murphy diagrams \cite{ehm2015quantiles}, which depict different scores along a parameter.
The standard way is to use the elementary scoring function, for the expectation given by $S_\theta(z, y) = \frac{1}{2}\abs{\theta - y}\one\{\min(z, y) \leq \theta < \max(z, y)\}$, and to vary its parameter $\theta$, see \autoref{fig:murphy diagram}.
As we have worked with and emphasised the Tweedie deviances, we show the same type of diagram with varying Tweedie power parameter $p$ in \autoref{fig:tweedie diagram}.
We have left out the trivial model and the OLS for better readability of the charts.
One sees that the corrected XGBoost model performs best on a large range of scores, indicating that this model dominates its competitors.
This is in contrast to other models like the Poisson GLM which has good scores for Tweedie powers $p$ close to 1, corresponding to the Poisson deviance, but gets worse for larger $p$.

\FloatBarrier
\section{Probabilistic Binary Classification} \label{sec:probabilistic binary classification}

Binary classification might be the most common task in ML, be it in theory, in ML challenges or in real world applications.
While all the above theory parts fully apply here as well, we still want to highlight some special features and to present an example on probabilistic binary classification.

The term classification usually means predicting one discrete class out of a fixed set of $k$ classes, which can be identified with $\Y = \{0, \ldots, k-1\}$ upon renaming the class labels.
Examples for such classes are client states like \enquote{healthy}, \enquote{ill}, \enquote{dead} or \enquote{lapsed} and \enquote{renewed}.
Here, we use the term probabilistic classification to underpin that we mean predicting probabilities per class.
Thus, probabilistic classification amounts to probabilistic prediction for discrete response values.

In the sequel, we focus on the most relevant situation of $k=2$ classes, called binary classification.

\begin{summarybox*}{Binary Classification}
	\begin{itemize}
		\item The response $Y$ takes only the values $0$ and $1$.
		\item The probability for class $1$ is the expectation $\P(Y=1|\bX) = \E[Y|\bX]$.
		\item Probabilistic classification models $z = m(\bX)$ predicting the class probability are models for the conditional expectation with canonical identification function $V(z,y) = z - y$, \cf \autoref{tab:identification_functions}.
		Strictly consistent scoring functions are Bregman functions \eqref{eq:Bregman binary}.
		\item Deterministic classification, \ie predicting one class only, is decision making.
		Ideally, decisions ought be made based on the full information of $\P(Y=1|\bX)$, or, at least, on a good prediction thereof.
	\end{itemize}
\end{summarybox*}

\TheorySec

In probabilistic binary classification, the outcome space $\Y$ consists of dichotomous events, leading to $\Y = \{0,1\}$.
If we assign $p = \P(Y=1|\bX)$ and $1 - p = \P(Y=0|\bX)$, we immediately obtain $p = \E[Y|\bX]\in[0,1]$.
This shows the well-known fact that the entire (conditional) distribution of $Y$ is solely parametrised by the expectation of $Y$, which is a number in the interval $[0,1]$.
Hence, probabilistic modelling and prediction amounts to modelling the (conditional) mean of $Y$.
This is in sharp contrast to continuous responses where probabilistic predictions are much harder than point predictions.
It also means that the notions of scoring functions (for the expectation) and scoring rules (for the probability distribution) coincide, \cf \cite{GR07, gneiting2014, Buja2005LossFF}.

\subsubsection{Probabilistic classification models}
Probabilistic models make \emph{probabilistic predictions} $m(\bx) \in [0,1]$.
The ideal model is
\begin{equation*}
	m^\star(\bx) = \P(Y=1|\bX=\bx)=\E[Y|\bX=\bx]\,.
\end{equation*}
Since this conditional probability actually corresponds to a conditional expectation, any strictly consistent scoring function takes the form \eqref{eq:Bregman}, which is
\begin{equation} \label{eq:Bregman binary}
	S(z,y) = \phi(y)- \phi(z) + \phi'(z)(z- y) + a(y), \quad z\in [0,1], \ y\in\{0,1\}\,,
\end{equation}
where $\phi$ is some strictly convex function on $[0,1]$.

In this context, the squared error, arising from $\phi(z) = z^2$ and $a=0$, is called Brier score.
Its auto-resolution-miscalibration decomposition is referred to as Brier score decomposition \cite{ANewVectorPartitionoftheProbabilityScore} and given in \eqref{eq:squared score decomp} upon replacing the conditioning on $\bX$ by conditioning on $m(\bX)$.
The most prominent strictly consistent score for binary classification is the log loss
\begin{equation}
	\label{eq:log loss}
	S_{\log}(z, y) = - y \log(z) - (1-y) \log(1-z) \,,
\end{equation}
see \autoref{tab:scoring_functions}.\footnote{%
	The log loss is known under many different names: logarithmic score, log loss, logistic loss, Bernoulli or Binomial deviance, Bernoulli log-likelihood.
	It gives rise to the Kullback-Leibler divergence, the Shannon entropy, and the binary cross-entropy.\\
	It arises from \eqref{eq:Bregman binary} upon using the strictly convex function
	$\phi(z) = z\log(z) + (1-z)\log(1-z)$, $z\in(0,1)$, with its continuous extension to $[0,1]$ such that $\phi(0) = \phi(1) = 0$, and $a(y) = -\phi(y)$.
}
In contrast to the Brier score, the log loss penalises false certainty, \ie $z=0$ while $y=1$ or $z=1$ while $y=0$, with an infinite score.

\subsubsection{Deterministic classification: decision rule}	
The alternative to probabilistic models are models that make discrete predictions $m(\bx) \in \{0, 1\}$, called decision rule or simply zero-one prediction.
Hence, the target functional needs to attain values in $\{0, 1\}$ only, with the most important case being a quantile.
In a business context, this can easily be motivated as follows.
For a decision maker, the cost $c_0$ of false positives (predicting $m(\bX)=1$ while $Y=0$ materialises) and the cost $c_1$ for false negatives (predicting $m(\bX)=0$ while $Y=1$ materialises) can be very different and very much context dependent, \cf
\cite{ThompsonBrier1955,
%Murphy1966CostLossRatio
 Elkan2001Foundations, Granger1999EconomicAS}.\footnote{%
	An example is fraud detection for insurance claims.
	Typically, the cost of missing a fraud case (false negatives) is much higher than the cost for wrongly predicting fraud (false positives).
}
With cost ratio $c = \frac{c_0}{c_0 + c_1} \in [0, 1]$, this leads to the cost-weighted misclassification error:\footnote{%
	In this context of decision theory, such loss or cost functions are called utility.
}
\begin{equation*}
	S_c(z, y) = (1-c) y (1-z) + c (1-y) z = (\one\{z \geq y\} - (1-c))(z-y)\,, \quad z,y\in\{0,1\}\,.
\end{equation*}
It turns out that this is the pinball loss of \autoref{tab:scoring_functions} for the $\alpha$-quantile with $\alpha=1-c$ and optimal predictions are the $\alpha$-quantiles of $\P(Y=1|\bX)$ given by
\begin{equation} \label{eq:Bayes classifier}
m^\star(x) =
\begin{cases}
0, & \P(Y=1\,|\,\bX=\bx) < c \\
1, & \P(Y=1\,|\,\bX=\bx) > c \\
0 \text{ or } 1, & \P(Y=1\,|\,\bX=\bx) = c
\end{cases}
\,,
\end{equation}
where the threshold $c$ is identical to the cost ratio.
This representation reveals that deterministic classification is only concerned about the (proximity of the) decision boundary $\P(Y=1|\bX) = c$ in contrast to probabilistic classification.

In light of \eqref{eq:Bayes classifier}, the cost-weighted misclassification error is often written in terms of probabilistic predictions $z \in [0, 1]$ where the thresholding is made explicit
\begin{equation*}
	S_c(z, y) = y (1-c) \cdot \one\{z \leq c\} + (1-y) c \cdot \one\{z > c\}\,, \quad y \in \{0, 1\}, \ z \in [0, 1]\,.
\end{equation*}

In the case of threshold $c=\frac{1}{2}$, \ie equal cost for false positives and false negatives, the ideal model is the median of the distribution of $Y$, conditional on $\bX$.
For a dichotomous response variable $Y$, the conditional median equals the conditional mode.
Then, the only strictly consistent scoring function for the mode---up to equivalence---is given by
\begin{equation*}
	S(z,y) = 2 S_{\frac{1}{2}}(z, y) = \one\{z\neq y\} = \abs{z - y} \,, \quad z,y\in\{0,1\}\,,
\end{equation*}
which is the zero-one loss of \eqref{eq:zero-one loss}, \cite{Gneiting2017}, also known as $1 - \text{accuracy}$.

\InsightsSec

As already outlined in the introduction \textcquote{gneiting_2011}{\textins{i}deally, forecasts ought to be probabilistic, taking the form of predictive distributions over future quantities and events.}
For binary responses, probabilistic forecasts translate to predicting $\P(Y=1|\bX)=\E[Y|\bX]$ with the help of which an informed decision can be made.
In contrast to continuous responses, probabilistic forecasts for binary events are much simpler as they are just point forecasts for the conditional expectation.
Disadvantages of deterministic classifiers that directly predict a discrete class are that they lose a lot of information and prematurely make the decision in choosing a threshold $c$ in \eqref{eq:Bayes classifier} or a loss function like the zero-one loss.
Further arguments in favour of probabilistic classification can be found in Chapter~1.3 of \cite{harrell2015}.
To conclude this point: \textcquote{MurphyWinkler1984}{%
	\textins{D}eterministic forecasts suffer from two serious deficiencies.
	First, a forecaster or forecast system is seldom, if ever, certain which event will occur.
	Second, categorical forecasts do not provide users of the forecast with the information that they need to make rational decisions in uncertain situations.%
}
Therefore, without use case specific reasons to do otherwise, we generally recommend the use of calibrated probabilistic classifiers, in particular for actuarial applications.

\subsubsection{Miscellaneous scores}
\begin{itemize}
	\item Accuracy: $S(z, y) = \one\{z=y\}$ for $z\in\{0,1\}$ or $z\in[0,1]$.\newline
	As stated above, the accuracy, is $1 - \text{zero-one loss}$.
	Hence, upon reversing the orientation of strict consistency (meaning that larger values are preferable), it is strictly consistent for the median\,/\,mode.
	As such, it fails to be a strictly proper scoring rule, even if $z$ attains values in $[0,1]$.
	One advantage of accuracy is the ease of communication and the direct interpretability of the average accuracy of a model, attaining values in $[0,1]$.
	However, we would like to caution such a use.
	Scoring functions are primarily tailored to forecast \emph{comparison}.
	The calibration--resolution decomposition \eqref{eq:score decomposition} underpins the dependence of the expected or average score on the unconditional distribution of $Y$ via the uncertainty term.
	For binary responses, the more imbalanced the unconditional distribution is, the smaller the uncertainty.
	That means it is \enquote{easier} to make accurate predictions without making use of the information contained in the features at all.
	If, \eg $\P(Y=1)=0.95$, a calibrated trivial model maximising accuracy always predicts \num{1} such that an accuracy of \num{0.95} is achieved by this trivial model.
	This highlights the importance of reporting skill scores\,/\,universal coefficients of determination.
	
	\item Absolute loss $S(z, y) = \abs{z - y}$ for $z\in\{0,1\}$ or $z\in[0,1]$.\newline
	Since the absolute loss is strictly consistent for the median and, for binary responses, the median equals the mode, the same comments as for accuracy apply here.
	(However, with the usual sign convention that smaller scores reflect a better predictive performance.)
	
	\item Hinge loss $S(z, y) = \max(0, 1 - \tilde{y}z)$ with $\tilde{y} = 2y -1 \in\{-1, +1\}$ for $z\in[-1,1]$ or $z\in\R$.\newline
	The hinge loss is commonly used to fit support vector machines and deterministic classification models.
	It is minimised by 
	\begin{equation*}
		m^\star(x) =
		\begin{cases}
			\le -1, & \P(Y=1\,|\,\bX=\bx) < 1/2 \\
			\ge 1, & \P(Y=1\,|\,\bX=\bx) > 1/2 \\
			\le -1 \text{ or } \ge 1, & \P(Y=1\,|\,\bX=\bx) = \frac{1}{2}
		\end{cases}
		\,,
	\end{equation*}
	We see that the additional flexibility issuing predictions on the whole real line is not really essential.
	The best action is actually one-to-one with \eqref{eq:Bayes classifier} for $c=1/2$.
	The best model aims at modelling the conditional mode.
	
	\item Confusion matrix related scores for deterministic classification.\newline
	There is a wealth of scores derived from the confusion matrix, also known as binary contingency table, see \cite{Powers2007evaluation, Tharwat2020ClassificationAM} and \cite[Section 2]{Wilks2011}.
	Next to accuracy---and again being flexible about the orientation of the scores---established ones are the hit rate (HR)\footnote{%
		HR is also know as sensitivity, recall and true positive rate.
	} (number of observations correctly predicted as positive, $z=y=1$, over number of positive outcomes, $y=1$) and false alarm rate (FAR)\footnote{%
		FAR is also known as probability of false detection, fall-out and false positive rate.
		It equals minus specificity, selectivity, or true negative rate
	} (number of observations falsely predicted as positive, $z=1$ while $y=0$, over number of negative outcomes, $y=0$).
	On a population level, the hit rate describes the conditional probability
	$\P(m(\bX)=1|Y=1)$ and the false alarm rate describes $\P(m(\bX)=1|Y=0)$.
	\newline
	A straightforward application of \cite[Theorem~5]{gneiting_2011} yields that HR is optimised by the constant model $m^\star(\bx)=1$, and equally FAR is optimised by $m^\star(\bx)=0$.
	Therefore, they are strictly consistent scoring functions for trivial (and uninteresting) functionals, not taking into account the distribution of the response variable.
	Hence, while HR, FAR and further variations (\eg precision, $F_1$-score, \etc) have their place in diagnostics, reporting and decision making, they are unfit for model comparison.
	
	\item Area under the curve (AUC) for probabilistic forecasts $z\in\R$.\newline
	AUC is defined as the area under the ROC curve, see further below and \cf \cite{Fawcett2006, Byrne2016, gneiting2021receiver}.
	It has a value between 0 and 1 with the orientation the higher the better.
	Note, however, that a random guessing classifier has $\mathrm{AUC}=0.5$ such that sensible classifiers should have $\mathrm{AUC} > 0.5$.
	While the AUC can be appealing when it comes to diagnostics, its deficiency is that it ignores calibration of the forecast, basically only assessing its resolution or discrimination, \cite[Section 4.7]{Wilks2011}.
	In particular, equation (3.4) in \cite{Byrne2016} illustrates that the AUC ignores the marginal distribution of $Y$.
	Hence, it cannot be a proper scoring rule for a binary response.	
	This coincides with the findings of \cite{Byrne2016}, who also formulated partial exceptions where propriety holds.
\end{itemize}

\subsubsection{Graphical tools}
\begin{itemize}
	\item Reliability diagrams\newline
	A good diagnostic tool for assessing auto-calibration of binary responses are reliability diagrams \cite{DimitriadisGneitingJordan2021} which draw observed outcomes vs predicted values.
	\textcquote{DimitriadisGneitingJordan2021}{%
	In a nutshell, \textelp{}
	\textins{a probabilistic} classifier is calibrated or reliable if, when looking back at a series of extant forecasts, the conditional event frequencies match the predictive probabilities.%
	}
	\item Receiver operating characteristic (ROC)\newline
	The ROC is a graphical tool illustrating the trade-off when choosing different cost ratios or thresholds $c$, see \cite{Fawcett2006}.
	\newline
	Let $m(\bX)\in[0,1]$ be a probabilistic model.
	Following \cite{gneiting2021receiver}, the ROC curve is a plot of the hit rate against the false alarm rate.
	To that end, one first turns the probabilistic forecast $m(\bX)$ into a deterministic classification, using a threshold $c\in[0,1]$.
	So the classification will be 1 if $m(\bX)>c$.
	Then we can define the hit rate at level $c$, $\mathrm{HR}(c)$, and the false alarm rate at level $c$,  $\mathrm{FAR}(c)$, as 
	\[
	\mathrm{HR}(c) = \P(m(\bX)>c|Y=1), \qquad \mathrm{FAR}(c) = \P(m(\bX) >c|Y=0).
	\]
	The ROC curve is then formally defined as the graph
	\[
	\{(\mathrm{FAR}(c),\mathrm{HR}(c))\colon c\in[0,1]\}\subset [0,1]^2.
	\]
	This definition on the population level has a clear counterpart on the sample level.
\end{itemize}

\subsubsection{Imbalanced classes}

There is a large strand of literature about the so-called imbalanced learning problem \cite{He2013ImbalancedLF}.
For binary classification, this means that one class significantly outnumbers the other one.
Typically, main interest is in the rare class $Y=1$, called minority class, with $\P(Y=1) \ll \P(Y=0)$, \eg with a ratio of 1:10, 1:1000 or even more unbalanced.\footnote{%
	An example are Covid-19 cases. Main interest is in the minority group of infected persons, luckily outnumbering the group of not infected people by a large margin.
}

Poor models in this setting are often a result of---any combination of---fitting deterministic classifiers, using re-sampling or re-weighting methods to balance class frequencies in the training data and evaluating the model with a score such as accuracy.
As mentioned above, accuracy is a poor score for model comparison, especially so for imbalanced classes where the cost-ratio $c$, if known, is likely far away from $1/2$.
The frequently encountered practice of over-sampling the minority class or under-sampling the majority class might help to get better (accuracy) scores, but it does not help with the underlying problem:
If the minority class is very rare in absolute terms, no matter the size of the training set, one has effectively a small sample problem.
No re-sampling technique will magically generate more information out of the few cases with the rare class.\footnote{%
	Let us play coin toss with a probability $p$ for $Y=1=\text{\enquote{head}}$.
	Every toss generates a data point adding the Shannon entropy $H = -p\log_2 p - (1-p) \log_2 (1-p)$.
	$H$ is maximised for $p=1/2$ giving $H=1$ bit of information and minimised by $p \in \{0, 1\}$ giving $H=0$ bits.
	Data points for imbalanced classes therefore have little information content.
}
Finally, we generally recommend probabilistic classifiers based on consistent scoring rules over deterministic classifiers.
For instance, a well specified logistic regression model with logit link and intercept fulfils \eqref{eq:canonical score equation} and therefore, at least on the training set, is guaranteed to predict the class probabilities according to their observed frequencies.
We refer to \cite{king_zeng_2001} for more information about logistic regression with rare events, \eg data collection, finite sample corrections and correcting estimates.

\ExampleSec{Binary Classification of Customer Churn} \label{subsec:probabilistic binary classification.Example}

As an example for binary classification, we take the Telco Customer Churn data set\footnote{%
	\url{https://www.openml.org/d/42178}
}.
Each of the $n = 7043$ rows represents a customer of a telecommunication company and the response variable $Y = (\variable{Churn} = \text{\enquote{Yes}})$ indicates if the customer left within the last month.
For the sake of brevity, we refrain from repeating the detailed steps of the regression example and instead focus on the new aspects of this classification task.

Listing~\ref{ls:common data preprocessing classification} shows the short preprocessing and the features used for modelling.
\begin{lstlisting}[caption=Common data preprocessing for classification example, label=ls:common data preprocessing classification]
library(tidyverse)
library(OpenML)


df_origin <- getOMLDataSet(data.id = 42178L)
df <- tibble(df_origin$data)

df[df == "No internet service" | df == "No phone service"] <- "No"

df <- df %>%
  mutate(
    LogTotalCharges = log(as.numeric(TotalCharges)),
    Churn = (Churn == "Yes") + 0,
  ) %>% 
  replace_na(list(LogTotalCharges = median(.$LogTotalCharges, na.rm = TRUE))) %>% 
  mutate_if(is.character, as.factor)

y_var <- "Churn"
x_continuous <- c("tenure", "MonthlyCharges", "LogTotalCharges")
x_discrete <- setdiff(colnames(select_if(df, is.factor)), 
                      c("customerID", y_var, "TotalCharges"))
x_vars <- c(x_continuous, x_discrete)
\end{lstlisting}
For a short exploratory data analysis, we refer to \autoref{subsec:EDA classification}.

We use a stratified train--test split:
\begin{lstlisting}
library(splitTools)

set.seed(34621L)
inds <- partition(df$Churn, p = c(train = 0.75, test = 0.25))
train <- df[inds$train, ]
test <- df[inds$test, ]
y_train <- train[[y_var]]
y_test <- test[[y_var]]
\end{lstlisting}
This guarantees the same average churn frequency on both sets.
We obtain $\bar{y} = \num{0.265}$ on the training and $\bar{y} = \num{0.266}$ on the test set.
This helps to better compare results across both sets.\footnote{%
	If those numbers deviated too much, the \iid assumption would clearly be violated.
}

We train four different models:
\begin{itemize}
	\item Trivial model with $m_{\text{trivial}}(\bx) = 0.2652907$;
	\item Logistic regression (LogReg), \ie a GLM with binomial family and logistic link function;
	\begin{lstlisting}
form <- reformulate(x_vars, y_var)
fit_glm <- glm(form, data = train, family = binomial())
	\end{lstlisting}
	\item Random forest (RF) with \proglang{R} package \href{https://cran.r-project.org/package=ranger}{ranger};
	\begin{lstlisting}
fit_rf <- ranger(
	form, 
	data = train, 
	probability = TRUE, 
	seed = 774, 
	min.node.size = 30, 
	oob.error = TRUE
)
	\end{lstlisting}
	\item XGBoost model with log loss as objective function, logit link function and optimised hyperparameters via cross-validation similar to the regression case.
\end{itemize}

The unconditional calibration with $V(z, y) = z - y$ is summarised in \autoref{tab:unconditional calibration classification}.
\begin{table}
	\centering
	\begin{tabular}{lll}
		\toprule
		& \multicolumn{2}{c}{$\overline{V}(m; D)$} \\
		model & $D=\text{train}$ & $D=\text{test}$ \\
		\midrule
		Trivial  & \num{0}        & \textbf{\num{-0.000317}}\\
		LogReg   & \num{0}        &         \num{-0.00533}\\
		RF       & \num{0.00117}  &         \num{-0.00329}\\
		XGBoost  & \num{0.000483} &         \num{-0.00523}\\
		\bottomrule
	\end{tabular}
	\caption{Assessment of classifiers' unconditional calibration in terms of bias.}
	\label{tab:unconditional calibration classification}
\end{table}
The GLM with canonical link, \ie the logistic regression, once more shows the balance property on the training set.
On the test set, the trivial model has quite a good calibration and, among the feature using models, the random forest is the best unconditionally calibrated one.

\begin{figure}[htbp]
	\centering
	\includegraphics[width=1.00\textwidth]{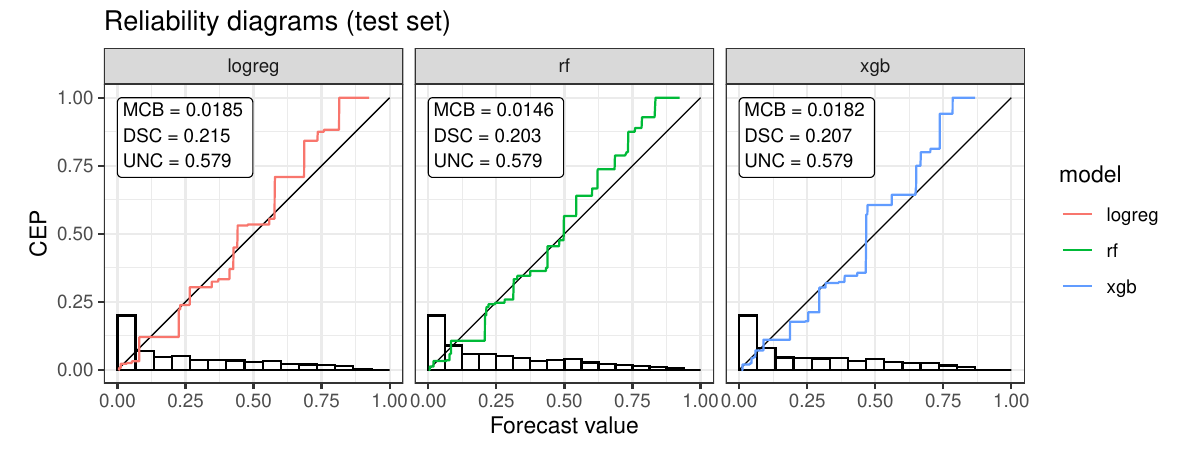}
	\caption{Reliability diagrams. The $x$-axis shows predictions/forecast values, the $y$-axis conditional event probabilities (CEP).
	At the bottom of each plot, the histogram of forecast values is drawn.
	The values in the boxes are the decomposition of the log loss \eqref{eq:log loss} into miscalibration (MCB), discrimination (DSC) and uncertainty (UNC), such that $\overline{S}_{\log} = \mathrm{MCB} - \mathrm{DSC} + \mathrm{UNC}$; see also \eqref{eq:score decomposition}.}
	\label{fig:reliability diagrams}
\end{figure}

As graphical tool to check auto-calibration, we show reliability diagrams on the test set with the CORP approach of \cite{DimitriadisGneitingJordan2021} in \autoref{fig:reliability diagrams}.
The CORP approach essentially performs the binning in an optimal way such that the conditional event probabilities (CEP), which are binned observed frequencies, are isotonic.
The random forest seems to be the best auto-calibrated one.
This is confirmed by the score decomposition of the log loss \eqref{eq:log loss} according to the auto-calibration variant of \eqref{eq:score decomposition} given in \autoref{fig:reliability diagrams}.
The lowest miscalibration (MCB) is obtained by the random forest.
On the other hand, the logistic regression has the highest discriminative power (DSC).

\begin{figure}
	\centering
	\includegraphics[width=.90\textwidth]{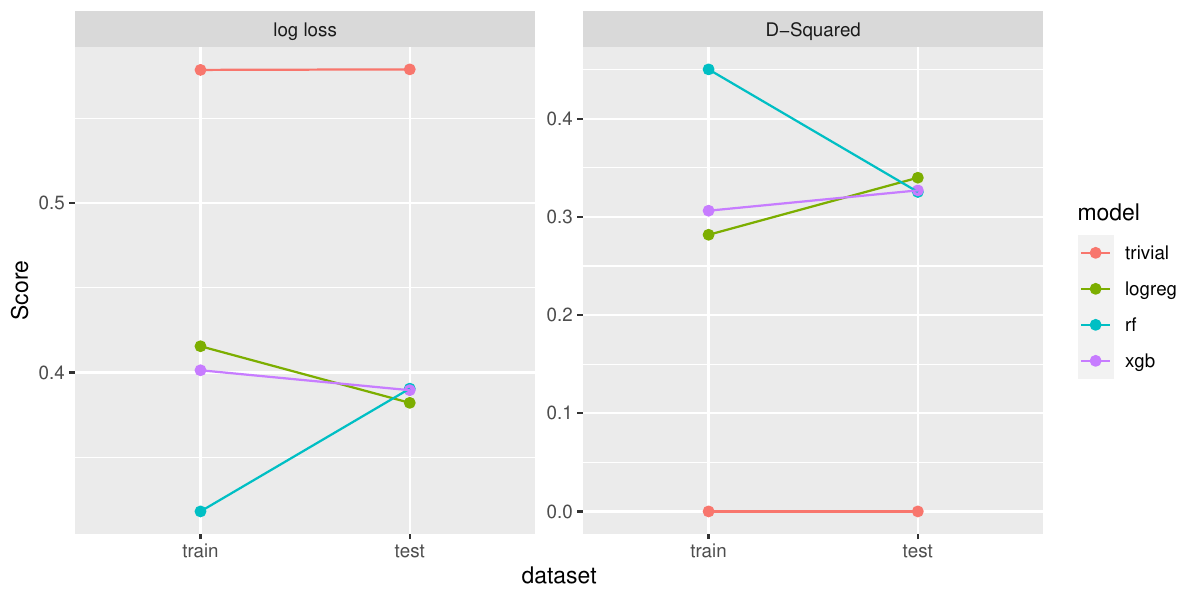}
	\caption{Performance on training and validation data in terms of mean log loss (smaller is better), its relative reduction $\Rstar$ (larger is better).}
	\label{fig:performance classification}
\end{figure}

Finally, we assess the predictive performance with the log loss and the relative reduction in log loss \wrt the trivial model, $\Rstar$, see \autoref{tab:performance classification} and \autoref{fig:performance classification}.
\begin{table}[hbt]
	\centering
	\begin{tabular}{lllllcll}
		\toprule
		        & \multicolumn{2}{c}{mean log loss} & \multicolumn{2}{c}{$\Rstar$} & & \multicolumn{2}{c}{AUC}\\
		Model   &  train & test & train & test & \hspace{0.4cm} & train & test\\
		%\midrule
		\cmidrule(lr){1-5} \cmidrule(lr){7-8}
		Trivial & \num{0.579} &         \num{0.579}  & \num{0    } &         \num{0    }  & & \num{0.500} &         \num{0.500}\\  
		LogReg  & \num{0.415} & \textbf{\num{0.382}} & \num{0.282} & \textbf{\num{0.340}} & & \num{0.845} & \textbf{\num{0.875}}\\
		RF      & \num{0.318} &         \num{0.390}  & \num{0.450} &         \num{0.325}  & & \num{0.935} &         \num{0.868}\\
		XGB     & \num{0.401} &         \num{0.390}  & \num{0.306} &         \num{0.327}  & & \num{0.859} &         \num{0.871}\\
		\bottomrule
	\end{tabular}
	\caption{Performance on training and validation data in terms of mean log loss (smaller is better), its relative reduction $\Rstar$ and AUC (for both larger is better).}
	\label{tab:performance classification}
\end{table}
The random forest clearly shows overfitting as it has the lowest mean log loss on the training set but also the second highest one behind the trivial model on the test set. 
This is not surprising as random forests are typically built of very deep trees.
The XGBoost model performs a touch better than the random forest on the test set.
Note, however, that this time no correction factor was required.
Despite the fact that we did not put much effort in the logistic regression, \eg we did not include any interaction terms nor splines or polynomials of degree two or higher, it ranks on top of all other models regarding predictive performance.
A possible explanation for this ranking is the modest size of the data set.

We also show the AUC.
As the models show different calibration, their ranking in terms of AUC could be very different from the ranking based on log loss.
Coincidentally, AUC shows the same out-of-sample ranking as log loss.
This can be explained by the score decomposition of the log loss in \autoref{fig:reliability diagrams}.
The discrimination components DSC are an order of magnitude larger than the miscalibration components MCB and give the same ranking as according to the log loss.
Despite the appeal of AUC, we generally prefer to evaluate the models with a strictly consistent scoring function, \ie log loss in our case.

\begin{figure}
	\centering
	\includegraphics[width=.90\textwidth]{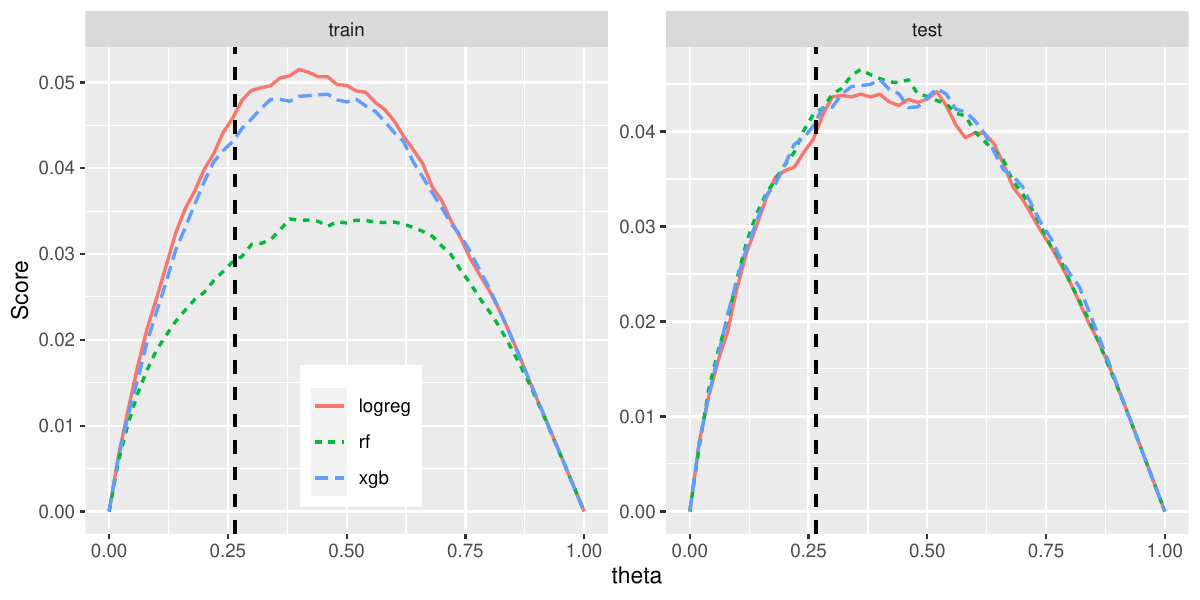}
	\caption{Murphy diagram with elementary scores for $\theta$ from \num{0} to \num{1}.
	The black vertical line indicates the average response on the training data set.}
	\label{fig:murphy diagram classification}
\end{figure}

We conclude this example with the Murphy diagram in \autoref{fig:murphy diagram classification}.
We observe the small values of the random forest on the training set for a large range of scores.
From the right-hand side of the figure, we are visually inclined to award the logistic regression with the best out-of-sample performance.
However, the differences out-of-sample seem to be relatively small such that we refrain from a definite answer as to what model exhibits the best predictive accuracy measured by a wide range of consistent scoring functions.

\section{Conclusion} \label{sec:conclusion}

In this user guide, we have elaborated on best statistical practice in model assessment and model comparison, important tasks in Machine Learning and actuarial science.
We have argued that for both tasks, specifying the modelling target in the form of a statistical functional is crucial. 
The tools for calibration assessment and model comparison---identification functions and scoring functions---need to be chosen in line with the modelling target at hand.
Calibration assessment via strict identification functions checks whether there are any systematic errors in the predictive model.
On the other hand, strictly consistent scoring functions simultaneously encourage good calibration and discrimination ability, honouring the information content of a model.
The methodology has been illustrated with running examples in regression and in binary classification, based on two (almost) real world data sets.
Making a clear distinction between learning (model building) and model evaluation, we have stressed the importance of thorough train--test split of the data.

Clearly, it was not possible to elaborate on all practically relevant aspects in this field. 
For instance, we have entirely omitted the situation of missing or incomplete data. 
Examples from insurance are IBNR (incurred but not yet reported) for frequency models or RBNS (reported but not settled) for claims severity.
Moreover, we have mainly focused on cross-sectional data, only providing some comments on data with serial dependence.

While the majority of the used guide at hand is devoted to a definition of what predictive performance is and how it can be assessed, we would like to close it by remarking that there are sometimes other important (and sometimes conflicting) goals in modelling such as calibration, explainability \cite{LorentzenMayer2020}, impartiality \cite{JohnsonFosterStine2016}, or robustness of predictions \cite{mayer2019}.

\section*{Acknowledgements}
The authors are very grateful to J\"urg Schelldorfer and Mario W\"uthrich as well as to Timo Dimitriadis, Daniel Meier and Marc-Oliver Pohle for their comprehensive reviews and their innumerable inputs which led to substantial improvements of this work.
Christian Lorentzen and Michael Mayer like to thank their company la Mobili\`ere for its support.

\printbibliography

@article{WitkowskiETAL2021,
	archiveprefix = {arXiv},
	author = {J. Witkowski and R. Freeman and J. Wortman Vaughan and D. M. Pennock and A. Krause},
	eprint = {2101.01816},
	journal = {arXiv},
	primaryclass = {cs.GT},
	title = {Incentive-Compatible Forecasting Competitions},
	year = {2021}}

@unpublished{FKP2022,
	author = {T. Fissler and F. Kr\"uger and M.-O. Pohle},
	note = {Mimeo},
	title = {Regression Diagnostics via Generalized Residuals},
	year = {2022}}

@article{LerchETAL2017,
	author = {Lerch, Sebastian and Thorarinsdottir, Thordis L. and Ravazzolo, Francesco and Gneiting, Tilmann},
	da = {2017/02},
	doi = {10.1214/16-STS588},
	isbn = {0883-4237},
	journal = {Statist. Sci.},
	la = {en},
	number = {1},
	pages = {106--127},
	publisher = {The Institute of Mathematical Statistics},
	title = {Forecaster's Dilemma: Extreme Events and Forecast Evaluation},
	ty = {JOUR},
	url = {https://projecteuclid.org:443/euclid.ss/1491465630},
	volume = {32},
	year = {2017}}

@incollection{MurphyDaan1985,
	author = {A. H. Murphy and H. Daan},
	booktitle = {Probability, {S}tatistics and {D}ecision {M}aking in the {A}tmospheric {S}ciences},
	editor = {A. H. Murphy and R. W. Katz},
	pages = {379--437},
	publisher = {Westview Press, Boulder, Colorado},
	title = {{Forecast Evaluation}},
	year = {1985}}

@article{LorentzenMayer2020,
	author = {Lorentzen, C. and Mayer, M.},
	doi = {10.2139/ssrn.3595944},
	journal = {SSRN Manuscript ID 3595944},
	title = {Peeking into the Black Box: An Actuarial Case Study for Interpretable Machine Learning},
	year = {2020}}

@article{JohnsonFosterStine2016,
	archiveprefix = {arXiv},
	author = {Johnson, K. D. and Foster, D. P. and Stine, R. A.},
	eprint = {1608.00528},
	primaryclass = {stat.ME},
	title = {Impartial Predictive Modeling and the Use of Proxy Variables},
	year = {2016}}

@book{RusselNorvig2010,
	author = {S. J. Russell and P. Norvig},
	edition = {Third},
	isbn = {978-0-13-604259-4},
	publisher = {Prentice Hall},
	title = {Artificial Intelligence: A Modern Approach},
	year = {2009}}

@article{FFHR2021,
	author = {T. Fissler and R.Frongillo and J. Hlavinov{\'a} and B. Rudloff},
	booktitle = {Electronic Journal of Statistics},
	doi = {10.1214/21-EJS1808},
	journal = {Electronic Journal of Statistics},
	journal1 = {Electronic Journal of Statistics},
	number = {1},
	pages = {1034--1084},
	title = {Forecast evaluation of quantiles, prediction intervals, and other set-valued functionals},
	ty = {JOUR},
	volume = {15},
	year = {2021}}

@article{BrehmerGneiting2021,
	archiveprefix = {arXiv},
	author = {J. R. Brehmer and T. Gneiting},
	booktitle = {Bernoulli},
	doi = {10.3150/20-BEJ1298},
	eprint = {2007.05709},
	journal = {Bernoulli},
	journal1 = {Bernoulli},
	number = {3},
	pages = {1993--2010},
	primaryclass = {math.ST},
	title = {Scoring interval forecasts: Equal-tailed, shortest, and modal interval},
	ty = {JOUR},
	volume = {27},
	year = {2021}}

@article{DFZ2020,
	archiveprefix = {arXiv},
	author = {T. Dimitriadis and T. Fissler and J. F. Ziegel},
	eprint = {2010.14146},
	journal = {arXiv},
	primaryclass = {math.ST},
	title = {{The Efficiency Gap}},
	year = {2020}}

@article{HeinrichFissler2021,
	archiveprefix = {arXiv},
	author = {Heinrich-Mertsching, C. and Fissler, T.},
	doi = {10.1093/biomet/asab065},
	eprint = {2109.00464},
	journal = {Biometrika (forthcoming)},
	title = {Is the mode elicitable relative to unimodal distributions?},
	year = {2021}}

@article{SteinwartPasinETAL2014,
	author = {I. Steinwart and C. Pasin and R. Williamson and S. Zhang},
	journal = {JMLR Workshop Conf. Proc.},
	pages = {1--45},
	title = {{Elicitation and Identification of Properties}},
	url = {http://proceedings.mlr.press/v35/steinwart14.html},
	volume = {35},
	year = {2014}}

@article{DawidVovk1999,
	author = {A. P. Dawid and V. G. Vovk},
	booktitle = {Bernoulli},
	doi = {bj/1173707098},
	journal = {Bernoulli},
	journal1 = {Bernoulli},
	number = {1},
	pages = {125--162},
	title = {Prequential probability: principles and properties},
	ty = {JOUR},
	volume = {5},
	year = {1999}}

@article{Davis2016,
	author = {Davis, M. H. A.},
	doi = {10.1515/strm-2015-0007},
	journal = {Statistics \& Risk Modeling},
	number = {3--4},
	pages = {67--93},
	title = {Verification of internal risk measure estimates},
	volume = {33},
	year = {2016}}

@article{NoldeZiegel2017,
	author = {Nolde, N. and Ziegel, J. F.},
	doi = {10.1214/17-AOAS1041},
	journal = {The Annals of Applied Statistics},
	number = {4},
	pages = {1833--1874},
	title = {{Elicitability and backtesting: Perspectives for banking regulation}},
	volume = {11},
	year = {2017}}

@article{HenziZiegel2021,
	archiveprefix = {arXiv},
	author = {Henzi, A. and {Ziegel}, J.~F.},
	eprint = {2103.08402},
	journal = {arXiv},
	primaryclass = {stat.ME},
	title = {Valid sequential inference on probability forecast performance},
	year = {2021}}

@article{FisslerZiegel2016,
	author = {T. Fissler and J. F. Ziegel},
	doi = {10.1214/16-AOS1439},
	fjournal = {Annals of Statistics},
	journal = {The Annals of Statistics},
	number = {4},
	pages = {1680--1707},
	sici = {0090-5364(2016)44:4<1680:HOEAOP>2.0.CO;2-U},
	title = {{Higher order elicitability and Osband's principle}},
	volume = {44},
	year = {2016}}

@phdthesis{Osband1985,
	author = {K. H. Osband},
	doi = {10.5281/zenodo.4355667},
	school = {University of California, Berkeley},
	title = {{Providing Incentives for Better Cost Forecasting}},
	year = {1985}}

@article{Gneiting2017,
	archiveprefix = {arXiv},
	author = {T. Gneiting},
	doi = {10.1002/sta4.148},
	eprint = {1704.08979},
	journal = {Stat},
	number = {1},
	pages = {204--206},
	primaryclass = {math.ST},
	title = {When is the mode functional the Bayes classifier?},
	volume = {6},
	year = {2017}}

@article{GR07,
	author = {Gneiting, T. and Raftery, A.~E.},
	doi = {10.1198/016214506000001437},
	journal = {Journal of the American Statistical Association},
	pages = {359--378},
	title = {Strictly Proper Scoring Rules, Prediction, and Estimation},
	url = {http://www.stat.washington.edu/people/raftery/Research/PDF/Gneiting2007jasa.pdf},
	volume = {102},
	year = {2007}}

@article{HolzmannEulert2014,
	author = {H. Holzmann and M. Eulert},
	doi = {10.1214/13-AOAS709},
	journal = {The Annals of Applied Statistics},
	pages = {79--83},
	title = {The role of the information set for forecasting -- with applications to risk management},
	volume = {8},
	year = {2014}}

@article{Pohle2020,
	archiveprefix = {arXiv},
	author = {Pohle, M.-O.},
	eprint = {2005.01835},
	journal = {arXiv},
	primaryclass = {stat.ME},
	title = {The Murphy Decomposition and the Calibration-Resolution Principle: A New Perspective on Forecast Evaluation},
	year = {2020}}

@article{DimitriadisGneitingJordan2021,
	author = {Dimitriadis, T. and Gneiting, T. and Jordan, A. I.},
	doi = {10.1073/pnas.2016191118},
	journal = {Proceedings of the National Academy of Sciences},
	journal1 = {Proc Natl Acad Sci USA},
	n2 = {Probabilistic classifiers assign predictive probabilities to binary events, such as rainfall tomorrow, a recession, or a personal health outcome. Such a system is reliable or calibrated if the predictive probabilities are matched by the observed frequencies. In practice, calibration is assessed graphically in reliability diagrams and quantified via the reliability component of mean scores. Extant approaches rely on binning and counting and have been hampered by ad hoc implementation decisions, a lack of reproducibility, and inefficiency. Here, we introduce the CORP approach, which uses the pool-adjacent-violators algorithm to generate optimally binned, reproducible, and provably statistically consistent reliability diagrams, along with a numerical measure of miscalibration based on a revisited score decomposition.A probability forecast or probabilistic classifier is reliable or calibrated if the predicted probabilities are matched by ex post observed frequencies, as examined visually in reliability diagrams. The classical binning and counting approach to plotting reliability diagrams has been hampered by a lack of stability under unavoidable, ad hoc implementation decisions. Here, we introduce the CORP approach, which generates provably statistically consistent, optimally binned, and reproducible reliability diagrams in an automated way. CORP is based on nonparametric isotonic regression and implemented via the pool-adjacent-violators (PAV) algorithm---essentially, the CORP reliability diagram shows the graph of the PAV-(re)calibrated forecast probabilities. The CORP approach allows for uncertainty quantification via either resampling techniques or asymptotic theory, furnishes a numerical measure of miscalibration, and provides a CORP-based Brier-score decomposition that generalizes to any proper scoring rule. We anticipate that judicious uses of the PAV algorithm yield improved tools for diagnostics and inference for a very wide range of statistical and machine learning methods.The probability of precipitation forecast data at Niamey, Niger, are from the paper by Vogel et al. (ref. 7, figure 2), where the original data sources are acknowledged. Precipitation forecasts and realizations data have been deposited at GitHub (https://github.com/TimoDimi/replication DGJ20). Additional data analyses, simulation studies, technical discussion, and the proofs of Theorems 1 and 2 have been relegated to SI Appendix. Reproduction material for both the main article and SI Appendix, including data and code in the R software environment (51), are available online (38, 53). Open-source code for the implementation of the CORP approach in the R language and environment for statistical computing (51) is available on CRAN (38).},
	number = {8},
	pages = {e2016191118},
	title = {Stable reliability diagrams for probabilistic classifiers},
	ty = {JOUR},
	volume = {118},
	year = {2021}}

@article{Fawcett2006,
	author = {Fawcett, T.},
	booktitle = {ROC Analysis in Pattern Recognition},
	da = {2006/06/01/},
	doi = {10.1016/j.patrec.2005.10.010},
	journal = {Pattern Recognition Letters},
	number = {8},
	pages = {861--874},
	title = {An introduction to ROC analysis},
	ty = {JOUR},
	volume = {27},
	year = {2006}}

@article{Byrne2016,
	author = {Byrne, S.},
	da = {2016},
	doi = {10.1214/16-EJS1109},
	journal = {Electronic Journal of Statistics},
	la = {en},
	number = {1},
	pages = {380--393},
	publisher = {The Institute of Mathematical Statistics and the Bernoulli Society},
	title = {A note on the use of empirical AUC for evaluating probabilistic forecasts},
	ty = {JOUR},
	volume = {10},
	year = {2016}}

@article{Diebold1995,
	author = {F. X. Diebold and R. S. Mariano},
	doi = {10.2307/1392185},
	journal = {Journal of Business \& Economic Statistics},
	number = {1},
	pages = {253-265},
	publisher = {American Statistical Association},
	title = {Comparing Predictive Accuracy},
	url = {https://citeseerx.ist.psu.edu/viewdoc/download?doi=10.1.1.454.4490&rep=rep1&type=pdf},
	volume = {13},
	year = {1995}}

@inbook{Wilks2011,
	author = {Wilks, D. S.},
	booktitle = {Statistical Methods in the Atmospheric Sciences},
	chapter = {8},
	doi = {10.1016/B978-0-12-385022-5.00008-7},
	edition = {3},
	isbn = {978-0-12-385022-5},
	pages = {301-394},
	publisher = {Elsevier},
	series = {International Geophysics Series},
	title = {Forecast Verification},
	volumne = {100},
	year = {2011}}

@article{ehm2015quantiles,
	archiveprefix = {arXiv},
	author = {W. Ehm and T. Gneiting and A. I. Jordan and F. Kr\"uger},
	doi = {10.1111/rssb.12154},
	eprint = {1503.08195},
	journal = {Journal of the Royal Statistical Society: Series B (Statistical Methodology)},
	number = {3},
	pages = {505--562},
	primaryclass = {math.ST},
	title = {Of quantiles and expectiles: consistent scoring functions, Choquet representations and forecast rankings},
	volume = {78},
	year = {2016}}

@book{HuberRonchetti2009,
	address = {Hoboken, New Jersey},
	author = {P. J. Huber and E. M. Ronchetti},
	edition = {Second},
	publisher = {John Wiley \& Sons, Inc.},
	title = {Robust Statistics},
	year = {2009}}

@article{Huber1964,
	author = {Huber, P. J.},
	doi = {10.1214/aoms/1177703732},
	journal = {The Annals of Mathematical Statistics},
	month = {03},
	number = {1},
	pages = {73--101},
	title = {{Robust Estimation of a Location Parameter}},
	volume = {35},
	year = {1964}}

@book{hastie01statisticallearning,
	address = {New York, NY, USA},
	author = {Hastie, T. and Tibshirani, R. and Friedman, J.},
	publisher = {Springer New York Inc.},
	series = {Springer Series in Statistics},
	title = {The Elements of Statistical Learning},
	url = {https://web.stanford.edu/~hastie/ElemStatLearn/},
	year = 2001}

@book{harrell2015,
	author = {F. E. Harrell},
	doi = {10.1007/978-3-319-19425-7},
	edition = {2},
	isbn = {978-3-319-19424-0},
	publisher = {Springer International Publishing Switzerland 2015},
	series = {Springer Series in Statistics},
	title = {Regression Modeling Strategies},
	year = {2015}}

@article{gneiting_2011,
	archiveprefix = {arXiv},
	author = {Gneiting, T.},
	doi = {10.1198/jasa.2011.r10138},
	eprint = {0912.0902},
	journal = {Journal of the American Statistical Association},
	number = {494},
	pages = {746--762},
	primaryclass = {math},
	title = {Making and Evaluating Point Forecasts},
	volume = {106},
	year = {2011}}

@article{gneiting2014,
	author = {T. Gneiting and M. Katzfuss},
	doi = {10.1146/annurev-statistics-062713-085831},
	journal = {Annual Review of Statistics and Its Application},
	number = {1},
	pages = {125--151},
	publisher = {Annual Reviews},
	title = {Probabilistic Forecasting},
	volume = {1},
	year = {2014}}

@techreport{Buja2005LossFF,
	author = {A. Buja and W. Stuetzle and Y. Shen},
	institution = {University of Pennsylvania},
	title = {Loss Functions for Binary Class Probability Estimation and Classification: Structure and Applications},
	url = {http://www-stat.wharton.upenn.edu/~buja/PAPERS/paper-proper-scoring.pdf},
	year = {2005}}

@article{Patton2011,
	author = {A. J. Patton},
	doi = {10.1016/j.jeconom.2010.03.034},
	journal = {Journal of Econometrics},
	number = {1},
	pages = {246--256},
	publisher = {Elsevier {BV}},
	title = {Volatility forecast comparison using imperfect volatility proxies},
	url = {http://public.econ.duke.edu/~ap172/Patton_vol_proxies_JoE_2011.pdf},
	volume = {160},
	year = {2011}}

@article{Raschka2018,
	archiveprefix = {arXiv},
	author = {S. Raschka},
	bibsource = {dblp computer science bibliography, https://dblp.org},
	eprint = {1811.12808},
	journal = {arXiv},
	primaryclass = {cs.LG},
	title = {Model Evaluation, Model Selection, and Algorithm Selection in Machine Learning},
	year = {2018}}

@book{ShalevShwartz2014,
	author = {S. Shalev-Shwartz and S. Ben-David},
	doi = {10.1017/cbo9781107298019},
	publisher = {Cambridge University Press},
	title = {Understanding Machine Learning},
	url = {https://www.cse.huji.ac.il/~shais/UnderstandingMachineLearning},
	year = {2014}}

@inproceedings{Tweedie1984AnIW,
	author = {M. C. K. Tweedie},
	pages = {579--604},
	series = {Statistics: Applications and New Directions.},
	title = {An index which distinguishes between some important exponential families},
	volume = {Proceedings of the Indian Statistical InstituteGolden Jubilee International Conference},
	year = {1984}}

@book{Jorgensen1997,
	address = {London},
	author = {B. J{\o}rgensen},
	isbn = {978-0-412-99711-2},
	publisher = {Chapman and Hall/CRC},
	title = {The Theory of Dispersion Models},
	year = {1997}}

@book{kroese2019DSML,
	address = {Boca Raton},
	author = {Kroese, D. P. and Botev, Z. I. and Taimre, T. and Vaisman, R.},
	isbn = {978-1-138-49253-0},
	lccn = {2019030250},
	publisher = {CRC Press},
	series = {Chapman \& Hall/CRC machine learning \& pattern recognition},
	title = {Data Science and Machine Learning: Mathematical and Statistical Methods},
	url = {https://people.smp.uq.edu.au/DirkKroese/DSML/},
	year = {2019}}

@article{vonluxburg2008statistical,
	archiveprefix = {arXiv},
	author = {U. von Luxburg and B. Schoelkopf},
	eprint = {0810.4752},
	primaryclass = {stat.ML},
	title = {Statistical Learning Theory: Models, Concepts, and Results},
	year = {2008}}

@book{mitchell1997ML,
	address = {USA},
	author = {Mitchell, T. M.},
	edition = {1},
	isbn = {978-0-07-042807-2},
	publisher = {McGraw-Hill, Inc.},
	title = {Machine Learning},
	year = {1997}}

@article{bates2021crossvalidation,
	archiveprefix = {arXiv},
	author = {S. Bates and T. Hastie and R. Tibshirani},
	eprint = {2104.00673},
	primaryclass = {stat.ME},
	title = {Cross-validation: what does it estimate and how well does it do it?},
	year = {2021}}

@article{Nadeau_2003,
	author = {C. Nadeau and Y. Bengio},
	doi = {10.1023/a:1024068626366},
	journal = {Machine Learning},
	number = {3},
	pages = {239--281},
	publisher = {Springer Science and Business Media LLC},
	title = {Inference for the Generalization Error},
	volume = {52},
	year = {2003}}

@inbook{Bouckaert_Frank_2004,
	author = {Bouckaert, R. R. and Frank, E.},
	booktitle = {Advances in Knowledge Discovery and Data Mining},
	doi = {10.1007/978-3-540-24775-3_3},
	pages = {3--12},
	publisher = {Springer Berlin Heidelberg},
	title = {Evaluating the Replicability of Significance Tests for Comparing Learning Algorithms},
	url = {http://www.cs.waikato.ac.nz/~ml/publications/2004/bouckaert-frank.pdf},
	year = {2004}}

@article{gneiting2021receiver,
	archiveprefix = {arXiv},
	author = {T. Gneiting and P. Vogel},
	doi = {10.1007/s10994-021-06115-2},
	eprint = {1809.04808},
	journal = {arXiv},
	primaryclass = {stat.ME},
	title = {Receiver operating characteristic (ROC) curves: equivalences, beta model, and minimum distance estimation},
	year = {2021}}

@article{gneiting2021regression,
	archiveprefix = {arXiv},
	author = {T. Gneiting and J. Resin},
	eprint = {2108.03210},
	journal = {arXiv},
	primaryclass = {stat.ME},
	title = {Regression Diagnostics meets Forecast Evaluation: Conditional Calibration, Reliability Diagrams, and Coefficient of Determination},
	year = {2021}}

@article{10.1214/19-EJS1552,
	author = {T. Fissler and J. F. Ziegel},
	doi = {10.1214/19-EJS1552},
	journal = {Electronic Journal of Statistics},
	number = {1},
	pages = {1166--1211},
	publisher = {Institute of Mathematical Statistics and Bernoulli Society},
	title = {{Order-sensitivity and equivariance of scoring functions}},
	volume = {13},
	year = {2019}}

@article{fissler2021deep,
	archiveprefix = {arXiv},
	author = {T. Fissler and M. Merz and M. V. W{\"u}thrich},
	eprint = {2112.03075},
	journal = {arXiv},
	primaryclass = {stat.ME},
	title = {Deep Quantile and Deep Composite Model Regression},
	year = {2021}}

@article{Wuthrich2021StatisticalFoundations,
	author = {M. V. W\"uthrich and M. Merz},
	day = {3},
	doi = {10.2139/ssrn.3822407},
	journal = {SSRN Manuscript ID 3822407},
	month = {11},
	title = {Statistical Foundations of Actuarial Learning and its Applications},
	year = {2021}}

@article{MurphyWinkler1984,
	author = {A. H. Murphy and R. L. Winkler},
	doi = {10.2307/2288395},
	issn = {01621459},
	journal = {Journal of the American Statistical Association},
	number = {387},
	pages = {489--500},
	publisher = {[American Statistical Association, Taylor & Francis, Ltd.]},
	title = {Probability Forecasting in Meterology},
	volume = {79},
	year = {1984}}

@article{ThompsonBrier1955,
      address = {Boston MA, USA},
      author = {J. C. Thompson and G. W. Brier},
      doi = {10.1175/1520-0493(1955)083<0249:TEUOWF>2.0.CO;2},
      journal = {Monthly Weather Review},
      number = {11},
      pages = {249 - 253},
      publisher = {American Meteorological Society},
      title = {The Economic Utility of Weather Forecasts},
      volume = {83},
      year = {1955}
}

@inproceedings{Elkan2001Foundations,
	address = {San Francisco, CA, USA},
	author = {Elkan, C.},
	booktitle = {Proceedings of the 17th International Joint Conference on Artificial Intelligence},
	isbn = {978-1-55860-812-2},
	numpages = {6},
	pages = {973--978},
	publisher = {Morgan Kaufmann Publishers Inc.},
	series = {IJCAI'01},
	title = {The Foundations of Cost-Sensitive Learning},
	url = {https://citeseerx.ist.psu.edu/viewdoc/download?doi=10.1.1.29.514&rep=rep1&type=pdf},
	year = {2001}}

@techreport{Powers2007evaluation,
	archiveprefix = {arXiv},
	author = {D. M. W. Powers},
	eprint = {2010.16061},
	institution = {School of Informatics and Engineering Flinders University, Adelaide},
	number = {SIE-07-001},
	primaryclass = {cs.LG},
	title = {Evaluation: From Precision, Recall and F-Factor to ROC, Informedness, Markedness \& Correlation},
	year = {2007}}

@article{Tharwat2020ClassificationAM,
	author = {A. Tharwat},
	doi = {10.1016/j.aci.2018.08.003},
	journal = {New England Journal of Entrepreneurship},
	number = {1},
	pages = {168--192},
	publisher = {Emerald Publishing Limited},
	title = {Classification assessment methods},
	volume = {17},
	year = {2020}}

@article{ArlotCelisse2010,
	archiveprefix = {arXiv},
	author = {S. Arlot and A. Celisse},
	doi = {10.1214/09-ss054},
	eprint = {0907.4728},
	journal = {Statistics Surveys},
	pages = {40 -- 79},
	primaryclass = {math.ST},
	publisher = {Amer. Statist. Assoc., the Bernoulli Soc., the Inst. Math. Statist., and the Statist. Soc. Canada},
	title = {{A survey of cross-validation procedures for model selection}},
	volume = {4},
	year = {2010}}

@article{Roberts2017CrossvalidationSF,
	author = {D. R. Roberts and V. Bahn and S. Ciuti and M. S. Boyce and J. Elith and G. Guillera‐Arroita and S. Hauenstein and J. J. Lahoz‐Monfort and B. Schr\"oder and W. Thuiller and D. I. Warton and B. A. Wintle and F. Hartig and C. F. Dormann},
	doi = {10.1111/ecog.02881},
	journal = {Ecography},
	number = {8},
	pages = {913-929},
	title = {Cross-validation strategies for data with temporal, spatial, hierarchical, or phylogenetic structure},
	volume = {40},
	year = {2017}}

@article{Stein2007BenchmarkingDP,
	author = {R. M. Stein},
	doi = {10.21314/JRMV.2007.002},
	journal = {The Journal of Risk Model Validation},
	number = {1},
	pages = {77-113},
	title = {Benchmarking default prediction models: pitfalls and remedies in model validation},
	url = {http://www.rogermstein.com/wp-content/uploads/BenchmarkingDefaultPredictionModels_TR030124.pdf},
	volume = {1},
	year = {2007}}

@inproceedings{Hyndman2021ForecastingPA,
	author = {R. J. Hyndman and G. Athanasopoulos},
	edition = {3},
	publisher = {OTexts: Melbourne, Australia},
	title = {Forecasting: principles and practice},
	url = {https://otexts.com/fpp3},
	year = {2021}}

@techreport{Schnaubelt2019comparison,
	address = {N\"{u}rnberg},
	author = {M. Schnaubelt},
	copyright = {http://www.econstor.eu/dspace/Nutzungsbedingungen},
	language = {eng},
	number = {11/2019},
	publisher = {Friedrich-Alexander-Universit\"{a}t Erlangen-N\"{u}rnberg, Institute for Economics},
	title = {A comparison of machine learning model validation schemes for non-stationary time series data},
	type = {FAU Discussion Papers in Economics},
	url = {http://hdl.handle.net/10419/209136},
	year = {2019}}

@article{Granger1999EconomicAS,
	author = {C. W. J. Granger and M. Hashem Pesaran},
	doi = {10.1002/1099-131X(200012)19:7<537::AID-FOR769>3.0.CO;2-G},
	journal = {Journal of Forecasting},
	pages = {537-560},
	title = {Economic and Statistical Measures of Forecast Accuracy},
	url = {https://www.repository.cam.ac.uk/bitstream/1810/421/1/gp1999.pdf},
	volume = {19},
	year = {1999}}

@book{He2013ImbalancedLF,
	doi = {10.1002/9781118646106},
	editor = {H. He and Y. Ma},
	isbn = {978-1-118-07462-6},
	publisher = {John Wiley \& Sons, Inc.},
	title = {Imbalanced Learning: Foundations, Algorithms, and Applications},
	year = {2013}}

@article{king_zeng_2001,
	author = {King, G. and Zeng, L.},
	doi = {10.1093/oxfordjournals.pan.a004868},
	journal = {Political Analysis},
	number = {2},
	pages = {137-163},
	publisher = {Cambridge University Press},
	title = {Logistic Regression in Rare Events Data},
	volume = {9},
	year = {2001}}

@inproceedings{chen2016,
	author = {T. Chen and C. Guestrin},
	booktitle = {Proceedings of the 22nd ACM SIGKDD International Conference on Knowledge Discovery and Data Mining},
	doi = {10.1145/2939672.2939785},
	location = {San Francisco, California, USA},
	numpages = {10},
	pages = {785--794},
	publisher = {Association for Computing Machinery},
	series = {KDD '16},
	title = {XGBoost: A Scalable Tree Boosting System},
	url = {http://doi.acm.org/10.1145/2939672.2939785},
	year = {2016}}

@inproceedings{10.1007/3-540-44581-1_15,
	address = {Berlin, Heidelberg},
	author = {Bartlett, P. L. and Mendelson, S.},
	booktitle = {Computational Learning Theory},
	doi = {10.1007/3-540-44581-1_15},
	editor = {Helmbold, D. and Williamson, B.},
	pages = {224--240},
	publisher = {Springer Berlin Heidelberg},
	title = {Rademacher and Gaussian Complexities: Risk Bounds and Structural Results},
	url = {http://www.jmlr.org/papers/volume3/bartlett02a/bartlett02a.pdf},
	year = {2001}}

@article{neyshabur2015search,
	archivePrefix = {arXiv},
	author = {B. Neyshabur and R. Tomioka and N. Srebro},
	eprint = {1412.6614},
	primaryClass = {cs.LG},
	title = {In Search of the Real Inductive Bias: On the Role of Implicit Regularization in Deep Learning}, 
	year={2015}}

@article{zhang2017understanding,
	archivePrefix = {arXiv},
	author = {C. Zhang and S. Bengio and M. Hardt and B. Recht and O. Vinyals},
	eprint = {1611.03530},
	primaryClass = {cs.LG},
	title = {Understanding deep learning requires rethinking generalization},
	year = {2017}}

@article {Bartlett30063,
	author = {Bartlett, P. L. and Long, P. M. and Lugosi, G. and Tsigler, A.},
	doi = {10.1073/pnas.1907378117},
	journal = {Proceedings of the National Academy of Sciences},
	number = {48},
	pages = {30063--30070},
	publisher = {National Academy of Sciences},
	title = {Benign overfitting in linear regression},
	volume = {117},
	year = {2020}}

@article {Belkin15849,
	author = {Belkin, M. and Hsu, D. and Ma, S. and Mandal, S.},
	doi = {10.1073/pnas.1903070116},
	journal = {Proceedings of the National Academy of Sciences},
	number = {32},
	pages = {15849--15854},
	publisher = {National Academy of Sciences},
	title = {Reconciling modern machine-learning practice and the classical bias{\textendash}variance trade-off},
	volume = {116},
	year = {2019}}

@article{Belkin2020,
	archivePrefix = {arXiv},
	author = {Belkin, M. and Hsu, D. and Xu, J.},
	doi = {10.1137/20m1336072},
	eprint = {1903.07571},
	issn = {2577-0187},
	journal = {SIAM Journal on Mathematics of Data Science},
	number = {4},
	pages = {1167--1180},
	primaryClass = {math.ST},
	publisher = {Society for Industrial & Applied Mathematics (SIAM)},
	volume = {2},
	title = {Two Models of Double Descent for Weak Features},
	year = {2020}}

@article{hastie2020surprises,
	archivePrefix = {arXiv},
	author = {Trevor Hastie and Andrea Montanari and Saharon Rosset and Ryan J. Tibshirani},
	eprint = {1903.08560},
	primaryClass = {math.ST},
	title = {Surprises in High-Dimensional Ridgeless Least Squares Interpolation}, 
	year = {2020}}

@article{dar2021farewell,
    archivePrefix = {arXiv},
    author = {Y. Dar and V. Muthukumar and R. G. Baraniuk},
    eprint = {2109.02355},
    primaryClass = {stat.ML},
    title = {A Farewell to the Bias-Variance Tradeoff? An Overview of the Theory of Overparameterized Machine Learning},
    year = {2021}}

@article{10.1162/neco.1992.4.1.1,
    author = {Geman, S. and Bienenstock, E. and Doursat, R.},
    doi = {10.1162/neco.1992.4.1.1},
    journal = {Neural Computation},
    month = {01},
    number = {1},
    pages = {1-58},
    title = {Neural Networks and the Bias/Variance Dilemma},
    volume = {4},
    year = {1992},
}

@article{mayer2019,
	author = {M. Mayer and S. C. Bourassa and M. Hoesli and D. Scognamiglio},
	doi = {10.1108/JERER-08-2018-0035},
	journal = {Journal of European Real Estate Research},
	language = {English},
	number = {1},
	pages = {134--150},
	publisher = {Emerald Group Publishing Ltd.},
	title = {Estimation and Updating Methods for Hedonic Valuation},
	url = {https://ssrn.com/abstract=3300193},
	volume = {12},
	year = {2019}
}

@article{ANewVectorPartitionoftheProbabilityScore,
	address = {Boston MA, USA},
    author = {Allan H.  Murphy},
	doi = {10.1175/1520-0450(1973)012<0595:ANVPOT>2.0.CO;2},
	journal = {Journal of Applied Meteorology and Climatology},
	number = {4},
	pages = {595--600},
	publisher = {American Meteorological Society},
	title = {A New Vector Partition of the Probability Score},
	volume = {12},
	year = {1973}
}

@article{KrugerZiegel2021,
	author = {Fabian Krüger and Johanna F. Ziegel},	
	doi = {10.1080/07350015.2020.1741376},
	journal = {Journal of Business \& Economic Statistics},
	number = {4},
	pages = {972-983},
	publisher = {Taylor & Francis},
	title = {Generic Conditions for Forecast Dominance},
	url = {https://doi.org/10.7892/boris.141834},
	volume = {39},
	year  = {2021}
}

@article{Saerens2000,
	author = {M. Saerens},
	doi = {10.1109/72.883416},
	journal = {IEEE Transactions on Neural Networks},
	number = {6},
	pages = {1263-1271},
	title = {Building cost functions minimizing to some summary statistics},
	volume = {11},
	year = {2000}
}

@article{Jordan2022,
  	author = {Jordan, Alexander I. and M\"uhlemann, Anja and Ziegel, Johanna F.},
	doi = {10.1007/s10463-021-00808-0},
	journal = {Annals of the Institute of Statistical Mathematics},
	number = {3},
	pages = {489--514},
	title = {Characterizing the optimal solutions to the isotonic regression problem for identifiable functionals},
	volume = {74},
	year = {2022}
}

\appendix

\section{Exploratory Data Analysis}

\subsection{Regression: Workers' Compensation Data Set} \label{subsec:EDA regression}

\begin{figure}[htbp]
	\centering
	\includegraphics[width=.90\textwidth]{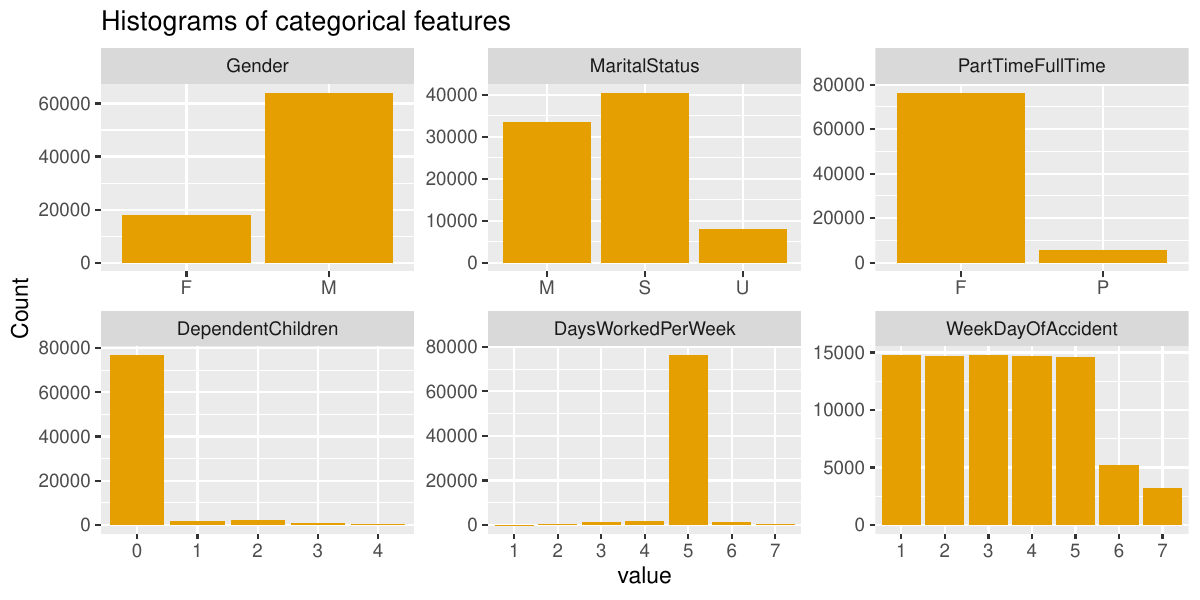}
	\caption{Histograms of categorical features of Workers' Compensation data set.}
	\label{fig:histogram categorical features}
\end{figure}

\begin{figure}[htbp]
	\centering
	\includegraphics[width=.90\textwidth]{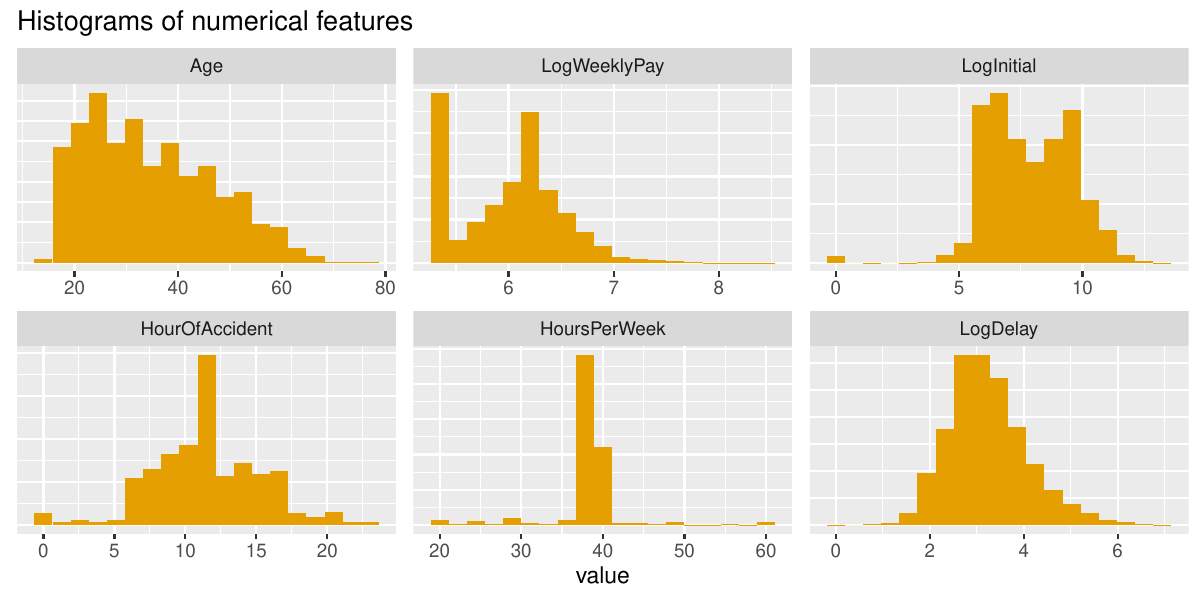}
	\caption{Histograms of numerical features of Workers' Compensation data set.}
	\label{fig:histogram numerical features}
\end{figure}

\begin{figure}[htbp]
	\centering
	\includegraphics[width=.90\textwidth]{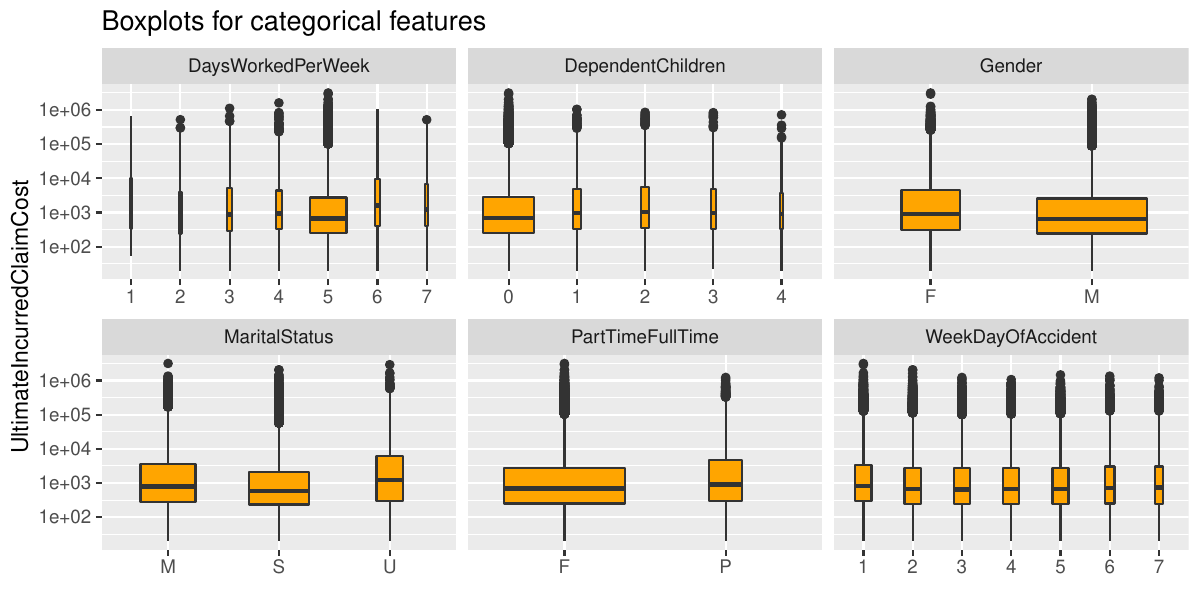}
	\caption{Boxplots of response \variable{UltimateIncurredClaimCost} conditional on categorical features of Workers' Compensation data set.}
	\label{fig:rel plot categorical features}
\end{figure}

\begin{figure}[htbp]
	\centering
	\includegraphics[width=.90\textwidth]{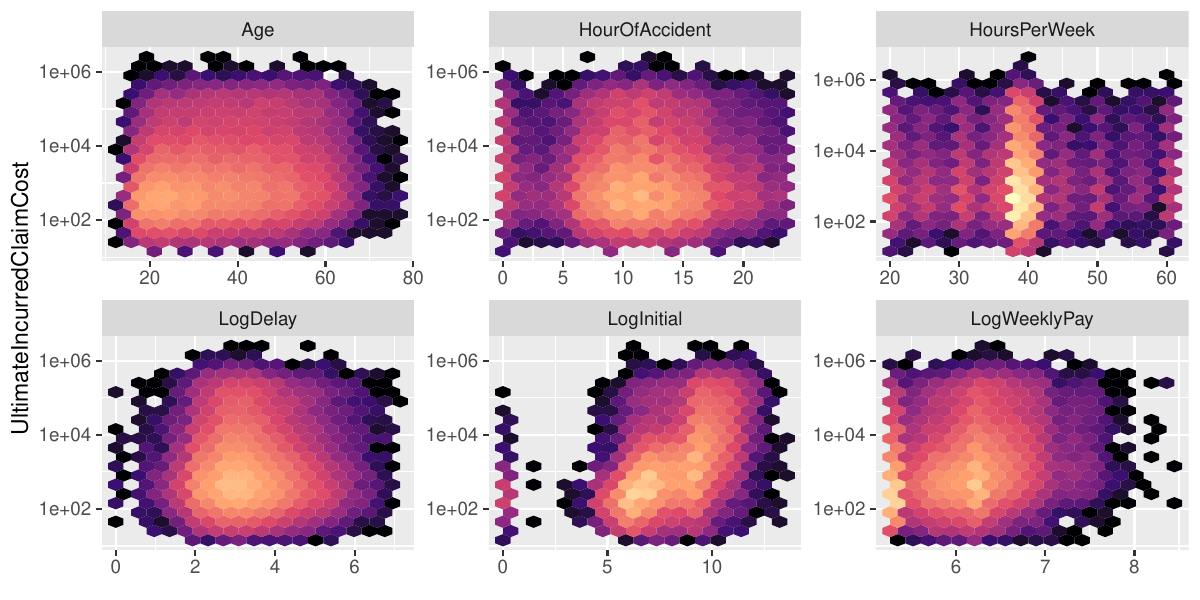}
	\caption{Density plots of response \variable{UltimateIncurredClaimCost} conditional on numerical features Workers' Compensation data set.}
	\label{fig:rel plot numerical features}
\end{figure}

See Figures~\ref{fig:histogram categorical features}, \ref{fig:histogram numerical features}, \ref{fig:rel plot categorical features} and \ref{fig:rel plot numerical features}.

\FloatBarrier

\subsection{Classification: Telco Customer Churn Data Set} \label{subsec:EDA classification}

\begin{figure}[htbp]
	\centering
	\includegraphics[width=.90\textwidth]{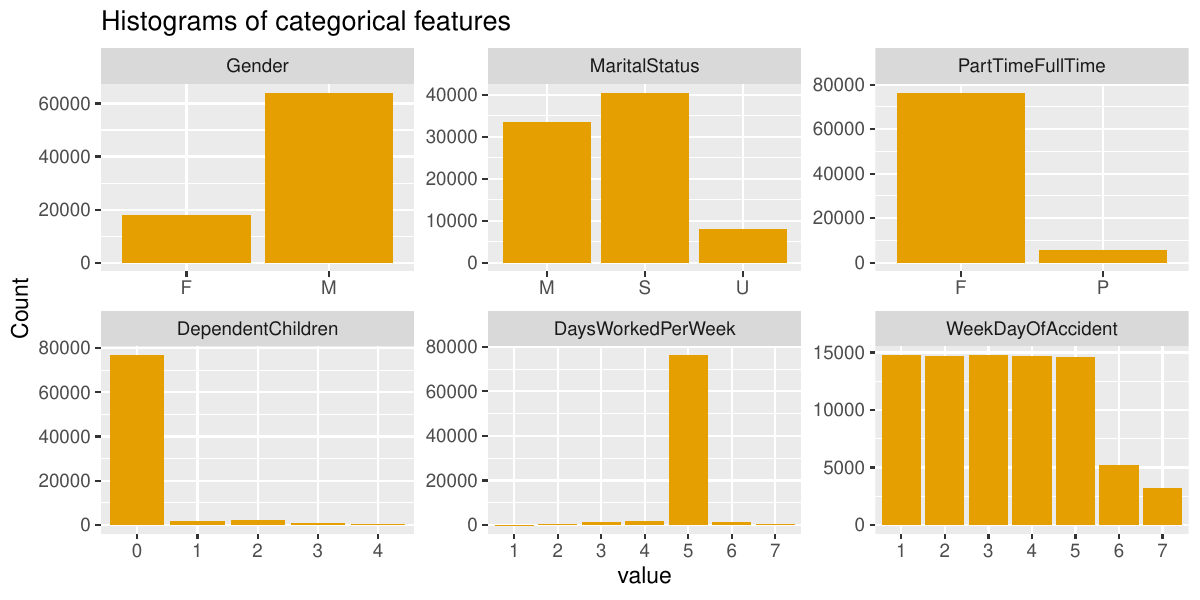}
	\caption{Histograms of categorical features of Telco Customer Churn data set.}
	\label{fig:histogram categorical features classification}
\end{figure}

\begin{figure}
	\centering
	\includegraphics[width=.90\textwidth]{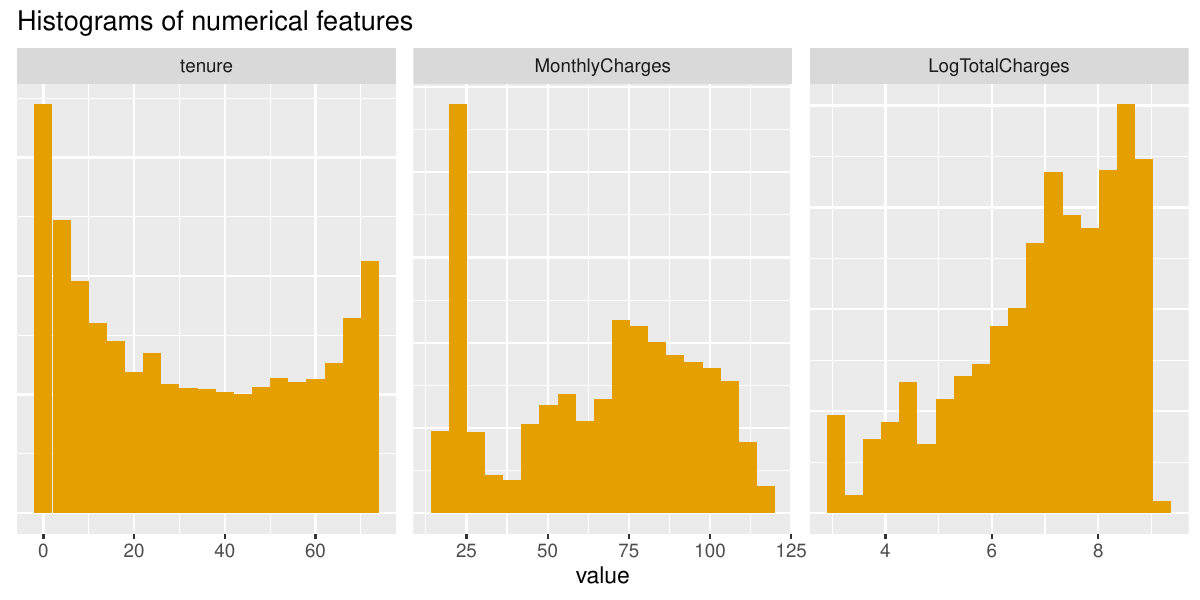}
	\caption{Histograms of numerical features of Telco Customer Churn data set.}
	\label{fig:histogram numerical features classification}
\end{figure}

\begin{figure}
	\centering
	\includegraphics[width=.90\textwidth]{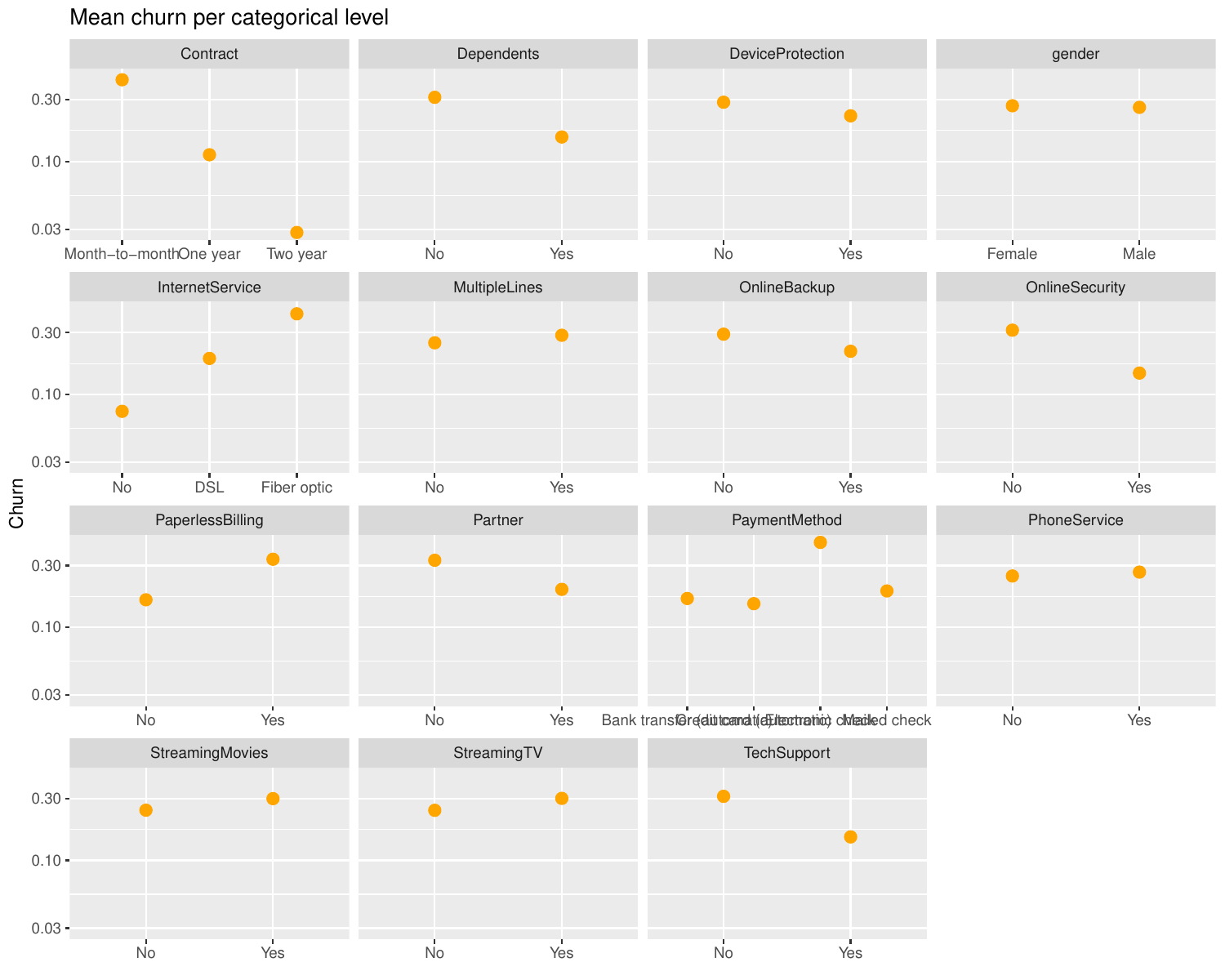}
	\caption{Mean response \variable{Churn} conditional on categorical features of Telco Customer Churn data set.}
	\label{fig:rel plot categorical features classification}
\end{figure}

\begin{figure}
	\centering
	\includegraphics[width=.90\textwidth]{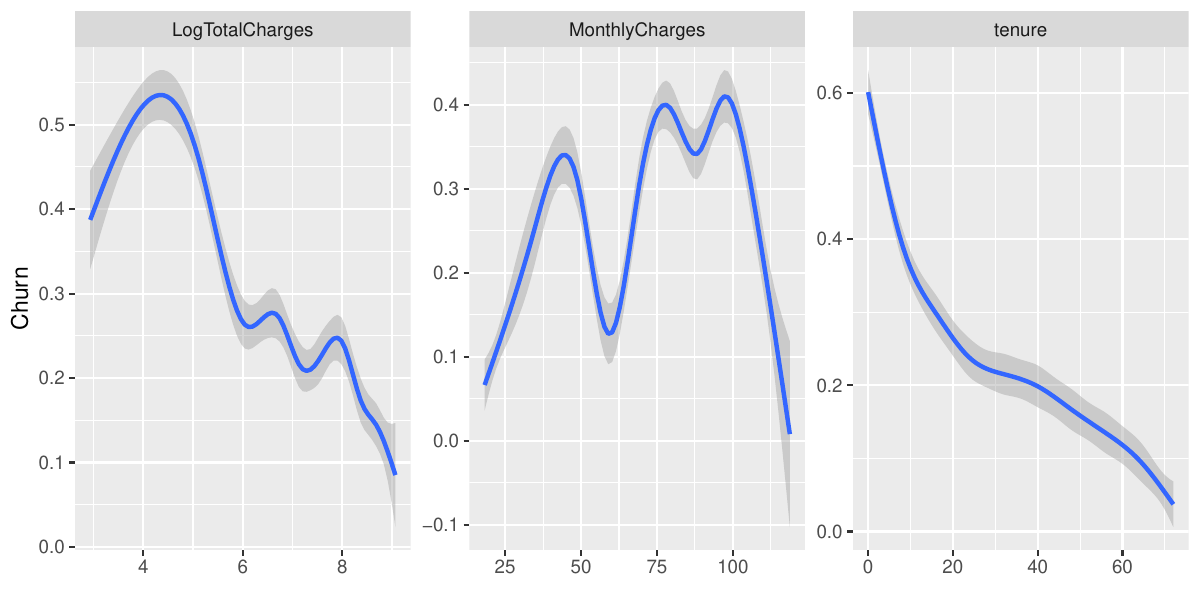}
	\caption{Density plots of response \variable{Churn} conditional on numerical features of Telco Customer Churn data set.}
	\label{fig:rel plot numerical features classification}
\end{figure}

See Figures~\ref{fig:histogram categorical features classification}, \ref{fig:histogram numerical features classification}, \ref{fig:rel plot categorical features classification} and \ref{fig:rel plot numerical features classification}.
\FloatBarrier

\section{Tweedie Deviance and Homogeneous Bregman Functions}
\label{sec:tweedie deviance}

In Eq.~(20) of \cite{gneiting_2011}, originally in \cite{Patton2011}, a rich family of homogeneous Bregman functions with parameter $b  \in \realnumbers$ is introduced for the positive real line $z, y\in \realnumbers_+$ as
\begin{equation*}
	S_b(z, y) = 
	\begin{cases}
		\frac{1}{b(b-1)} (y^b - z^b) - \frac{1}{b-1} z^{b-1} (y-z) & b \in \realnumbers \setminus \{0, 1\} \\
		y \log\frac{y}{z} -y+z & b=1  \\
		\frac{y}{z} - \log\frac{y}{z} -1 & b=0.
   \end{cases}
\end{equation*}
Up to a multiplicative constant, this coincides with the homogeneous scores in \autoref{tab:scoring_functions} for $b = a > 1$.

The Tweedie distributions $Tw_p(\mu, \varphi)$ \cite{Tweedie1984AnIW} are a subfamily of the exponential dispersion family (EDF), see \cite{Jorgensen1997,Wuthrich2021StatisticalFoundations}, and are characterised by their mean--variance relation $\Var[Y] = \varphi \cdot \mu^p$ with mean $\mu=\E[Y]$, dispersion parameter $\varphi > 0$ and power $p$.
Their deviance is given by
\begin{equation*}
	d_p(y, \mu) =
	2 \cdot
	\begin{cases}
		\frac{\max(0, y^{2-p})}{(1-p)\cdot(2-p)}-\frac{y\cdot\mu^{1-p}}{1-p}+\frac{\mu^{2-p}}{2-p} & p \in \realnumbers \setminus (0,1] \cup \{2\} \\
		y\log\frac{y}{\mu} - y + \mu & p=1 \\
		\frac{y}{\mu} - \log\frac{y}{\mu} - 1 & p=2
	\end{cases}
\end{equation*}
with valid domains:
\begin{center}
\begin{tabular}{llll} 
	\toprule
	response $y$& mean $\mu$ & power $p$\\ 
	\midrule 
	$y \in \realnumbers$ & $\mu \in \realnumbers_+$ & $p < 0$ \\
	$y \in \realnumbers$ & $\mu \in \realnumbers$  & $p = 0$ \\
	$y \in \realnumbers_+\cup \{0\}$ & $\mu \in \realnumbers_+$ & $1 \leq p < 2$\\
	$y \in \realnumbers_+$  & $\mu \in \realnumbers_+$  & $p \geq 2$\\
	\bottomrule
\end{tabular}
\end{center}
Well-known members are the Normal distribution ($p=0$), the Poisson distribution ($p=1$), the Gamma distribution ($p=2$), and the inverse Gaussian distribution ($p=3$).
Remarkably, for $p \in (0, 1)$, no Tweedie distribution exists while the Bregman function $S_b$ is defined for all $b \in \realnumbers$.

On the common domains of $y$ and $z$, it holds that
\begin{equation*}
	d_p(y, x) = 2 \cdot S_{2-p}(x, y) \,.
\end{equation*}
This indicates that Tweedie deviances for $p \in (0, 1)$ may be used as consistent scoring function for the expectation, although no distribution exists for this parameter range.
On the other hand, the domain of $z$ for a Bregman function can be extended to allow for $z=0$ for $0 < b \leq 1$.

Furthermore, Tweedie distributions are the only distributions within the EDF that are closed under scale transformations \cite{Jorgensen1997}: If $Y \sim Tw_p(\mu, \varphi)$ then $tY \sim Tw_p(t\mu, t^{2-p}\varphi)$ for any $t>0$.
We recover the degree of homogeneity of Tweedie deviances, $h=2-p$, as the law of transformation of the dispersion parameter of Tweedie random variables under scale transformations.

\section{Simulation study on efficiency}
\label{sec:Efficiency aspects}

As outlined in \autoref{subsubsec:Neyman-Pearson}, the coefficient of variation $\CV(Z) = \left(\Var[Z]/\E[Z]^2\right)^{\sfrac{1}{2}}$, where $Z$ is the score difference, determines the asymptotic power of a Diebold--Mariano test. % \sfrac instead of just 1/2 to avoid a too long line
To be more precise and to take into account the sample size $n$ of the test data set, the power is inversely linked to the coefficient of variation of the average score difference
\begin{equation*}
	\overline{Z}_n = \frac{1}{n} \sum_{i=1}^n S(m_A(\bX_i), Y_i) - S(m_B(\bX_i), Y_i)\,.
\end{equation*}

\begin{figure}
	\centering
	\includegraphics[width=.9\textwidth]{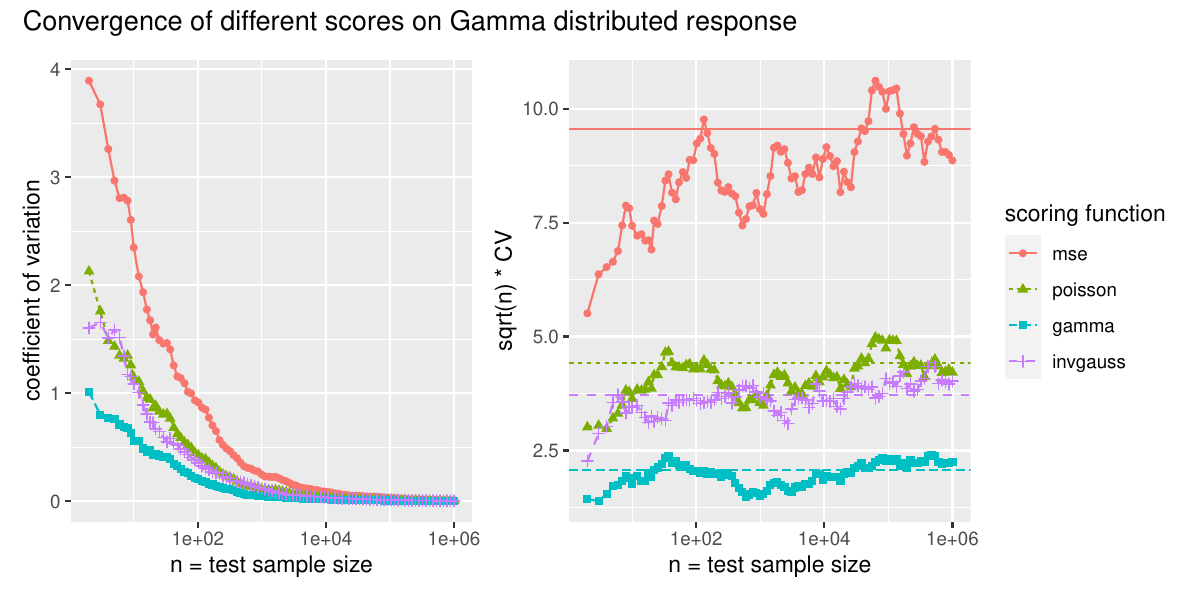}
	\caption{Speed of convergence with increasing test sample size of several scoring functions in terms of $\CV(\overline{Z}_n)$ evaluated over \num{100} simulations.
	Horizontal lines in the right figure indicate the value of $\CV(Z)$ based on $10^8$ data points.}
	\label{fig:efficiency}
\end{figure}
\begin{figure}
	\centering
	\includegraphics[width=.93\textwidth]{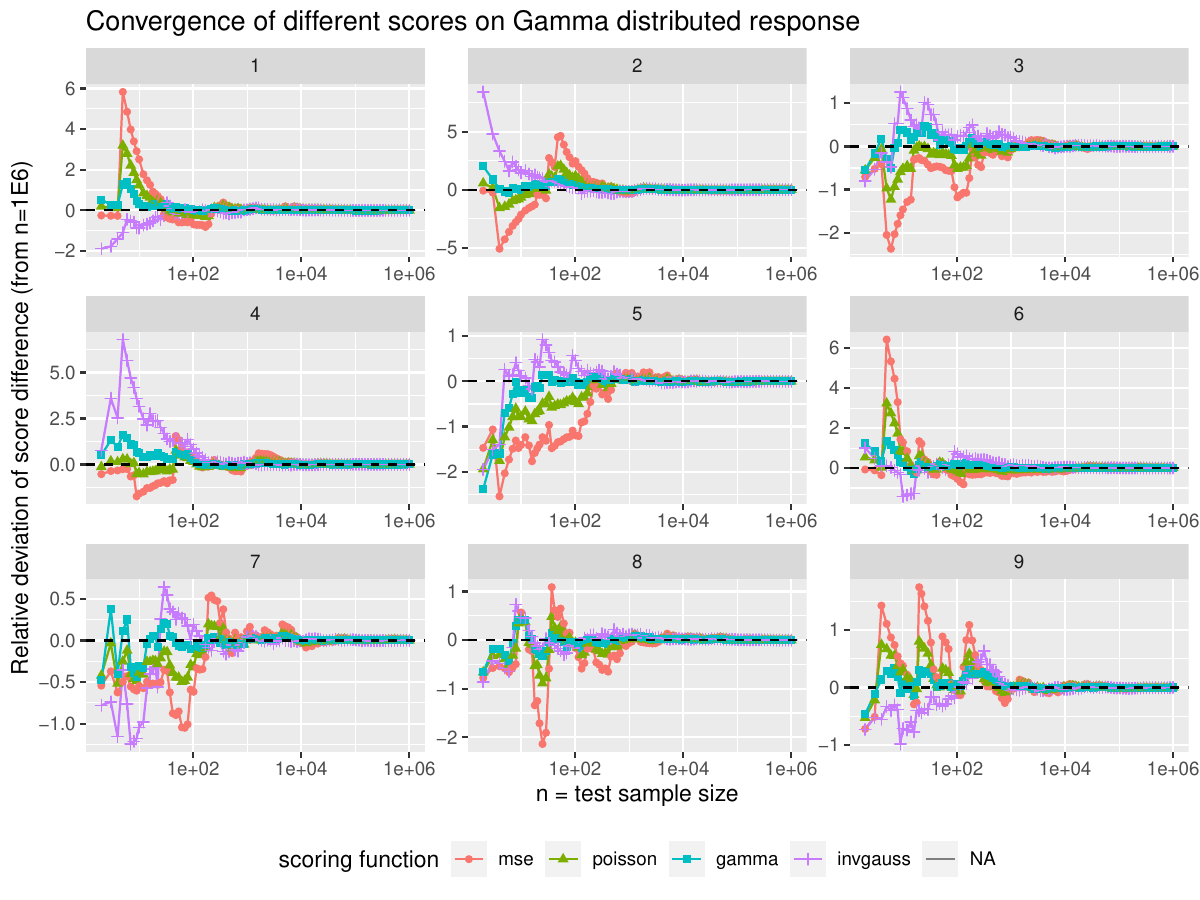}
	\caption{Speed of convergence with increasing test sample size of several scoring functions for 9 different simulations of test sets.
	Values on $y$-axis are the relative deviation of score differences with respect to their respective values at $n=10^6$.}
	\label{fig:efficiency_single_sims}
\end{figure}
As an empirical illustration of the convergence speed, we simulate a Gamma distributed response, conditionally on two features $\bX = (X_1,X_2)$, one categorical and one numerical feature, $Y | \bX \sim \mathrm{Gamma}$.
\begin{lstlisting}
generate_data <- function(n_samples, seed = NA) {
	if (!is.na(seed) && seed%%1==0) {
		set.seed(seed)
	}
	df <- tibble(
		color = sample(c("red", "blue"), n_samples, replace = TRUE, prob = c(0.2, 0.8)),
		length = runif(n_samples, min = -2, max = 2)) %>% 
	mutate(color = as.factor(color))
	
	true_mean <- exp(model.matrix(~ color + length, data = df) %*% c(4, -2, 1))
	dispersion <- 2  # results are highly sensitive to this parameter
	# E[Y] = shape * scale = mu
	# Var[Y] = shape * scale**2 = dispersion * mu**2
	# shape = 1 / dispersion
	# scale = mu * dispersion
	df$y <- rgamma(n_samples, shape = 1 / dispersion, scale = true_mean * dispersion)
	df
}
\end{lstlisting}
We fit a trivial model as $m_A$ as well as a (correctly specified) Gamma GLM with log link as $m_B$ on a training sample of $\text{size} = \num{1000}$.
We evaluate different scoring functions---squared error (corresponding to Gaussian deviance), Poisson deviance, Gamma deviance and inverse Gaussian deviance, which are Tweedie deviances with power $p=0, 1, 2, 3$---on a sequence of nested test sets with increasing size $n = 1, \ldots, 10^6$.

Take the following results with a grain of salt as they are very sensitive to slight changes of the distribution of the data.
The relative deviation of the score difference $Z$ on different test sets can be seen in \autoref{fig:efficiency_single_sims}.
We repeat this simulation \num{100} times with different random seeds and calculate the coefficient of variation for each test size $n$.
\autoref{fig:efficiency} visualises $\CV(\overline{Z}_n)$ as well as $\sqrt{n}\CV(\overline{Z}_n)$, taking into account the convergence rate of $n^{-1/2}$ of $\CV(\overline{Z}_n)$.
It further exhibits the theoretical quantity of $\sqrt{n}\CV(Z)$ (based on a Monte Carlo simulation with size $10^8$).
It is well visible that the Gamma deviance converges fastest with inverse Gaussian and Poisson deviance close together as second.
The squared error seems to suffer from some data points on which it produces outliers.\footnote{%
	We even observed erratic behaviour of the inverse Gaussian deviance for values of $y$ very close to zero when experimenting with different parameters of data distributions.
}
This simulation study suggests that scores derived from a quasi maximum likelihood approach of the underlying data distribution exhibit a fast convergence.
This underpins the theoretical intuition provided in \autoref{subsubsec:Neyman-Pearson}. 
We emphasise that this intuition ignored the presence of features.
This could be the reason for the sensitivity of the results with respect to small changes in the distribution of the data.
We encourage further research which takes the presence of feature information into account.

\end{document}